\documentclass[lettersize,journal,compsoc]{IEEEtran}
\usepackage{graphicx}
\usepackage{amssymb}
\usepackage{pifont}
\usepackage{amsmath,amsfonts}
\usepackage{algorithmic}
\usepackage{algorithm}
\usepackage{caption}
\usepackage{array}
\usepackage{textcomp}
\usepackage{stfloats}
\usepackage{url}
\usepackage{verbatim}
\usepackage{graphicx}
\usepackage{cite}
\usepackage{multirow}
\usepackage{soul}
\usepackage[table]{xcolor}
\usepackage{color}
\usepackage{multirow,booktabs,hhline}
\usepackage{xspace}
\usepackage{tabularx}
\usepackage{makecell}
\usepackage{arydshln}
\usepackage{hyperref}

\definecolor{lightergray}{RGB}{230,230,230}
\definecolor{DarkGreen}{RGB}{30,130,30}
\definecolor{red}{RGB}{255,0,0}
\newcommand{\cmark}{\textcolor{DarkGreen}{\ding{51}}}
\newcommand{\xmark}{\textcolor{red}{\ding{55}}}

\newcolumntype{C}[1]{>{\centering\arraybackslash}p{#1}}
\newcommand{\ie}{{\emph{i.e.}}\xspace}
\newcommand{\eg}{{\emph{e.g.}}}
\newcommand{\etc}{\emph{etc}\xspace}
\newcommand{\etal}{{\emph{et al.}}\xspace}

\definecolor{lightergray}{RGB}{230,230,230}
\definecolor{DarkGreen}{RGB}{30,130,30}
\definecolor{red}{RGB}{255,0,0}

\begin{document}

\title{LLMs Meet Multimodal Generation and Editing:\\A Survey}

\author{
Yingqing He$^\dagger$$^*$, Zhaoyang Liu$^\dagger$$^*$, Jingye Chen$^*$, Zeyue Tian$^*$, Hongyu Liu$^*$, Xiaowei Chi$^*$, \\Runtao Liu$^*$, Ruibin Yuan$^*$, Yazhou Xing$^*$, Wenhai Wang, Jifeng Dai, Yong Zhang, \\
Wei Xue, Qifeng Liu, Yike Guo, Qifeng Chen\\
 
\IEEEcompsocitemizethanks{
\IEEEcompsocthanksitem{$^\dagger$ Project leaders; $^*$ Co-first authors.}
\IEEEcompsocthanksitem{Yingqing He, Zhaoyang Liu, Jingye Chen, Zeyue Tian, Hongyu Liu, Xiaowei Chi, 
Runtao Liu, Ruibin Yuan, Yazhou Xing, Wei Xue, Qifeng Liu, Yike Guo, and Qifeng Chen are with The Hong Kong University of Science and Technology, Hong Kong SAR.}
\IEEEcompsocthanksitem{Wenhai Wang is with The Chinese University of Hong Kong, Hong Kong SAR.}
\IEEEcompsocthanksitem{Jifeng Dai is with Tsinghua University, China.} 
\IEEEcompsocthanksitem{Yong Zhang is with the Tencent AI Lab, China.}
}}



\IEEEtitleabstractindextext{
\begin{abstract}
With the recent advancement in large language models (LLMs), there is a growing interest in combining LLMs with multimodal learning. 
Previous surveys of multimodal large language models (MLLMs) mainly focus on multimodal \textit{understanding}.
This survey elaborates on \textit{multimodal generation and editing} across various domains, comprising image, video, 3D, and audio.
Specifically, we summarize the notable advancements with milestone works in these fields and categorize these studies into LLM-based and CLIP/T5-based methods.
Then, we summarize the various roles of LLMs in multimodal generation and exhaustively investigate the critical technical components behind these methods and the multimodal datasets utilized in these studies. 
Additionally, we dig into tool-augmented multimodal agents that can leverage existing generative models for human-computer interaction.
Lastly, we discuss the advancements in the generative AI safety field, investigate emerging applications, and discuss future prospects. 
Our work provides a systematic and insightful overview of multimodal generation and processing, which is expected to advance the development of Artificial Intelligence for Generative Content (AIGC) and world models.
A curated list of all related papers can be found at \url{https://github.com/YingqingHe/Awesome-LLMs-meet-Multimodal-Generation}.
\end{abstract}

}

\maketitle

\begin{figure*}[th]
    \centering
    \includegraphics[width=\linewidth]{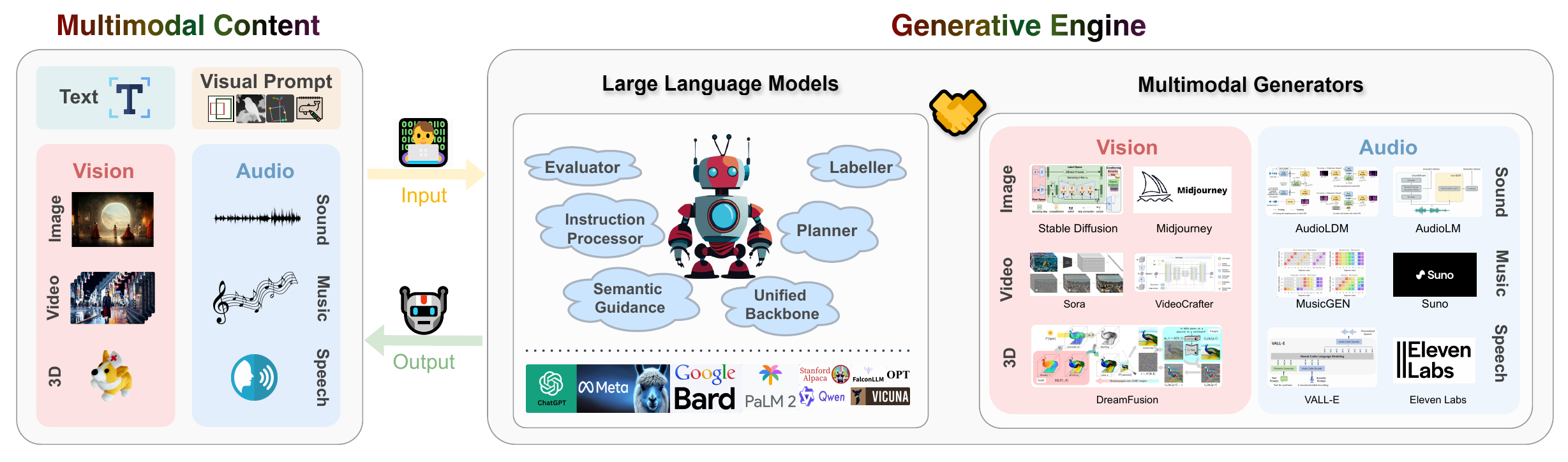}
    \caption{
    Our main goal is to investigate the roles of LLMs in the task of language-guided multimodal generation.
    The modalities we focused on consist of image, video, 3D, and audio (including sound, music, and speech).
    }
    \label{fig:intro}
\end{figure*}
\section{Introduction}
\label{sec:intro}
The interactions between human beings and the physical world involve information from multiple modalities, such as language, vision, and audio. 
Therefore, realizing a world simulator also requires the model to be able to perceive and respond with multimodal information in a flexible manner.
Recently, OpenAI introduced a foundation text-to-video generation model termed Sora~\cite{sora} that is capable of generating highly realistic videos as world simulators.
It makes great progress in simulating or generating real-world scenes but is unable to generate other modalities, such as text, 3D, and audio. 
Also, it lacks the ability to perceive other modalities such as image, video, 3D, and audio, making it an incomprehension world simulator.

In the past few years, researchers focused on the generation of each single modality and have achieved great progress:
In terms of text generation, we have witnessed a qualitative leap in the performance of natural language processing tasks:
From BERT~\cite{devlin2018bert}, GPT1~\cite{2018gpt1}, GPT2~\cite{2019gpt2}, GPT3~\cite{gpt3}, GPT4~\cite{gpt4} to ChatGPT~\cite{chatgpt}, LLaMA~\cite{li2023llama,gao2023llama}, the number of model parameters and training samples has grown rapidly, resulting in the continual growth of modal abilities and product deployment.
In the visual generation field, with the rapid progress of diffusion models and large-scale image-text datasets, text-to-image (T2I) generation has achieved remarkable achievement and can synthesize high-quality images based on various user-provided text prompts, such as SDXL~\cite{sdxl} and PIXART-$\alpha$~\cite{chen2023pixart}.
%
Subsequently, significant advancements have been made in the field of text-to-video generation through the utilization of video diffusion models~\cite{vdm} and large-scale video-language datasets~\cite{Webvid}. 
Notably, several milestone works have emerged, such as~\cite{he2022lvdm, zhou2022magicvideo, singer2022make-a-video, ho2022imagen-video, villegas2022phenaki, chen2023videocrafter1, guo2023animatediff, bar2024lumiere, girdhar2023emuvideo, videocrafter2} and Sora~\cite{sora}.
For the 3D generation, with the emergence of the CLIP~\cite{radford2021learning} model, some methods~\cite{sanghi2022clip,mohammad2022clip,michel2022text2mesh} try to align the text information to the rendered images from 3D representations, \ie, mesh, point cloud,  NeRF~\cite{wang2023nerf} and gaussian splatting~\cite{kerbl3Dgaussians}). 
These approaches have led to significant developments in text-to-3D generation.  
Additionally, the integration of Stable Diffusion (SD)~\cite{ldm} with text-to-image rendering has enabled a series of works in text-to-3D generation~\cite{yi2023gaussiandreamer, tang2023dreamgaussian, hoellein2023text2room, liang2023_luciddreamer, yu2023_csd, li2023_sweetdreamer, wang2023prolificdreamer, lorraine2023_att3d, xu2023_dream3d, zhu2023_hifa, Chen_2023_ICCV, tsalicoglou2023textmesh, poole2022dreamfusion, lin2023magic3d,seo2023let}. 
The powerful text-to-image model helps the 3D generation achieve higher performance and better visual results. 
In the area of text-to-audio generation, a series of representative works tackle different audio domains 
such as~\cite{liu2023audioldm, liu2023audioldm2, kreuk2022audiogen} for text-to-audio,~\cite{agostinelli2023musiclm, copet2024musicgen, forsgren2022riffusion} for text-to-music, and~\cite{tan2024naturalspeech, shen2023naturalspeech2, ju2024naturalspeech3, wang2023valle, jiang2023megatts2, ren2020fastspeech2} for text-to-speech, and they have achieved significant performance in generating high-quality natural sounds, music, and human-level speech.

With the notable progress and significant performance improvements of Large Language Models (LLMs), other non-text modalities have started harnessing the power of LLMs to either enhance their generation quality or integrate multiple modalities in a unified system to achieve more powerful functionalities.
%
In the context of image generation, the integration of LLMs can be divided into two categories.
%
The first category involves encoding visual information into discrete token indices, trying to unify visual understanding and generation~\cite{ge2023planting, zeqiang2023mini, tang2023codi, ge2023making, sun2023emu2, zhao2023making}. 
Specifically, visual information is encoded into token representations, and LLMs directly comprehend and generate visual tokens, enabling simultaneous visual understanding and generation.
The second category focuses on leveraging LLMs to enhance the generation quality of existing pretrained T2I models: 
One type of approach utilizes an LLM as a layout planner to incorporate knowledge of object spatial positions, quantity, and object size, enabling the generation of required bounding boxes~\cite{chen2023textdiffuser2, lian2023llm, feng2023layoutgpt, zhang2023controllable, qu2023layoutllm}.
After obtaining the bounding boxes, the images can be generated through a grounded T2I model such as GLIGEN~\cite{li2023gligen}.
Another approach utilizes LLMs to expand input user prompts~\cite{betker2023improving}: 
By providing highly detailed and comprehensive prompts, LLMs generate images with high quality and richness.
With the assistance of LLMs, image generation has achieved higher generation quality, improved prompt following capabilities, dialogic function, and user-friendly interface.

Similar to image domain, in video generation, LLMs serve as the general backbone for unified multimodal joint generation~\cite{kondratyuk2023videopoet, yu2023language}, video layout planning~\cite{fei2023empowering, lin2023videodirectorgpt, lian2023llm, lv2023gpt4motion, lu2023flowzero} and temporal prompt generation~\cite{hong2023large, huang2023free, wang2023intercontrol, liu2023plan, long2024videodrafter} for temporal dynamics guidance. 
For 3D generation and editing, LLMs serve as a bridge between users and 3D assets, which improves interaction efficiency~\cite{sun20233d,feng2023posegpt} and helps users understand the 3D assets~\cite{chen2023ll3da,wu2023gpteval3d}.
In the context of audio generation and editing,
the role of LLMs primarily lies in serving as coordinated backbones for multimodal audio~\cite{zhang2023speechgpt, gong2023listen,deshmukh2023pengi, rubenstein2023audiopalm, liu2024music, chen2023lauragpt, gardner2023llark, tang2023salmonn, chu2023qwen-audio, hussain2023m,shu2023llasm,yuan2024chatmusician,ding2024songcomposer}, conditioners for specific tasks~\cite{wu2023music, ghosal2023text, wu2024improving}, labelers for audio understanding~\cite{huang2023make2, wang2023assessing, vyas2023audiobox}, agents for interactive generation and editing~\cite{shen2023hugginggpt, huang2023audiogpt, liu2023wavjourney, yu2023musicagent, zhang2023loop, zhuo2023lyricwhiz}, as well as inspiration for novel approaches~\cite{wang2023valle, agostinelli2023musiclm, dhariwal2020jukebox, copet2024musicgen, borsos2023audiolm, yang2023uniaudio}.
The growing utilization of LLMs in the audio domain is not only transforming our way of engaging with sound and music but also expanding the boundaries at the crossroads of AGI and audio technologies. 
Besides, multimodal agents~\cite{shen2023hugginggpt,liu2023controlllm,yang2023gpt4tools,wu2023visual,liu2023internchat,li2023modelscope} integrate lots of AIGC tools into the framework as a universal system, which relies on LLMs to invoke tools but endows LLMs with the ability to comprehend and generate content of non-text modalities.
Generally, LLMs significantly play an indispensable role in generating various modes of content.

To promote the development of multimodal generation and empower the world simulator, in this work, we provide a comprehensive review of works involving LLMs in the generation of multiple modalities. 
As shown in Fig.~\ref{fig:intro}, 
we summarize the roles of LLMs into several key aspects, such as evaluator, labeller, instruction processor, planner, provider of semantic guidance, or as backbone architectures. 
Additionally, we discuss the development in the generative AI safety topic in Sec.~\ref{sec: safety}, with the emerged applications, and the potential future prospects in Sec.~\ref{sec: applications} and Sec.~\ref{sec: future_work}.

We summarize our contributions as follows:
\begin{itemize}
    \item We present the first systematic review of LLMs applied to the generation and editing of multiple modalities, including images, videos, 3D, and audio. 
    \item We discuss the evolution of generative techniques through a comparative analysis of pre-LLM and post-LLM eras, offering a clear perspective on the progression and refinement of these approaches. 
    \item We summarize the various roles of LLMs in the generation or editing process for each modality from a technical view.
    \item  We discuss important AI safety issues, investigate emerging applications and explore future directions to boost the development of multimodal generation and world models.
\end{itemize}

\subsection{Scope}
This survey explores the generation of multiple modalities, including images, videos, 3D models, and audio. 
The multimodal generation in our survey includes the separate generation of different modalities as well as the joint generation of multiple modalities.
We will not dig into pure text generation and processing extensively, as there have already been many surveys specifically focusing on the advancements in that field~\cite{survey-llm-2023, survey-llm-2024, survey-llm-eval}. 
Our primary focus is on how the recent emergence of LLMs in the past few years can assist in the generation of other vision and audio modalities, especially in the open-domain generation. 
This will aid us in designing better unified generative models for multi-modalities. 
Note that the tasks and works we discussed are primarily \textit{language-based} generation and editing.
Unconditional generation and other non-text-based editing are not our primary focus since they are either limited to a small domain or lack flexibility and controllability.
In detail, we focus on the following tasks:

\begin{itemize}
    \item \textbf{Text-to-image generation and editing}: Image generation aims to create various open-domain image contents, including pictures, photos, or stylized drawings from user-provided textual descriptions. Image editing aims to modify the input image content and can be based on user instructions.
    \item \textbf{Text-to-video generation and editing}, where models generate or modify arbitrary and various dynamic visual contents guided by free-form text descriptions.
    \item \textbf{Text-to-3D generation and editing}, which is a task for generating and editing 3D objects, scenes, or avatars with user-provided textual descriptions.
    \item \textbf{Text-to-audio generation and editing}, where textual descriptions are used to generate audio, including general sounds, music, and speech. Audio editing tasks, such as adding, removing, or inpainting, can all be performed by modifying existing audio content through textual descriptions.
    \item \textbf{Multimodal generative agents}, which enable LLMs to handle data across different modalities by utilizing a variety of specialized multimodal tools.
    \item \textbf{Generative AI Safety}, which focuses on reducing toxic and biased content, protecting copyright, and addressing the creation of fabricated content by multimodal generative models.
\end{itemize}

\subsection{Content Overview}
We first review related surveys on both single modality generation and LLMs in Sec.~\ref{sec: related_work}. 
We then briefly review the basic techniques of representative generative models, multimodal alignment models, LLMs, and MLLMs in Sec.~\ref{sec: prelimi}. 
Next, we review the LLM-based vision generation in different visual modalities, including image in Sec.~\ref{sec: image}, video in Sec.~\ref{sec: video}, 3D in Sec.~\ref{sec: 3D}, audio modality in Sec.~\ref{sec: audio}, and multimodal agents in Sec.~\ref{sec: agent}, respectively. 
Lastly, we review the safety aspect of generative AI in Sec.~\ref{sec: safety}, emerging applications in Sec.~\ref{sec: applications}, and potential future directions for the LLMs-based multimodal generation field in Sec.~\ref{sec: future_work}.

\section{Related Surveys}
\label{sec: related_work}
\textbf{Survey on modality-specific generation.}
A series of surveys focus on single modality generation, such as \cite{survey-t2idiffusion} for image generation, \cite{survey-vdm} for video generation, \cite{survey-3d-representation} for 3D generation, \cite{survey-audio-speech} for audio generation. 
However, the previous generation paradigm mainly employs pretrained CLIP \cite{radford2021learning}, CLIP-related variant \cite{eva-clip}, or language encoder T5 \cite{T5} to achieve open-domain text-guided generation. 
With the emergence of LLMs, there is a growing trend of leveraging powerful LLMs to enhance the generation of content for each modality. 
Our work aims to provide a comprehensive survey on the role of LLMs in the generation of various types of modalities, an aspect that is absent from previous surveys.

\noindent\textbf{Survey on LLMs and MLLMs.}
Numerous surveys have been conducted to explore various aspects of LLMs. 
For example, \cite{survey-llm-agent} offers a comprehensive examination of LLMs-based autonomous agents. 
Additionally, \cite{survey-mllm}, and \cite{survey-mllm-2} look into MLLMs, introducing papers that combine LLMs with other non-text modalities.
They review papers on multimodal understanding and generation in a mixed manner, introducing primarily multimodal understanding works and focusing less on multimodal generation. 
In contrast, our work primarily concentrates on the \textit{generation} aspect, aiming to investigate the performance and functionality improvement that LLMs bring to each modality's generation process, leading to a better AI-generated world with various modalities.

\section{Preliminaries}
\label{sec: prelimi}

In this section, we first review different types of generative models in Sec.~\ref{sec: prelimi_gen}. 
Then, we illustrate multimodal alignment models in Sec.~\ref{sec: prelimi_mmalign}.
Lastly, We introduce the technical principle of large language models in Sec.~\ref{sec: prelimi_llm}, and explain multimodal large language models in Sec.~\ref{sec: prelimi_mllm}

\begin{figure}[h]
    \centering
    \includegraphics[width=1\linewidth ]{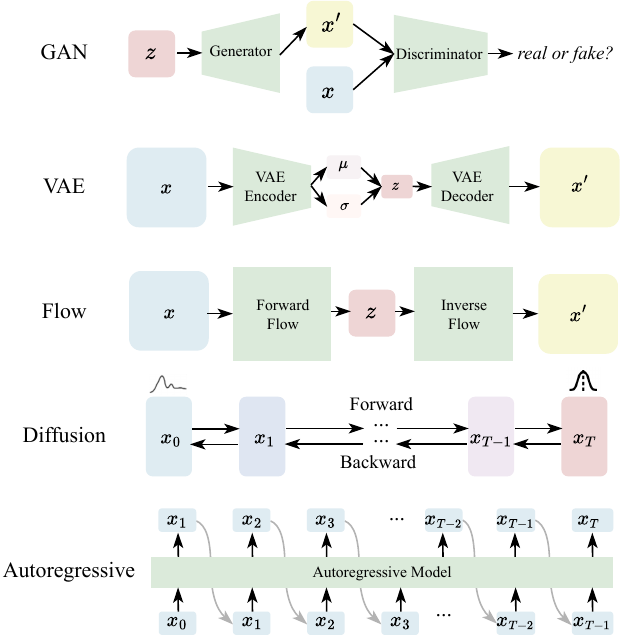}
    \caption{The illustration of generative models.
    In this picture, $x$ and $x_0$ indicate the sample from the real data distribution,  $x'$ stands for the sample from the model's estimated data distribution, and $z$ means the latent sampled from a prior distribution (typically a Gaussian distribution).
    }
    \label{fig:preliminaris_gen_model}
\end{figure}

\subsection{Generative Models}
\label{sec: prelimi_gen}
 
We review the core principles and basic concepts of classic generative models, including generative adversarial Networks (GANs), variational autoencoders (VAEs), flow-based models, diffusion models, and autoregressive models.

The generation process of generative models can be characterized as a transformation from a latent sample $z$ drawn from a prior distribution $p_\mathbf{z}(\mathbf{z})$ to a generated sample $\mathbf{x}'$ from a real data distribution $p_\text{data}(\mathbf{x})$ that aligns with the target data distribution. 
Specifically, the latent variable $\mathbf{z}$ is passed through a parametric function, typically implemented as a neural network, which learns to map the prior distribution to the target data distribution. 
The output $\mathbf{x}'$ of this transformation is then regarded as a synthetic instance that mimics the statistical properties of the original data distribution, which may correspond to various modalities such as images, videos, 3D representations, audio, or text.

\subsubsection{Generative Adversarial Networks}
GAN~\cite{goodfellow2020generative} has achieved promising results in various tasks during years of development. 
As shown in Fig.~\ref{fig:preliminaris_gen_model}, GAN comprises two crucial components: a discriminator ($D$) and a generator ($G$). The discriminator is designed to distinguish between real and fake samples. 
The generator aims to create fake samples that are indistinguishable from the real data and try to fool the discriminator.
During training, the $G$ and $D$ are trained simultaneously and play a two-player minimax game.
The optimization objective is formulated as follows:
\begin{align}
& \min_G \max_D V(D, G) = \notag\\
& \mathbb{E}_{\mathbf{x} \sim p_{\text{data}}(\mathbf{x})}[\log D(\mathbf{x})] + \mathbb{E}_{\mathbf{z} \sim p_\mathbf{z}(\mathbf{z})}[\log(1 - D(G(\mathbf{z})))] \notag \\
\end{align}
Where $D(\mathbf{x})$ indicates the probability of real sample $\mathbf{x}$ being a real sample, and $D(G(\mathbf{z}))$ indicates the probability of a generated sample being a fake sample.
$ \mathbb{E}_x$ is the expectation value over all samples. 

\subsubsection{Variational Auto-Encoder}
Variational auto-encoder~\cite{vae} contains an encoder and a decoder to learn latent representations from the input data, as shown in the second row in Fig.~\ref{fig:preliminaris_gen_model}.

Encoder is a neural network that maps the input data $\mathbf{x}$ to the distribution of the latent space variable $\mathbf{z}$. Then, the variational posterior distribution  $q(\mathbf{z}|\mathbf{x})$ is usually assumed to be a Gaussian distribution $N(\mu,\sigma^2)$. In this case, the encoder gives the $\mu$ and $ {\sigma}^2$.

Decoder maps the latent space variable $\mathbf{z}$ back to the input space $\mathbf{x}'$, yielding the conditional distribution of the generated data $x$,$q(x|z)$.

The training optimization target of VAE is to maximize the lower bound of the marginal log-likelihood of the data. This target can be achieved through stochastic gradient descent and the reparameterization trick. It is also known as Evidence Lower Bound (ELBO). Specifically, the ELBO can be written in the following form:

\begin{align}
& \mathcal{L}(\theta; \mathbf{X}) = \notag \\
& - \left[KL(q(\mathbf{z}|\mathbf{x}; \theta)||p(\mathbf{z})) - \mathbb{E}_{q(\mathbf{z}|\mathbf{x}; \theta)}[\log p(\mathbf{x}|\mathbf{z}; \theta)]\right]  \notag \\
\end{align}

where $KL(·||·)$ denotes the KL divergence, which measures the difference between the posterior distribution inferred by the encoder $q(\mathbf{z}|\mathbf{x})$ and the prior distribution $p(z)$. The second term is the reconstruction error, which represents the match between the $\mathbf{x'}$ generated from $z$ and the actual data $x$.

\textbf{Vector Quantized Variational Auto-Encoder} VQ-VAE\cite{esser2021taming} is a variant of the VAE that introduces a discrete latent space, which significantly improves the quality of the generated samples compared to the original VAE.
In VQ-VAE, the encoder maps the continuous output of the encoder to the nearest point in a predefined discrete codebook and outputs a discrete latent representation $\mathbf{z}$. The codebook is learned jointly with the rest of the model parameters. Using a discrete latent space allows VQ-VAE to capture more global and structured information about the data, leading to better generation quality.

\subsubsection{Flow-based Model}
Flow-based model, also known as normalizing flows, is a class of generative models successfully applied in various tasks, including image synthesis, variational inference, and unsupervised representation learning.
The architecture of a flow-based model consists of a sequence of invertible transformations (or flows). Each flow is parameterized by a neural network, which learns to transform the data distribution step by step into the simpler prior distribution.
The objective function for training a flow-based model is the negative log-likelihood of the data under the model, which can be computed exactly due to the invertibility of the flows. The function is given by:
\begin{align} 
\mathcal{L}(\theta) = -\mathbb{E}_{\mathbf{x} \sim p_{\text{data}}(\mathbf{x})}[\log p_{\text{model}}(\mathbf{x}; \theta)] 
\end{align}
where $p_{\text{model}}(\mathbf{x}; \theta)$ is the probability density function of the model. In practice, the transformations used in flow-based models are chosen to be easily invertible and have easily computable Jacobians, such as affine transformations.

\subsubsection{Diffusion Model}
The diffusion model is proposed in~\cite{sohl2015deep}, which first gives the prototype of recent diffusion models. However, the foundational structure of the modern diffusion model, which led to a revolution of generation paradigms, was proposed in Denoised Diffusion Probabilistic Models\cite{ho2020denoising}. It is elegant in training and mythical improvement and only introduced a simple regression loss.
As shown in Fig.~\ref{fig:preliminaris_gen_model}, the diffusion model turned the complex work into a series of denoising tasks, primarily consisting of two steps: the injection of prior noise into the data and the denoising prediction.

\textbf{Forward Noise Injection}
In the forward noise injection process, the model sequentially introduces Gaussian noise $\mathbf{\zeta_t}$ into the data in each step $t$ of $T$ steps. The process can be represented as follows:
\begin{equation}
\mathbf{x_{t+1}} = \sqrt{1 - \alpha_t^2}x_t + \alpha_t \mathbf{\zeta_t}
\end{equation}
Where $\mathbf{x_t}$ is the data at time $t$, $\mathbf{\zeta_t}$ is the Gaussian noise at time $t$, and $\mathbf{\alpha_t}$ is a noise schedule determining the amount of noise to be added at each step. The noise schedules $\mathbf{\alpha_t}$ typically start close to 0 and gradually increase to 1 over the $T$ steps.

The noise at each step is assumed to follow a Markov transition process, which means the noise at time $t$, $\mathbf{\zeta_t}$, is independent of the noises at all previous times. This assumption simplifies the model and makes it tractable to train.

\textbf{Reverse Denoising}
After the forward noise injection, the model aims to reverse the process by predicting the original data from the noisy version. 
This is done by learning a denoising function, which is typically parameterized as a deep neural network. The denoising function inputs the noisy data at time $t$ and attempts to predict the noise added at that step. This process is repeated for each time step, going backward from $T$ to 1.
The denoising function can be represented as follows:
\begin{equation}
\mathbf{\hat{\zeta}_t} = D_{\theta}(\mathbf{x_t}, t)
\end{equation}
where $D_{\theta}$ is the denoising function parameterized by $\theta$, $x_t$ is the noisy data at time $t$, and $\mathbf{\hat{\zeta}_t}$ is the predicted noise.

The model's objective during training is to minimize the difference between the predicted noise $\mathbf{\hat{\zeta}_t}$ and the actual noise $\mathbf{\zeta_t}$ added during the forward process. This can be measured using a simple mean squared error loss.

Training the model to perform this reverse denoising prediction allows it to generate data similar to the original data distribution, making the diffusion model a powerful tool for generative tasks. 

\subsubsection{Autoregressive Model}
Autoregressive models are another class of generative models widely used in various tasks, including time series forecasting, speech synthesis, and natural language processing. Their architecture predicts future values based on past values.

The objective function for training an autoregressive model is also the negative log-likelihood of the data under the model, which can be computed exactly due to the sequential nature of the model. This is given by:
\begin{align} 
\mathcal{L}(\theta) = -\mathbb{E}{\mathbf{x} \sim p_{\text{data}}(\mathbf{x})}[\log p_{\text{model}}(\mathbf{x_{t+1}}|\mathbf{x_{\leq t}}; \theta)] 
\end{align}

where $p_{\text{model}}(\mathbf{x_{t+1}}|\mathbf{x_{\leq t}}; \theta)$ is the conditional probability density function of the model. In practice, the model is trained to maximize the likelihood of the next value in the sequence given the previous values.

\subsection{Multimodal Alignment Model}
\label{sec: prelimi_mmalign}

CLIP~\cite{radford2021learning} is a ground-breaking image-language alignment model that simultaneously learns an image encoder and a text encoder to produce visual and textual representations in a shared semantic space, trained on a diverse range of internet text and images through contrastive learning~\cite{contrastivelearning}.
After its large-scale contrastive pretraining, it is capable of tackling various downstream tasks, including fine-grained object recognition, video action recognition, facial emotion recognition, geo-localization, and many others in a zero-shot manner.
Thanks to its web-scale pertaining, it can understand plenty of semantics. Thus, it has become one of the most widely used visual and textual encoders in various vision generation and editing works such as DALLE-2~\cite{dalle2hierar} and LDM~\cite{ldm} for text-to-image generation, VideoCrafter~\cite{chen2023videocrafter1} for text-to-video and CLIP-Nerf~\cite{wang2022clip} for 3D. 

Besides text and vision alignment, CLAP~\cite{elizalde2023clap} aligns text and audio information. The audio-aligned text embedding representation is used as the condition of AudioLDM~\cite{liu2023audioldm} for text-guided audio generation.

CAVP makes a further step towards video-audio alignment, which is trained in Diff-Foley~\cite{diff-foley} for the video-to-audio generation task. 
After training CAVP, Diff-Foley further trains a latent diffusion model, which is conditioned on the audio-aligned video representation to output synchronized audio signals.

Different from previous methods for the alignment of paired modalities, ImageBind\cite{girdhar2023imagebind} aligns six different modalities in one shared semantic space.
The supporting modalities include text, image, video, audio, depth, and thermal.
It has been used in multimodal generation tasks such as Next-GPT~\cite{wu2023next}, Seeing-and-Hearing~\cite{seeing-and-hearing}, and also multimodal understanding works such as PandaGPT~\cite{pandagpt}.

\subsection{Large Language Models}
\label{sec: prelimi_llm}

Modern Large Language Models utilize the transformer architecture to generate contextually rich embeddings. These models are trained on a large corpus of text and then fine-tuned for specific tasks. They generate text by predicting the next word in a sequence, given the previous words. 

Typical examples include LlaMA\cite{touvron2023llama} and GPT\cite{radford2018improving,radford2019language,brown2020language}, which are the autoregressive models that use only the left context to make predictions. They are mainly built by transformer decoder. The models would be pre-trained on large, diverse datasets to acquire a strong foundation of language understanding and generation capabilities and then fine-tuned on datasets that provide explicit instructions or guidance on specific tasks, such as question-answering summarization or code generation. Furthermore, tricks like Chain-of-Thought (CoT)~\cite{gpt3} fine-tuning and Reinforcement Learning from Human Feedback (RLHF)~\cite{rlhf} improve the models' task-specific ability.

\subsection{Multimodal Large Language Models}
\label{sec: prelimi_mllm}

Multimodal Large Language Models (MLLMs) are recently emerging models that aim to equip LLMs with the ability to understand or generate other modalities. 
MLLMs usually incorporate a few key components: extra pre-trained modality-specific encoders for feature extraction and input projectors for multimodal hidden feature alignment with the LLM backbone. 
For MLLMs with generation ability, they usually contain extra output projectors and corresponding modality generators as the generation ends. 
A series of works incorporating extra pre-trained encoders for encoding multimodal information to a pre-trained LLM and train modality alignment modules to achieve this~\cite{zhu2023minigpt, chen2023llava, wu2023next, pi2023perceptiongpt, pi2023detgpt, gao2023g}. 
Other works train the whole multimodal system in an end-to-end manner~\cite{sun2023generative}.
In the following sections, we will illustrate a range of recent MLLMs works, especially MLLMs on multimodal generation.

\begin{figure*}[h]
    \centering
    \includegraphics[width=\linewidth]{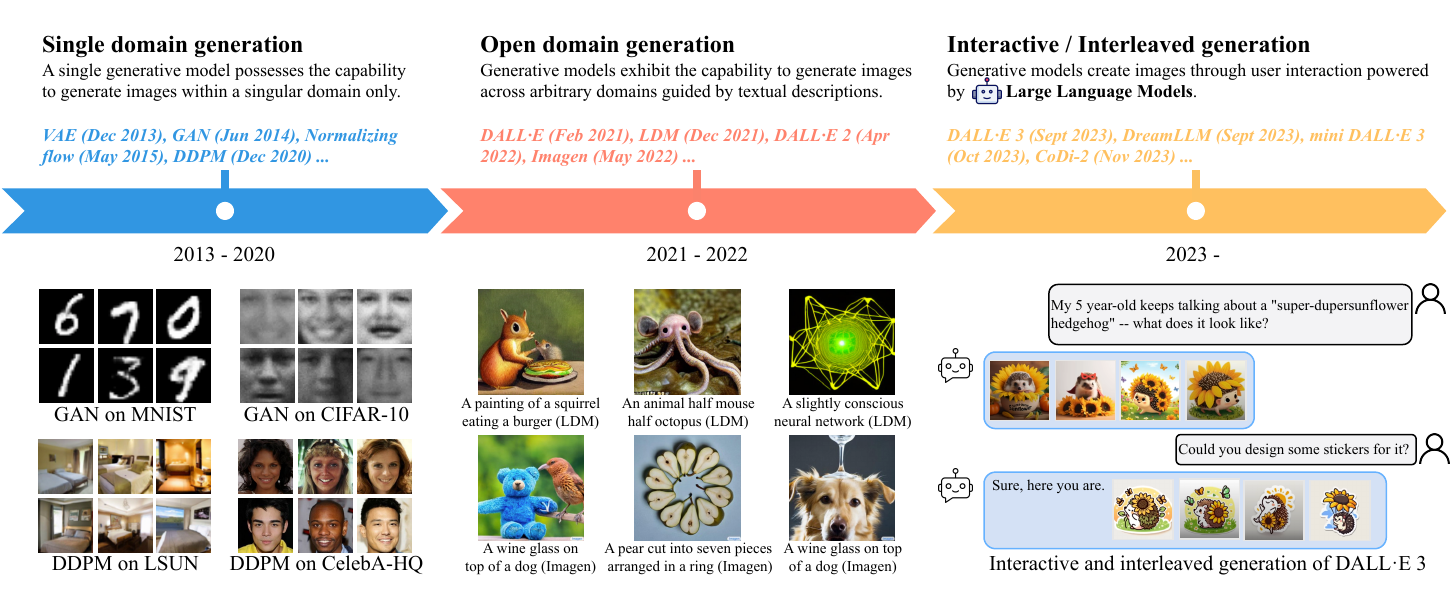}
    \caption{History review on the development trajectory of image generation. 
    Early works on image generation predominantly concentrated on synthesizing images within specific narrow domains, such as human faces or bedrooms~\cite{yu2015lsun,liu2018large}. 
    Subsequently, DALL-E~\cite{dalle} and Latent Diffusion Models (LDM)~\cite{ldm} have progressed to generate images through user prompts and support the synthesis of open-domain images. 
    In the recent two years, powered by LLMs, research has trended toward achieving a more intuitive and interactive image generation process, such as iterative generation through conversations~\cite{dong2023dreamllm,dalle3}.
    }
    \label{fig:image-history}
\end{figure*}

\begin{table*}[t]
\centering
\caption{
    Overview of existing methods using LLMs for language-based image generation. According to the role of LLMs in this task, these methods can be divided into four categories: multimodal LLMs for generation, image layout planning, prompt synthesis and refinement, and image quality evaluation. 
    In the "Task`` column, the "T`` and "I`` are the abbreviations of "text`` and "image``, respectively, while "Any`` represents the universal generation supporting text, image, video, and audio modality. 
    "-`` indicates that the information is not available in the official paper. 
}
\scalebox{1}{
\begin{tabular}{lcccccc}
 \toprule
\textbf{Method} & \textbf{Venue}  & \textbf{Task} & \textbf{LLM} & \textbf{Generative Model} & \textbf{Training Cost} \\
\midrule
\multicolumn{7}{l}{\textbf{\textit{Multimodal LLMs with image generation}}\;\;\;\;\;} \vspace{0.05cm}\\

FROMAGe~\cite{koh2023grounding} & ICML 2023 &  TI$\rightarrow$TI & OPT  & Retrieval & 1$\times$A6000, 24Hrs \\ 

GILL~\cite{koh2023generating} & NeurIPS 2023 &  TI$\rightarrow$TI & OPT & Retrieval/SD & 2$\times$A6000, 48Hrs \\ 

SPAE~\cite{yu2023spae} & NeurIPS 2023 & Tokenization & PaLM2/GPT-3.5  & CNN & - \\ 

Emu~\cite{sun2023generative} & ICLR 2024 &  TI$\rightarrow$TI & LLaMa  & SD & -  \\ 

SEED~\cite{ge2023planting} & ICLR 2024 & Tokenization & OPT & SD & 64$\times$V100, 44Hrs \\ 

CM3Leon~\cite{yu2023scaling} & arXiv 2023 & TI$\rightarrow$I, I$\rightarrow$T & CM3Leon & CM3Leon & 64$\times$A100 \\ 

NExT-GPT~\cite{wu2023next} & arXiv 2023 & Any$\rightarrow$Any & Vicuna & SD etc. & - \\ 

DreamLLM~\cite{dong2023dreamllm} & ICLR 2024 & TI$\rightarrow$TI & Vicuna & SD & 128$\times$A800, 17.5Hrs \\ 
MiniGPT-5~\cite{zheng2023minigpt} & arXiv 2023 & TI$\rightarrow$TI & Vicuna  & SD & 4$\times$A6000 \\ 
OpenLEAF~\cite{an2023openleaf} & arXiv 2023 & T$\rightarrow$TI & GPT-4 & SDXL & Training-free \\ 

Mini-DALLE3~\cite{zeqiang2023mini} & arXiv 2023 & TI$\rightarrow$TI & GPT-3.5/GPT-4 etc. & SD-XL/DALLE-3 etc. & Training-free  \\

EasyGen~\cite{zhao2023making} & arXiv 2023 & I$\rightarrow$T,T$\rightarrow$I & FlanT5XL/Vicuna & BiDiffuser & 120$\times$A100 Hrs  \\ 

TEAL~\cite{yang2023teal} & arXiv 2023 & Any$\rightarrow$Any & LLaMa-Adapter  & VQGAN & 8$\times$A100  \\

LLMGA~\cite{xia2023llmga} & arXiv 2024 & T$\rightarrow$I & LLaVA-1.5  & SD & - \\ 

ChatIllusion~\cite{chi2023chatillusion} & arXiv 2023 & TI$\rightarrow$TI & LLaMa-AdapterV2  & SDXL & 4$\times$A6000, 80Hrs \\ 

CoDi-2~\cite{tang2023codi} & CVPR 2024 & Any$\rightarrow$Any & Llama 2 & SD & -  \\

CAFE~\cite{zhou2023customization} & CVPR 2024 & T$\rightarrow$I & Llama 2 & SD & 10000$\times$A100 Hrs  \\ 

StoryGPT-V~\cite{shen2023storygpt} & arXiv 2024 & Story Generation & OPT/Llama2 & Char-LDM & - \\

ELLA~\cite{hu2024ella} & arXiv 2024 & T$\rightarrow$I & Llama 2  & SDXL & 1344$\times$A100 Hrs   \\

Lavi-Bridge~\cite{zhao2024bridging} & arXiv 2024 & T$\rightarrow$I & Llama 2  & SD/PixArt-$\alpha$ &  8$\times$A100, 48Hrs \\

\arrayrulecolor{gray}
\midrule
\arrayrulecolor{black}
\multicolumn{7}{l}{\textbf{\textit{Image layout planning }}\;\;\;\;\;} \vspace{0.05cm}\\
LMD~\cite{lian2023llm} & TMLR 2024 & T$\rightarrow$I & GPT-3.5/GPT-4  & SD & Training-free \\ 
LayoutGPT~\cite{feng2023layoutgpt} & NeurIPS 2023 & T$\rightarrow$I & GPT-3.5/GPT-4/Codex &  GLIGEN/SD & - \\ 
VP-GEN~\cite{cho2023visual} & NeurIPS 2023 & T$\rightarrow$I & Vicuna  & GLIGEN/SD & 4$\times$A6000, 48Hrs \\ 
Control-GPT~\cite{zhang2023controllable} & arXiv 2023 & T$\rightarrow$I & GPT-4  & ControlNet/SD & -  \\  
LayoutLLM-T2I~\cite{qu2023layoutllm} & MM 2023 & T$\rightarrow$I & GPT-3.5  & GLIGEN/SD & -  \\ 
LLM Blueprint~\cite{gani2023llm} & ICLR 2024 & T$\rightarrow$I & GPT-3.5  & LMD & 1$\times$A100  \\ 
SLD~\cite{wu2023self} & CVPR 2024 & T$\rightarrow$I & GPT-4  & DALLE3/SD & Training-free  \\ 
TextDiffuser-2~\cite{chen2023textdiffuser} & arXiv 2023 & T$\rightarrow$I & Vicuna & SD & 8$\times$A100, 168Hrs  \\ 
COLE~\cite{jia2023cole} & arXiv 2023 & T$\rightarrow$I & Llama 2/LLaVA & IF & -  \\ 

\arrayrulecolor{gray}
\midrule
\arrayrulecolor{black}

\multicolumn{7}{l}{\textbf{\textit{Prompt synthesis and refinement }}\;\;\;\;\;} \vspace{0.05cm}\\

SUR-Adapter~\cite{zhong2023adapter} & MM 2023 & T$\rightarrow$I & LLaMa  & SD & - \\ 
ChatGenImage~\cite{yu2023interactive} & arXiv 2023 & Data Synthesis & GPT-3.5  & SD & 1$\times$GTX3090  \\ 
SwitchGPT~\cite{wang2023switchgpt} & arXiv 2023 & T$\rightarrow$TI & Llama 2/GPT-3.5 & SD & 4$\times$A100, 3Hrs \\ 
TIAC~\cite{liao2023text} & arXiv 2023 & T$\rightarrow$I & GPT-3.5  & SD & -  \\  
Idea2Img~\cite{yang2023idea2img} & arXiv 2023 & T$\rightarrow$I & GPT-4V  & IF/SD & Training-free  \\ 
WordArt Designer~\cite{he2023wordart} & EMNLP 2023 & T$\rightarrow$I & GPT-3.5  & ControlNet & 1$\times$V100  \\ 

\arrayrulecolor{gray}
\midrule
\arrayrulecolor{black}
\multicolumn{7}{l}{\textbf{\textit{Image quality evaluation} }\;\;\;\;\;} \vspace{0.05cm}\\

DreamSync~\cite{sun2023dreamsync} & arXiv 2023 & T$\rightarrow$I & PaLM2/TIFA  & SD-XL & -  \\ 

LLMScore~\cite{lu2024llmscore} & NeurIPS 2023 & T$\rightarrow$I & GPT-4  & SD & - \\

\bottomrule
\end{tabular}}
\label{tab:image-generation}
\end{table*}

\begin{table*}[!ht]
    \centering
    \caption{Image-language Datasets that can be adopted for language-based image generation.}
    \scalebox{0.85}{
    \begin{tabular}{llllllll}
    \toprule
        \textbf{Name} & \textbf{Date} & \textbf{Venue} & \textbf{Org} & \textbf{Domain} & \textbf{Source} & \textbf{\#Images} & \textbf{Caption} \\ \midrule

        Im2Text~\cite{ordonez2011im2text} & 12 Dec 2011 & NeurIPS 2011 & SBU & Open & Internet & 1M & Manual \\ 
        Microsoft-COCO~\cite{lin2014microsoft}	& 1 May 2014 &	ECCV 2014 &	Microsoft &	Common Objects	& Internet & 328K &	Manual \\ 
        ALIGN~\cite{jia2021scaling} & 11 Feb 2021 & ICML 2021 & Google & Open & Internet & 1.8B & Manual \\ 
        Conceptual 12M~\cite{changpinyo2021conceptual} & 17 Feb 2021 & CVPR 2021 & Google & Open & Internet & 12M & Manual \\ 
        WIT~\cite{srinivasan2021wit} & 2 Mar 2021 & SIGIR 2021 & Google & Open & Wikipedia & 11.5M & Manual \\
        LAION-400M~\cite{schuhmann2021laion} & 3 Nov 2021 & NeurIPS 2021 & LAION & Open & Internet & 400M & Manual \\ 
        LAION-FACE~\cite{zheng2022general} & 6 Dec 2021 & CVPR 2022 & Microsoft & Face & LAION & 20M & Manual \\ 
        M3W~\cite{alayrac2022flamingo} & 29 Apr 2022 & NeurIPS 2022 & Deepmind & Interleave & Internet & 43M & Manual \\ 
        LAION-COCO~\cite{laioncoco} & 15 Sep 2022 & - & LAION & Open & LAION & 600M & Synthetic \\ 
        LAION-5B~\cite{schuhmann2022laion} & 16 Oct 2022 & NeurIPS 2022 & LAION & Open & Internet & 5B & Manual \\ 
        Coyo-700M~\cite{coyo} & 31 Aug 2022 & - & Kakao Brain & Open & Internet & 700M & Manual \\ 
        KOSMOS-1~\cite{huang2024language} & 27 Feb 2023 & NeurIPS 2023 & Microsoft & Interleave & Internet & 355M & Manual \\ 
        Multimodal C4~\cite{zhu2024multimodal} & 14 Apr 2023 & NeurIPS 2023 & UCSB & Interleave & Internet & 571M & Manual \\ 
        LLaVA-instruct~\cite{liu2024visual} & 17 Apr 2023 & NeurIPS 2023 & UWM & Instruction & COCO & 150k & Synthetic \\ 
        DATACOMP~\cite{gadre2024datacomp} & 27 Apr 2023 & NeurIPS 2023 & DATACOMP & Open & Internet & 12.8B & Manual \\ 
        MARIO-10M~\cite{chen2024textdiffuser} & 19 May 2023 & NeurIPS 2023 & Microsoft & Text within image & LAION, TMDB, OpenLibrary & 10M & Manual \\ 
        LAION-Glyph~\cite{yang2024glyphcontrol} & 29 May 2023 &	NeurIPS 2023 &	Microsoft &	Text within image &	LAION &	10M	& Manual \\
        MIMIC-IT~\cite{li2023mimic} & 8 Jun 2023 & arXiv 2023 & NTU & Interleave & Internet & 2.8M & Synthetic \\ 
    \bottomrule
    \end{tabular}
    }
    \label{tab:image-language-dataset}
\end{table*}

\begin{figure*}[h]
    \centering
    \includegraphics[width=\linewidth]{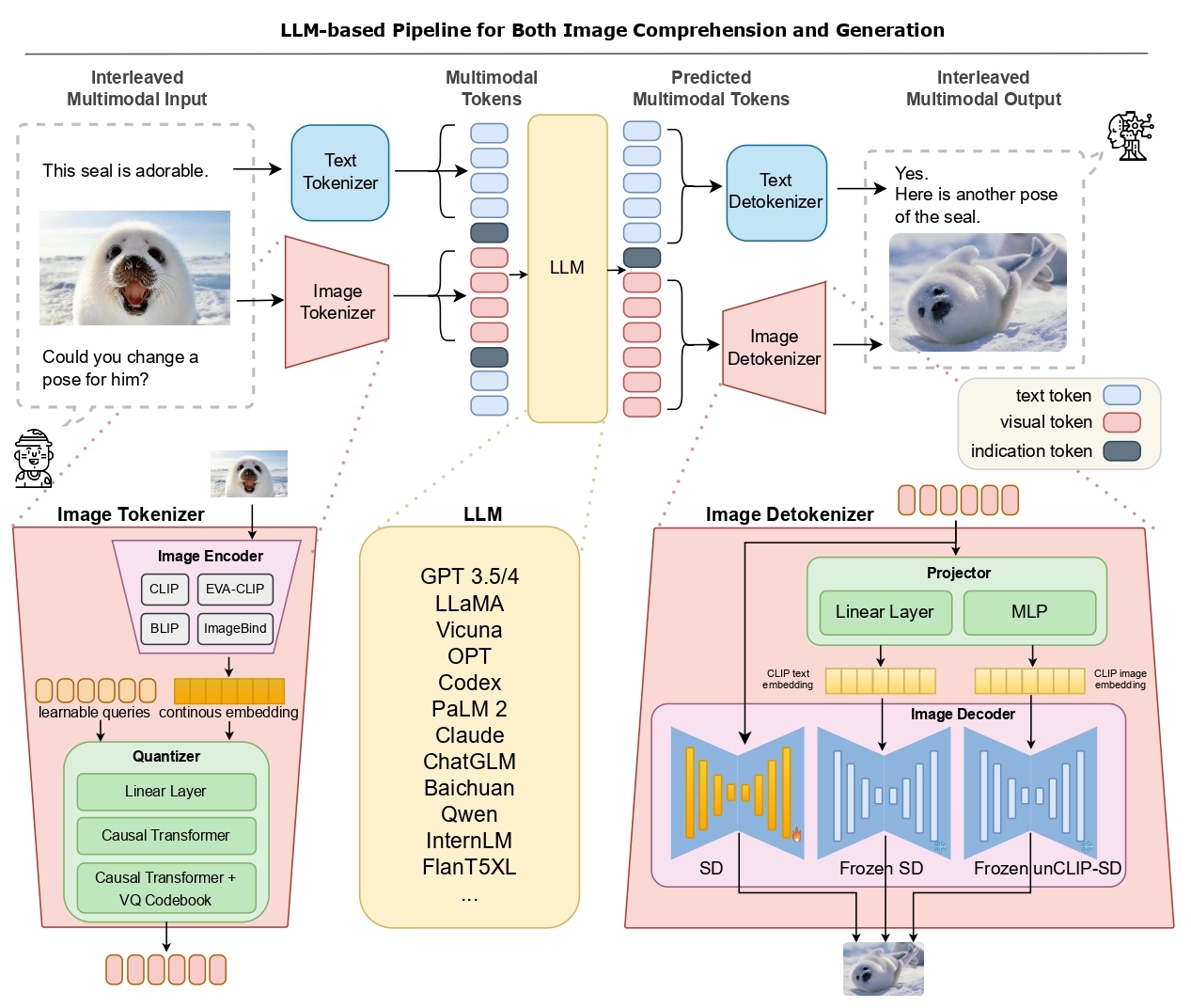}
    \caption{A generic pipeline of integrating image comprehension and generation ability on LLMs~\cite{sun2023emu2, ge2023planting, dong2023dreamllm, ge2023making}. 
    During inference time, users can input interleaved multimodal data (\textit{e.g.}, text and images). 
    The image tokenizer processes the information into image tokens and feeds them into the LLM.
    LLM outputs image tokens and then decodes them into textual responses and images.
    }
    \label{fig:imagepipe}
\end{figure*}

\begin{figure}[h]
    \centering
    \includegraphics[width=\linewidth]{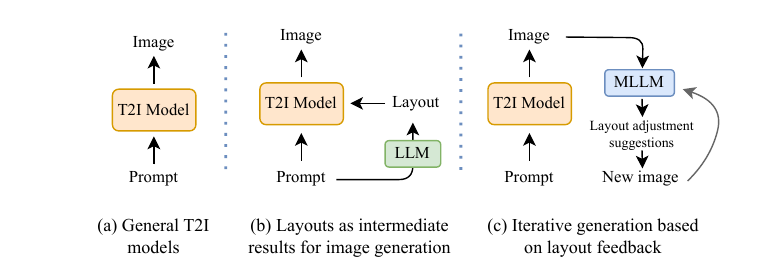}
    \caption{Pipeline comparisons of (a) standard text-to-image (T2I)~\cite{saharia2022photorealistic,ldm}, (b) T2I with LLMs as layout planners~\cite{feng2023layoutgpt,qu2023layoutllm,chen2023textdiffuser2,lian2023llm,cho2023visual,zhang2023controllable,gani2023llm}, and (c) T2I with LLMs for layout suggestions~\cite{jia2023cole,wu2023self}.
    }
    \label{fig:image_layout_planner}
\end{figure}

\section{Image Generation and Editing}
\label{sec: image}
\subsection{Image Generation}
Image generation has long been a fundamental task in the field of computer vision, playing a vital role in various applications such as digital art, entertainment, education, and communication~\cite{vetter1997linear,cohen1993radiosity,kirk1991unbiased}. 
In the beginning stage of image generation, the generated content is limited to a specific category, such as human faces, cats, or buildings.
Recent advancements in image generation have been particularly notable due to the introduction of text guidance and open-domain generation. 
Most recently, the power of LLMs has elevated image generation to a new level, enabling interactive or interleaved generation.
In Fig.~\ref{fig:image-history}, we summarize the history and development trajectory of image generation in detail. A curated list of recent image generation methods is shown in Table~\ref{tab:image-generation}. We also list representative datasets for image generation in Table~\ref{tab:image-language-dataset}.
These works allow for the creation of images that closely align with textual prompts, providing a powerful tool for visualizing ideas with creativity.

\subsubsection{Text-guided Image Generation with CLIP}
Previously, the adoption of image-text alignment models, such as CLIP~\cite{radford2021learning}, has played a crucial role in the development of text-guided image generation~\cite{dalle2hierar,chen2023textdiffuser,frans2022clipdraw,ldm,gu2022vector,wang2022clipgen}. The alignment capabilities of CLIP text encoder ensure that the generated image aligns with the given textual prompt, resulting in images that accurately match the intended descriptions, including desired objects, scenes, or attributes. Given the significant strides made by  CLIP in generating realistic images, it is natural to question: \textit{will more powerful LLMs further benefit the domain of image generation?}

It is worth mentioning that, the application of LLMs for the image domain has been extensively investigated, especially for image understanding. LLMs can effectively serve as unified processors of visual tokens and language tokens~\cite{liu2023visual,li2023blip,lv2023kosmos,ye2023mplug,ye2023mplug2,zhu2023minigpt,chen2023minigpt2,zhang2023llama,gao2023llama,dai2305instructblip,wang2023visionllm,bai2023qwen,wu2023next,wang2023cogvlm,li2023llama,jin2023unified,chen2023internvl,huang2023language}, tool coordinators~\cite{wu2023visual,shen2023hugginggpt,liu2023internchat,yang2023mm}, or analysts of upstream visual model outputs~\cite{wang2023chatcad,yang2022empirical,gu2023anomalygpt}. Inspired by these works, numerous works further take advantage of LLMs for image generation, and the milestone works are shown in Fig.~\ref{fig:image-history}. In the following, we will introduce the progress achieved in the task of image generation after the emergence of LLMs.

\subsubsection{Text-guided Image Generation with LLMs}
As demonstrated in Fig.~\ref{fig:imagepipe}, MLLMs have emerged as a transformative extension of LLMs, addressing the inherent limitations of LLMs in handling visual content. 
While LLMs excel in flexible text-based interactions, they are confined to textual inputs and outputs. The introduction of MLLMs stems from the necessity to bridge this gap and enable language models to comprehend and generate images. 
MLLMs offer a two-fold advantage: firstly, they serve as a unified interface for understanding and generating both textual and visual information, providing users with a seamless integration of language and images. 
Secondly, MLLMs introduce interactive generation capabilities, allowing users to send commands to modify image content iteratively. 
This interactive process empowers users with greater control over the image generation process, enhancing the overall user experience and controllability towards user-desired content.

Specifically, CM3Leon~\cite{yu2023scaling} is an autoregressive MLLMs designed for the simultaneous generation of textual and image outputs. Operating within a decoder-only, retrieval-augmented, token-based framework, CM3Leon offers a unique approach to multimodal language processing.
DreamLLM~\cite{dong2023dreamllm} presents the first MLLM capable of generating free-form interleaved content, supporting multi-round dialogue, and achieving remarkable results in image captioning and video question answering (VQA) without the need for fine-tuning. The entire framework is trained on interleaved multimodal contents in a truly end-to-end manner.
SEED-LLaMA~\cite{ge2023making}, similar to DreamLLM, enables LLMs to comprehend multimodal instructions and supports multi-turn in-context image and text generation. Notably, SEED-LLaMA emphasizes the design of the Image Tokenizer, proposing two crucial design principles for its functionality.
MiniGPT-5~\cite{zheng2023minigpt} introduces visual tokens (referred to as "voken") to traditional LLMs, enabling them to generate images. A two-stage training pipeline is proposed, involving an unimodal alignment stage and a multimodal learning stage, allowing LLMs to generate both text and images organically.
OpenLeaf~\cite{an2023openleaf} utilizes prompting of LLMs to generate interleaved textual and visual data, producing entity and style-consistent images and text. It supports various tasks such as how-to question answering, story-telling, graphical story rewriting, and webpage/poster generation.
EasyGen~\cite{zhao2023making} leverages a bidirectional conditional diffusion model, BiDiffuser, to endow LLMs with multimodal understanding and generation capabilities. Unlike previous CLIP-based approaches, EasyGen generates images based on this model.
TEAL~\cite{yang2023teal} utilizes existing tokenizers for different modalities and transforms the obtained tokens into a joint embedding space, enabling frozen LLMs to understand and generate across various modalities, including text, image, and audio.
ChatIllusion~\cite{chi2023chatillusion} introduces Genadapter and LLaMa-AdapterV2 to bridge the hidden embedding space of SD XL and enable LLMs to understand visual instructions and generate interleaved images and text, supporting image generation, editing, and storytelling.
Emu2~\cite{sun2023emu2} emphasizes the in-context learning capability of MLLMs, showcasing improved performance through scaling up the model and unified autoregressive training. It supports tasks such as visual prompting and object-grounded generation, achieving state-of-the-art results in question-answering benchmarks and open-ended subject-driven generation after instruction tuning.
ELLA~\cite{hu2024ella} and Lavi-Bridge~\cite{zhao2024bridging} incorporate large language models into the T$\rightarrow$I generation architecture by training several lightweight adapters. 
LLMGA~\cite{xia2023llmga} utilizes LLaVA to simultaneously encode images and instruction, enabling manipulating images based on stable diffusion.
StoryGPT-V~\cite{shen2023storygpt} takes advantage of LLM for coherent story script generation.

\subsubsection{Image Layout Planning via LLMs}

Despite the rapid development of T$\rightarrow$I generation, there remain several challenging issues that have yet to be fully addressed, including text rendering, spatial relationships, and quantity representation. Under this circumstance, some methods seek to utilize LLMs for layout planning and subsequently generate images based on the obtained layouts, as shown in Fig.~\ref{fig:image_layout_planner}. LayoutGPT~\cite{feng2023layoutgpt} leverages the inherent reasoning capabilities in LLMs to facilitate layout generation through contextual demonstrations. It utilizes GPT3.5/4 to transform user prompts to CSS-style output layouts, in which the position of each object is specified. LMD~\cite{lian2023llm} improves T$\rightarrow$I diffusion models by enhancing prompt understanding capabilities. It utilizes a two-stage approach, leveraging a pre-trained language model to generate scene layouts and guide image generation. VP-GEN~\cite{cho2023visual} breaks down the T2I task into object/count generation, layout generation, and image generation steps. By leveraging GPT-3.5-Turbo fine-tuned on text-layout pairs, VPGEN achieves better spatial control than end-to-end models. Control-GPT~\cite{zhang2023controllable} takes advantage of GPT-4 to output TikZ code to construct the sketch layouts according to the text descriptions. LayoutLLM-T2I~\cite{qu2023layoutllm} utilizes ChatGPT to induce the layout based on the user prompt. A prompt encoder module is then employed to model the text prompt, relation triplets, and the induced layout separately. To efficiently integrate the layout information, a Layout-aware Spatial Transformer based on the UNet is introduced. LLM Blueprint~\cite{gani2023llm} utilizes ChatGPT to produce detailed object descriptions, bounding box layouts, as well as background prompts. Subsequently, iterative refinement operations are conducted to remedy regional errors based on the layouts. SLD~\cite{wu2023self} improves T$\rightarrow$I generation by iteratively generating images from input prompts and correcting mistakes using an LLM-based layout planner. In particular, the layout planner can add, delete, or resize object boxes to help T2I models produce more accurate images. TextDiffuser-2~\cite{chen2023textdiffuser2} employs Vicuna-7B-1.5 for layout planning, generating the position and content of the text to be rendered based on user-provided prompts. COLE~\cite{jia2023cole} harnesses large language models to transform user prompts into detailed JSON files. These files encompass specifications, such as the content, position, and style, for the text to be added.

\subsubsection{Prompt Synthesis and Refinement via LLMs}
LLMs can be treated as a huge knowledge base. Some methods~\cite{yu2023interactive,yang2023idea2img,liao2023text,he2023wordart} have explored LLMs to synthesize or optimize prompts, guiding T$\rightarrow$I (T2I) models to generate images with rich and detailed content. For example, ChatGenImage~\cite{yu2023interactive} utilizes ChatGPT to generate prompts, directing AIGC models in generating preliminary images. Subsequently, it iteratively refines these prompts by incorporating automatically generated detailed annotations as local constraint prompts, resulting in the production of diverse and intricate scenes. Inspired by the three-layer artwork theory, TIAC~\cite{liao2023text} and WordArt Designer~\cite{he2023wordart} use LLMs to translate abstract concepts into semantically relevant physical objects, making it easier for downstream T$\rightarrow$I models. Idea2Img~\cite{yang2023idea2img} employs a Multimodal LLM to evaluate the generated images of T$\rightarrow$I models. Subsequently, based on the obtained feedback, the framework iteratively refines the initial prompts to generate satisfactory results.
DiffusionGPT~\cite{qin2024diffusiongpt} leverages LLMs to refine prompts for image generation. By parsing diverse prompts and utilizing domain-specific Trees-of-Thought, the model selects the most suitable generative model to generate high-quality images.
RPG~\cite{yang2024mastering} is a training-free framework for T$\rightarrow$I generation. It leverages multimodal LLMs to refine the original prompt, decomposing complex prompts into subregion tasks, and achieves superior performance in object composition and text-image alignment.
SUR-adapter~\cite{zhong2023adapter} leverages LLMs to improve its semantic understanding and reasoning capabilities, enabling it to create better textual semantic representations for T$\rightarrow$I generation.
SwitchGPT~\cite{wang2023switchgpt} introduces an innovative framework that enables traditional LLMs, like GPT, to interpret
the underlying intent of the given instruction, thus producing a more suitable response
non-text outputs.

\subsubsection{Image Quality Evaluation via LLMs}
A few works focus on using large language models to evaluate the quality of generated images. 
For example, DreamSync~\cite{sun2023dreamsync} employs two vision language models (VLMs) to evaluate the generated results, and select the best-generated image: one for text alignment and another for aesthetic quality. LoRA~\cite{hu2022lora} is then used to iteratively fine-tune the T2I model towards the selected best generations. 
LLMScore~\cite{lu2024llmscore} converts images into image-level and object-level visual descriptions. Subsequently, a set of instructions is given to the LLMs to check how closely between the images and the descriptions. Finally, a score is generated with reasons.

\subsection{Image Editing}
Image editing is a closely related task with generation and thus receives remarkable progress with the development of image generation models. A curated list of recent image editing methods is shown in Table~\ref{tab:image-edit}.
\subsubsection{Image Editing with CLIP/T5}
CLIP model enables language-based image editing~\cite{meng2021sdedit, kim2022diffusionclip, couairon2022diffedit, hertz2022prompt, mokady2023null, kawar2023imagic, tumanyan2023plug, morita2023interactive, zhang2023sine, parmar2023zero, goel2023pair, cao2023masactrl, nguyen2023visual, mirzaei2023watch}. 
PAIR-Diffusion~\cite{goel2023pair} identifies structure and appearance as the two most intuitive aspects of image editing. Thus, the PAIR-Diffusion is trained using structure and appearance information explicitly extracted from the training images, which enables editing of the structure and appearance separately during inference. 
Instead of relying on the heavy training process of diffusion models on large-scale datasets, another line of work~\cite{kawar2023imagic, zhang2023sine} slightly tunes the pre-trained diffusion models to edit the target images. Imagic~\cite{kawar2023imagic} proposes a generic model for text-based image editing. Notably, Imagic is free of source prompts, which is achieved by optimizing the target prompt embedding. After that, the entire diffusion model is fine-tuned for better reconstruction performance. The editing ability is achieved by interpolating the optimized prompt embeddings and the target prompt embeddings. 
There also exist many tuning-free methods for text-guided image editing~\cite{meng2021sdedit, kim2022diffusionclip, couairon2022diffedit, hertz2022prompt, mokady2023null, tumanyan2023plug, morita2023interactive, parmar2023zero, cao2023masactrl, nguyen2023visual, mirzaei2023watch}. 
SDEdit~\cite{meng2021sdedit} can generate realistic images from user inputs such as strokes, sketches, or masks, as well as edit existing images with text instructions. It works by first adding noise to the target image and then gradually denoising it with text instructions. The denoising process is guided by a diffusion model generative prior, which is trained on a large-scale image dataset. SDEdit outperforms state-of-the-art GAN-based methods on multiple image synthesis and editing tasks, according to a human perception study. 
Other tuning-free methods~\cite{parmar2023zero, cao2023masactrl, hertz2022prompt, tumanyan2023plug} achieve text-guided editing through the regularization or manipulation of latents, cross-attention maps, or UNet features. 
However, most text-guided image editing works rely on the CLIP model, whose capability limits the editing to simple text prompts and cannot understand complex human instructions. 

\begin{table*}[t]
\centering
\caption{An overview of language-based image editing methods based on CLIP and LLMs. 
We summarize the involved LLMs and generative models, as well as whether the method requires training or not. }
\scalebox{1}{
\begin{tabular}{lccccc}
 \toprule
\textbf{Method} & \textbf{Venue}  & \textbf{LLM}  & \textbf{Generative Model}  & \textbf{Training} \\
\midrule
\multicolumn{6}{l}{\textbf{\textit{CLIP for image editing}}\;\;\;\;\;} \vspace{0.05cm}\\
SDEdit~\cite{meng2021sdedit} & ICLR 2022 & - &  DDPM  & \xmark \\
DiffusionCLIP~\cite{kim2022diffusionclip} & CVPR 2022 & - &  DDPM  & \xmark \\
P2P~\cite{hertz2022prompt} & ICLR 2023 & - &  Imagen\&SD  & \xmark \\
NTI~\cite{mokady2023null} & ICLR 2023 & - &  SD  & \xmark \\
Imagic~\cite{kawar2023imagic} & CVPR 2023 & - &  Imagen  & \cmark \\
PaP~\cite{tumanyan2023plug} & CVPR 2023 & - &  SD  & \xmark \\
SINE~\cite{zhang2023sine} & CVPR 2023 & - &  SD  & \cmark \\
pix2pix-zero~\cite{parmar2023zero} & SIGGRAPH 2023 & - &  SD  & \xmark \\
PAIR-Diffusion~\cite{goel2023pair} & arXiv 2023 & - &  PAIR-Diffusion  & \cmark \\
MasaCtrl~\cite{cao2023masactrl} & SIGGRAPH 2023 & - &  SD  & \xmark \\
Dragondiffusion~\cite{mou2023dragondiffusion} & ICLR 2024 & - &  SD  & \cmark \\
DiffEditor~\cite{mou2024diffeditor} & CVPR 2024 & - &  SD  & \cmark \\

\midrule

\multicolumn{6}{l}{\textbf{\textit{LLMs for image editing}}\;\;\;\;\;} \vspace{0.05cm}\\

InstructPix2Pix~\cite{brooks2022instructpix2pix} & CVPR 2023 & GPT-3 &   SD  & \cmark \\
VisualChatGPT~\cite{wu2023visual} & arXiv 2023 & GPT-3 &   SD  & \cmark \\
CHATEDIT~\cite{cui2023chatedit} & EMNLP 2023 & GPT-3 &   StyleGAN  & \cmark \\
MGIE~\cite{fu2023guiding} & ICLR 2024 & LLaVA-7B &   SD  & \cmark \\
Emu Edit~\cite{sheynin2023emu} & arXiv 2023 & Llama 2-70B &   Emu  & \cmark \\
SLD~\cite{wu2023self} & CVPR 2024 & GPT-4 &   DALL-E 3  & \cmark \\
SmartEdit~\cite{huang2023smartedit} & CVPR 2024 & LLaVA-7B/13B &   InstructDiffusion~\cite{geng2023instructdiffusion}  & \cmark \\
\arrayrulecolor{gray}

\bottomrule
\end{tabular}}
\label{tab:image-edit}
\end{table*}

\begin{table*}[t]
\centering
\caption{An overview of video editing methods based on CLIP and LLMs. We summarize the involved LLMs and generative models, as well as whether the method requires training or not. }
\scalebox{1}{
\begin{tabular}{lccccc}
 \toprule
\textbf{Method} & \textbf{Venue}  & \textbf{LLM}  & \textbf{Generative Model}  & \textbf{Training} \\

\midrule

\multicolumn{6}{l}{\textbf{\textit{CLIP for video editing}}\;\;\;\;\;} \vspace{0.05cm}\\
Tune-A-Video~\cite{wu2023tune} & ICCV 2023 & - &   SD  & \cmark \\
Dreamix~\cite{molad2023dreamix} & arXiv 2023 & - &   Imagen-video  & \cmark \\
Video-P2P~\cite{liu2023video} & arXiv 2023 & - &   SD  & \cmark \\
FateZero~\cite{qi2023fatezero} & ICCV 2023 & - &   SD  & \xmark \\
Pix2Video~\cite{ceylan2023pix2video} & ICCV 2023 & - &   SD  & \xmark\\
StableVideo~\cite{chai2023stablevideo} & ICCV 2023 & - &   SD  & \xmark \\
Rerender-A-Video~\cite{yang2023rerender} & SIGGRAPH Asia 2023 & - &   SD  & \cmark \\
TokenFlow~\cite{geyer2023tokenflow} & ICLR 2024 & - &   SD  & \xmark \\
CoDeF~\cite{ouyang2023codef} & CVPR 2024 & - &   SD  & \cmark \\
MagicEdit~\cite{liew2023magicedit} & arXiv 2023 & - &   SD  & \cmark \\
MagicStick~\cite{ma2023magicstick} & arXiv 2023 & - &   SD  & \cmark \\
\midrule

\arrayrulecolor{black}
\multicolumn{6}{l}{\textbf{\textit{LLMs for language-based video editing }}\;\;\;\;\;} \vspace{0.05cm}\\
InstructV2V~\cite{cheng2023consistent} & ICLR 2024 & GPT-3 &  SD  & \xmark \\
InstructVid2Vid~\cite{qin2023instructvid2vid} & arXiv 2023 & GPT-3 &  SD  & \xmark \\

\bottomrule
\end{tabular}}
\label{tab:video-edit}
\end{table*}

\subsubsection{Image Editing with LLMs}
LLMs provide powerful chat-based or interactive editing capabilities for image editing~\cite{brooks2023instructpix2pix, wu2023visual, cui2023chatedit, fu2023guiding, sheynin2023emu, wu2023self}. 

InstructPix2pix~\cite{brooks2023instructpix2pix} proposes to use LLMs to construct the data tuples (original image, prompt, target image) to train a model that learns to edit the images following the editing prompt. The model is based on a conditional diffusion model that can process arbitrarily interleaved image and text inputs and produce coherent image (and text) outputs. To generate the data tuples, the authors leverage the knowledge of two large pretrained models: a language model (GPT-3) and a T$\rightarrow$I model (Stable Diffusion). The language model generates the editing instructions and the textual descriptions of the edited images, while the T$\rightarrow$I model renders the edited images based on the textual descriptions. The authors also introduce a mapping network that translates the hidden representations of the language model into the embedding space of the visual models, enabling the model to use the strong text representations of the LLM for visual outputs. 

CHATEDIT~\cite{cui2023chatedit} further leverages LLMs to contribute an interactive facial image editing system via dialogue. Specifically, CHATEDIT split the dialogue-based editing problem into (1) user edit request tracking, (2) image editing, and (3) response generation subtasks. The user edit request tracking module is responsible for extracting the user’s editing intentions from the dialogue history and updating them dynamically. The image editing module is based on a conditional diffusion model that can process both image and text inputs and outputs, and perform various editing operations such as changing hair color, adding glasses, or removing wrinkles. The response generation module is designed to generate natural and informative responses that reflect the editing results and guide the user to the next step. CHATEDIT is evaluated on a novel benchmark dataset proposed by the authors, which contains multi-turn dialogues and corresponding facial images annotated with user edit requests. 

MGIE~\cite{fu2023guiding} investigates the MLLMs to do the image editing task. The proposed MGIE can learn to convert expressive human instructions to editing guidance. The editing model is also trained to follow the editing guidance in an end-to-end way. The effectiveness of the MGIE is verified on photoshop-style manipulations, global photo optimization and local editing. SmartEdit~\cite{huang2023smartedit} is another recent work that utilizes MLLMs for complex instruction-based image editing. SmartEdit analyzes the performance of instruction-based image editing models under complex instructions and proposes a Bidirectional Interaction Module to make the image feature output by pre-trained image encoder and the output feature of LLaVA. They also fine-tune the pre-trained diffusion model to enhance the model's perception and reasoning capability. 

Emu edit~\cite{sheynin2023emu} trained the image editing model in a multi-task way. The tasks include region-based editing, free-form editing and other computer vision tasks, where all tasks are formulated into generative tasks. Emu edit leverages LLMs for instruction generation. Concretely, the authors provide the LLMs with a task description, a few task-specific exemplars, and a real image caption. Then, the LLM is expected to output an editing instruction, an output caption for an ideal output image, and which objects should be updated or added to the original image. 

Different from the above works that utilize LLMs to provide editing instructions, SLD~\cite{wu2023self} uses LLMs to correct the incorrect generation to enable object-level image editing. 

\subsection{Image-language datasets}

Image-language datasets play a crucial role in the training of T$\rightarrow$I models, providing the foundational data necessary for these models to learn how to generate accurate and relevant visual content from textual descriptions. Over ten years ago, the IM2Text~\cite{ordonez2011im2text} project gathered a huge collection of photos by searching through Flickr, a popular photo-sharing website. They sifted through a massive amount of data and carefully picked out one million images that had clear and directly related captions. MS-COCO~\cite{lin2014microsoft} collected images depicting complex everyday scenes with common objects in their natural settings. The researchers included five written captions to provide descriptive context. These captions offer a richer understanding of the scene and the objects within it. In recent years, the academic community has witnessed a proliferation of large-scale image-text datasets. Typically, researchers curate these datasets by crawling the internet. For example, LAION-5B~\cite{schuhmann2022laion} is a massive dataset that was collected by searching the internet for image-text pairs. Using the CLIP model to filter the results, the researchers ensured that the text was relevant to the images. This process resulted in a dataset containing 5.85 billion image-text pairs. Additionally, some researchers are digging into the LAION-5B to find specific types of content. For example, the Mario-10M dataset focuses on pulling out parts of the dataset where the images have text in them to study further. Meanwhile, LAION-FACE~\cite{zheng2022general} is all about images with faces. These specialized subsets help researchers focus on particular areas within the massive collection of image-text pairs. Further, to help image generators follow instructions during a conversation, the LLaVA Visual Instruct 150K dataset~\cite{liu2023visual} comprises a collection of multimodal data designed for instruction-following tasks, generated via a GPT model.

\begin{table*}[t]
\centering
\caption{
    Overview of existing methods leveraging LLMs for language-based video generation.
    We divide these methods into four categories: 
    multimodal LLMs for video generation, video layout planning, and temporal prompt generation.
    In each method, we summarize the input-output of the task, the involved LLM, and the generative model.
    In the "Task`` column, the "T`` and "V`` are the abbreviations of "text`` and "video``, respectively, while "Any`` represents the universal generation supporting text, image, video, and audio modalities.
    Tokenization is the task of converting video into discrete video tokens, which can be viewed as a submodule of some video generation pipelines.
}
\scalebox{1}{
\begin{tabular}{lccccc}
 \toprule
\textbf{Method} & \textbf{Venue}  & \textbf{Task} & \textbf{LLM} & \textbf{Generative Model} \\
\midrule
\multicolumn{6}{l}{\textbf{\textit{Multimodal LLMs for video generation}}\;\;\;\;\;} \vspace{0.05cm}\\

VideoPoet~\cite{kondratyuk2023videopoet} & arXiv 2023 & Any$\rightarrow$V & VideoPoet & VideoPoet \\
MAGVIT-v2~\cite{yu2023language} & ICLR 2024 & Tokenization & BERT & BERT \\
Video-LaVIT~\cite{yu2023language} & arXiv 2024 & TIV$\rightarrow$TIV & Llama 2-7B & SVD img2vid-xt  \\
\arrayrulecolor{gray}
\midrule
\arrayrulecolor{black}
\multicolumn{6}{l}{\textbf{\textit{Video layout planning }}\;\;\;\;\;} \vspace{0.05cm}\\

Dysen-VDM~\cite{fei2023empowering} & CVPR 2024 & T$\rightarrow$V & GPT-4 & Text2Video-Zero \\
VideoDirectorGPT~\cite{lin2023videodirectorgpt} & arXiv 2023 & TI$\rightarrow$V & GPT-4 & LayoutVid \\
LVD~\cite{lian2023llm} & ICLR 2024 & T$\rightarrow$V & GPT-3.5/GPT-4 & DSL-grounded generator \\
GPT4MOTION~\cite{lv2023gpt4motion} & arXiv 2023 & T$\rightarrow$V & GPT-4 & SDXL / ControlNet \\
FlowZero~\cite{lu2023flowzero} & arXiv 2023 & T$\rightarrow$V & GPT-4 & Gligen \\

\arrayrulecolor{gray}
\midrule
\arrayrulecolor{black}

\multicolumn{6}{l}{\textbf{\textit{Temporal prompt generation }}\;\;\;\;\;} \vspace{0.05cm}\\

DirecT2V~\cite{hong2023large} & arXiv 2023 & T$\rightarrow$V & GPT-4 & Text2Video-Zero \\
Free-Bloom~\cite{huang2023free} & NeurIPS 2023 & T$\rightarrow$V & GPT-3.5 & LDM \\
InterControl~\cite{wang2023intercontrol} & arXiv 2023 & T$\rightarrow$V & GPT-4 & HMDM \\
PRO-Motion~\cite{liu2023plan} & arXiv 2023 & T$\rightarrow$V & GPT-3.5 & Posture-Diffuser \\
VideoDrafter~\cite{long2024videodrafter} & arXiv 2024 & T$\rightarrow$V & ChatGLM-6B & SD-XL \\

\bottomrule
\end{tabular}}
\label{tab:video_generation_llm}
\end{table*}

\section{Video Generation and Editing}
\label{sec: video}
\subsection{Video Generation}
While video understanding has been thoroughly investigated~\cite{wu2021star,liu2021tam,zhao2022towards,bain2021frozen,yang2022zero,yang2022learning,lin2022swinbert,lin2019tsm,bertasius2021space,wu2019long,zhang2023video,chen2023video}, the past two years have witnessed the rapid development or video generation.
Particularly in the field of text-based video generation, numerous works have achieved remarkable results.
We list the milestone works in Fig.~\ref{fig:video_timeline} including both CLIP/T5-based and LLM-based approaches, and summarize the key technical components in Table~\ref{tab:video_generation_llm} and commonly-used video-language datasets in Table~\ref{tab:video_dataset}.

\subsubsection{Text to Video Generation with CLIP}

Based on the type of generative models, there are two main paradigms: one is based on diffusion models, and the other is based on autoregressive models built with transformer architectures and discrete codebooks, trained with the next-token prediction loss.
Diffusion models have become a mainstream paradigm due to their ease of training. Within the diffusion framework, there are pixel-level video diffusion models~\cite{vdm, singer2022make-a-video, ho2022imagen-video} and latent-level video diffusion models~\cite{he2022lvdm, zhou2022magicvideo, blattmann2023align, animate-a-story, follow-your-pose, make-your-video,freenoise, follow-your-click}.
Pixel-level approaches exhibit better text alignment but require substantial computational resources. 
On the other hand, latent-level models are more efficient as they reduce redundancy in video data.
Different works in video generation have different focuses. Some emphasize photorealistic or high-definition outputs, aiming to improve the quality of generated videos. Others focus on controllable generation, such as image-to-video approaches that enable local control over motion regions, trajectory control for object and camera movement directions, and the use of sketch, depth, and pose to control structure. Some works concentrate on generating longer videos or exploring better network architectures.

\subsubsection{Text to Video Generation with LLMs}
Recently, several works have also taken advantage of multimodal LLMs~\cite{kondratyuk2023videopoet, yu2023language} for the task of video generation. 
For example, VideoPoet~\cite{kondratyuk2023videopoet} leverages a pre-trained auto-regressive transformer model to handle multimodal data for synthesizing videos with temporal consistency and high-motion fidelity. It adapts LLMs training techniques, allowing for task-specific video generation, including tasks such as text-to-video and image-to-video conversion. 
Another work, MAGVIT-v2~\cite{yu2023language}, explores the video tokenization technique of MLLMs. 
It transforms visual inputs into discrete tokens, enhancing the performance of image and video generation tasks. 
It outperforms diffusion models on benchmarks such as ImageNet and Kinetics, provides video compression on par with advanced codecs, and improves action recognition through effective representation learning.

\subsubsection{Video Layout Planning via LLMs}
Numerous studies have validated the efficacy of LLMs in generating image layouts, and similarly, recent research has sought to explore the potential of LLMs in crafting video layouts. For instance, some methods employ LLMs to sequentially generate bounding boxes for objects in each frame to assist in the video generation process.
VideoDirectorGPT~\cite{lin2023videodirectorgpt} utilizes a language model to generate a video plan that includes object-bound boxes for each scene. These bounding boxes provide spatial coordinates for the entities, which are used to maintain object consistency and precise layout control throughout the video generation process.
LLM-grounded Video Diffusion (LVD)~\cite{lian2023llm} enhances video generation by first using an LLM to create detailed scene layouts from textual prompts, capturing complex motions. These layouts then guide a diffusion model through adjusted attention maps to produce videos that accurately reflect the prompted actions and dynamics, improving upon existing video generation methods.
FlowZero~\cite{lu2023flowzero} utilizes LLMs to generate a dynamic scene syntax that includes object bounding boxes, which are critical for defining object positions and movements within each frame. These bounding boxes guide the diffusion model to ensure accurate object placement and coherent motion throughout the video. 
Another line of works attempts to utilize layouts beyond conventional bounding boxes.
Dysen-VDM~\cite{fei2023empowering} enhances the quality of generated videos by employing a dynamic scene graph (DSG). The DSG is used to capture and organize the temporal dynamics of actions described in the text, which are then enriched with details and integrated into a diffusion model to produce more dynamic and realistic videos. 
GPT4Motion~\cite{lv2023gpt4motion} uses GPT-4 to script physical scenes in Blender. The simulated scenes are transformed into intermediate representations, such as depth maps, to serve as layout conditions. Then these conditions are fed to Stable Diffusion to produce the final video, ensuring motion consistency and efficiency in scenarios like object interactions and fluid dynamics.

\subsubsection{Temporal Prompt Generation via LLMs}

In contrast to image generation, video generation requires more elaborate and dense descriptions. The refinement and expansion of prompts can be facilitated by leveraging the capabilities of LLMs. For instance,
DirecT2V~\cite{hong2023large} improves narrative consistency and scene composition from user prompts. It employs instruction-tuned large language models to decompose a single user prompt into detailed frame-by-frame descriptions. These descriptions guide the generation of each video frame, allowing for the seamless integration of time-varying elements and coherent storytelling.
Free-Bloom~\cite{huang2023free} leverages large language models LLMs to create a sequence of semantically coherent prompts, guiding the video's narrative flow. Pre-trained latent diffusion models (LDMs) are then used as animators to generate high-fidelity frames that visually represent the evolving semantic content, like the process of a flower blooming.
InterControl~\cite{wang2023intercontrol} uses an LLM planner to convert textual interaction descriptions into detailed contact plans, improving the generation quality of motion videos.
PRO-MOTION~\cite{liu2023plan} uses LLMs to create a sequence of scripts detailing the key postures needed for the target motion. These scripts are based on simple templates, which are different from natural languages and are designed to comprehensively describe all possible postures, thereby simplifying the process for the subsequent generation process.
VideoDrafter~\cite{long2024videodrafter} utilizes LLMs to convert an input prompt into a detailed multi-scene script. This script capitalizes on the logical knowledge of LLMs to ensure the scenes make sense in sequence.

\begin{figure*}[h]
    \centering
    \includegraphics[width=\linewidth]{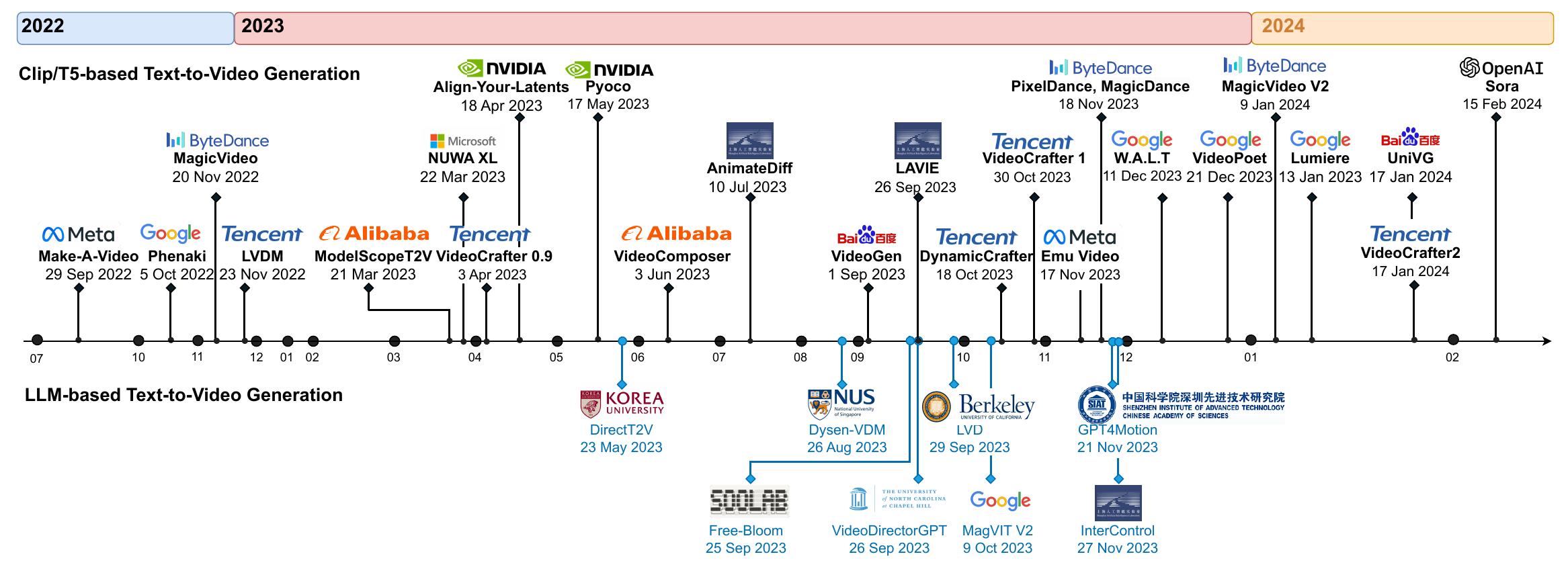}
    \caption{
    Milestone works of Clip/T5-based and LLM-based language-guided video generation.
    }
    \label{fig:video_timeline}
\end{figure*}

\subsection{Video Editing}
\subsubsection{Text-guided Video Editing with CLIP/T5}
CLIP~\cite{radford2021learning} enables language-based video editing. Here, we mainly discuss the diffusion-model-based video editing approaches. Tune-A-Video~\cite{wu2023tune} presents the early attempt at text-guided video editing with pre-trained diffusion models. Instead of training a giant video diffusion model, Wu et al.~\cite{wu2023tune} propose to inflate and tune the pre-trained text-to-image diffusion model on the target video in a one-shot way. After tuning, the inflated diffusion models support versatile video editing capabilities. Despite its simplicity, Tune-A-Video~\cite{wu2023tune} shows poor temporal stability and the limitation of keeping the unrelated regions unaltered. Video-P2P~\cite{liu2023video} and FateZero~\cite{qi2023fatezero} utilize better inversion techniques and propose manipulating attention maps to keep the background unchanged when editing. Pix2Video~\cite{ceylan2023pix2video} adopts editing the key frame first and then propagates the editing to other frames, which achieves improved temporal consistency and longer video editing capabilities. Different from these works, Rerender-A-video~\cite{yang2023rerender} and CoDeF~\cite{ouyang2023codef} focus their applications on video-to-video style translation and achieves impressive results through optical-flow-based regularization~\cite{yang2023rerender} or adoption of deformable content field~\cite{ouyang2023codef}. More recently, the developments of CLIP-based video editing are towards better temporal consistency~\cite{geyer2023tokenflow, chai2023stablevideo, liew2023magicedit}, more controllable~\cite{ma2023magicstick}, and more computational effective~\cite{bao2023latentwarp}. We list recent representative video editing works in Table~\ref{tab:video-edit}.

\subsubsection{Text-guided Video Editing with LLMs}
Existing works that utilize LLMs for video editing are relevantly limited. Current LLMs-based video editing follows a similar scheme as the InstructPix2pix~\cite{brooks2023instructpix2pix}, i.e., using LLMs to construct the training data more efficiently. 

InstructVid2Vid~\cite{qin2023instructvid2vid} is one of the works that involve LLMs inefficient training data construction. The method uses an LLM model to generate synthetic video-instruction pairs, which are then used to train an editing model for controllable video editing with natural language instructions. InstructVid2Vid leverages the pre-trained image generation model, i.e., stable diffusion, and a conditional 3D Unet to produce high-quality and temporally coherent videos that match the input video and instruction. To improve the diversity and realism of the synthetic data, InstructVid2Vid incorporates the knowledge and expertise of different models, such as ChatGPT, BLIP, and Tune-a-Video, to synthesize various instructions for the same video. The paper demonstrates the effectiveness of using LLMs to synthesize training data for complex and creative tasks, such as attribute editing, background change, and style transfer.

InsV2V~\cite{cheng2023consistent} is another approach that extends the paradigm of InstructPix2Pix to the video editing domain. InsV2V uses a large language model to construct synthetic data for training a video editing model, which can also follow natural language instructions to edit videos. InsV2V adopts a one-model-all-video strategy, eliminating the need for per-video-per-model fine-tuning or inversion, and simplifying the user interaction by only requiring an editing prompt. InsV2V leverages a pre-trained image generation model, Stable Diffusion, and a conditional 3D U-Net architecture to produce high-quality and temporally coherent videos that match the input video and instruction. InsV2V demonstrates the versatility and effectiveness of using LLMs to synthesize training data and perform text-based video editing for various tasks, such as object replacement, style transfer, and background change.

\subsection{Video-language Datasets}
The availability of captioned video datasets is crucial for text-to-video generation. 
To address this challenge, MSR-VTT~\cite{MSR-VTT} has introduced a large-scale open-domain video-language dataset that encompasses a wide range of categories and diverse content, setting a new benchmark for the video understanding task in 2016. 
It contained 200k clip-language pairs from 10K videos on the web, and every video had 20 human annotations in English.
Anna et al. have presented the large-scale movie description challenge (LSMDC)~\cite{LSMDC}, which consists of 202 movies accompanied by transcribed audio descriptions. These descriptions provide a narrative of the significant events depicted in the visual video.
In natural videos, multiple events often occur within a single video. For instance, a video may feature a man playing the piano, a girl singing, and a crowd applauding. To identify and describe each event, Ranjay et al. have proposed the ActivityNet Caption~\cite{ActivityNetCaptions} benchmark, which involves detecting events, describing them using natural language, and localizing them with start and end times.
How2~\cite{How2} and VATEX~\cite{VaTeX} are multilingual video description datasets. How2 is a large-scale instructional multimodal and multilingual video dataset that includes English and Portuguese descriptions, videos, speech, and English video-level summaries. VATEX comprises both English and Chinese descriptions, covering 600 human activities.
HowTo100M~\cite{HowTo100M} has introduced an automatic video captioning method that utilizes transcribed narrations from web videos instead of manual labeling, enabling fast and scalable data collection.
Jonathan et al. have observed that web videos often come with text metadata such as titles and descriptions. They have proposed a data collection process and gathered 70M video clips, known as WTS70M~\cite{WTS70M}, using metadata including titles, descriptions, tags, and channel names.
The WebVid~\cite{Webvid} dataset has been created for the text-to-video retrieval task. Recognizing the noise in previous datasets like HowTo100M, WebVid-2M, and WebVid-10M have been collected from the internet, featuring weak captions.
YT-Temporal-180M~\cite{YT-Temporal-180M} is a dataset that contains a diverse corpus of frames, where ASR is provided from a filtered set of 6 million YouTube videos, serving as a resource for multimodal representation learning.
HDVILA~\cite{HD-VILA} is a high-resolution large-scale dataset comprising 370k videos, covering 15 popular YouTube categories and offering diverse video content.
VideoCC3M~\cite{VideoCC3M} proposes a method to transfer captions from existing image captions in CC3M, creating a new weakly labeled audio-video captioning dataset.
VideoFactory~\cite{HD-VG-130M} introduces HD-VG-130M, a dataset consisting of 130M high-definition, widescreen, and watermark-free text-video pairs.
InternVid~\cite{InternVid} presents a scalable approach to building a high-quality video-text dataset. They employ a multi-scale approach that leverages Tag2Text, LLM, and BLIP2 to generate video captions.
Panda-70M~\cite{panda70m} is a high-quality and large-scale captioned video dataset proposed in 2024.
It contains 70M video clips from YouTube videos, and the caption is extracted via multiple teacher models to obtain multiple captions for one video and a well-trained caption retrieval model to select the best caption.
Vript~\cite{vript} is a fine-grained video-text dataset proposed in 2024 containing 12K annotated videos. Although the number of videos is limited, the caption of each video is fine-grained and comprises information on the shot type, camera movement, content, and scene title.

\subsection{Summary}
In Sec.~\ref{sec: video}, we have introduced the research works on the generation and editing of the video modality.
For each task, we divided the papers into two groups: CLIP/T5-based methods and LLMs-based methods to highlight the advancement brought by LLMs.
We summarize the key technical components of LLMs-based approaches in Table~\ref{tab:video_generation_llm}, the development of milestone works in the task of language-guided video generation in Fig.~\ref{fig:video_timeline}, and the related video-language dataset in Table~\ref{tab:video_dataset}

\begin{table*}[!ht]
    \centering
    \caption{
        Public video-language datasets can be adopted for language-guided video generation.
        For each dataset, we list the following information in each column: dataset name (Dataset), paper conference venue (Venue), dataset domain (Domain), video source (Vid. Source), video spatial resolution (Res.), average duration per clip (Dur./Clip), total number of clips ($\#$Clips), total number of videos ($\#$Videos), total number of hours ($\#$Hours), and caption source (Cap. Source).
        The dataset is sorted in ascending order of its released time.
        }
    \scalebox{0.85}{
    \begin{tabular}{llllllllllll}
    \toprule
        \textbf{Dataset}  & \textbf{Venue} & \textbf{Domain} & \textbf{Vid. Source} & \textbf{Res.} & \textbf{Dur./Clip} & \textbf{\#Clips} & \textbf{\#Videos} & \textbf{\#Hrs} & \textbf{Cap. Source} \\ \midrule
        MSR-VTT~\cite{MSR-VTT} & CVPR 2016 & Open & Internet & 240p & 15s & 10K & 7K & 41 & Human \\ 
        LSMDC~\cite{LSMDC} & IJCV 2017 & Movie & Amazon & 1080p & 4.8s & 118K & 202 & 158 & Audio Desc. \\ 
        ActivityNet Captions~\cite{ActivityNetCaptions} &  ICCV 2017 & Activity & Internet & - & 36s & 100K & 20K & 849 & Human \\ 
        How2~\cite{How2} & NIPS 2018 & \scriptsize{Instruction} & Youtube & - & 90s & 80K & - & 298 & Human \\ 
        VATEX~\cite{VaTeX} & ICCV 2019 & Open & Youtube & 240p & 10s & 41K & 41K & - & Human \\ 
        HowTo100M~\cite{HowTo100M} & ICCV 2019 & \scriptsize{Instruction} & Youtube & 240p & 4s & 136M & 1.2M & 134K & ASR \\ 
        WTS70M~\cite{WTS70M} & arXiv 2020 & Open & Youtube & - & 10s & 70M & 70M & 194K & Metadata \\ 
        WebVid-10M~\cite{Webvid} & ICCV 2021 & Open & Internet & 360p & 18s & 10.7M & 10.7M & 52K & Alt-text \\ 
        YT-Temporal-180M~\cite{YT-Temporal-180M} & NeurIPS 2021 & Open & Youtube & - & - & 180M & 6M & - & ASR \\ 
        HD-VILA-100M~\cite{HD-VILA} & CVPR 2022 & Open & Youtube & 720p & 13.4s & 103M & 3.3M & 372 & Algorithm \\ 
        CelebV-Text~\cite{CelebVText} & CVPR 2023 & Face & Internet & 512$^2$+ & 14s & 70K & - & 279 & Human+Algorithm \\ 
        VideoCC3M~\cite{VideoCC3M} & ECCV 2022 & Open & CC3M & - & 10s & 6.3M & - & 17.5K & CC3M \\ 
        HD-VG-130M~\cite{HD-VG-130M} & arXiv 2023 & Open & Youtube & 720p & - & 130M & - & - & Algorithm \\ 
        InternVid~\cite{InternVid} & ICLR 2024 & Open & Youtube & 720p & 12s & 234M & 7M & 760K & Algorithm \\ 
        Panda-70M~\cite{panda70m} & CVPR 2024 & Open & Youtube & 720p & 8.5s & 70.8M & - & 166.8K & Algorithm \\
        Vript~\cite{vript} & Github & Open & Youtube & 720p-2K & 11.7s & 400K & 12K & 1.3K & Algorithm \\ 
    \bottomrule
    \end{tabular}
    \label{tab:video_dataset}
    }
\end{table*}

\section{3D Generation and Editing}
\label{sec: 3D}

\begin{figure*}[h]
    \centering
    \includegraphics[width=\linewidth]{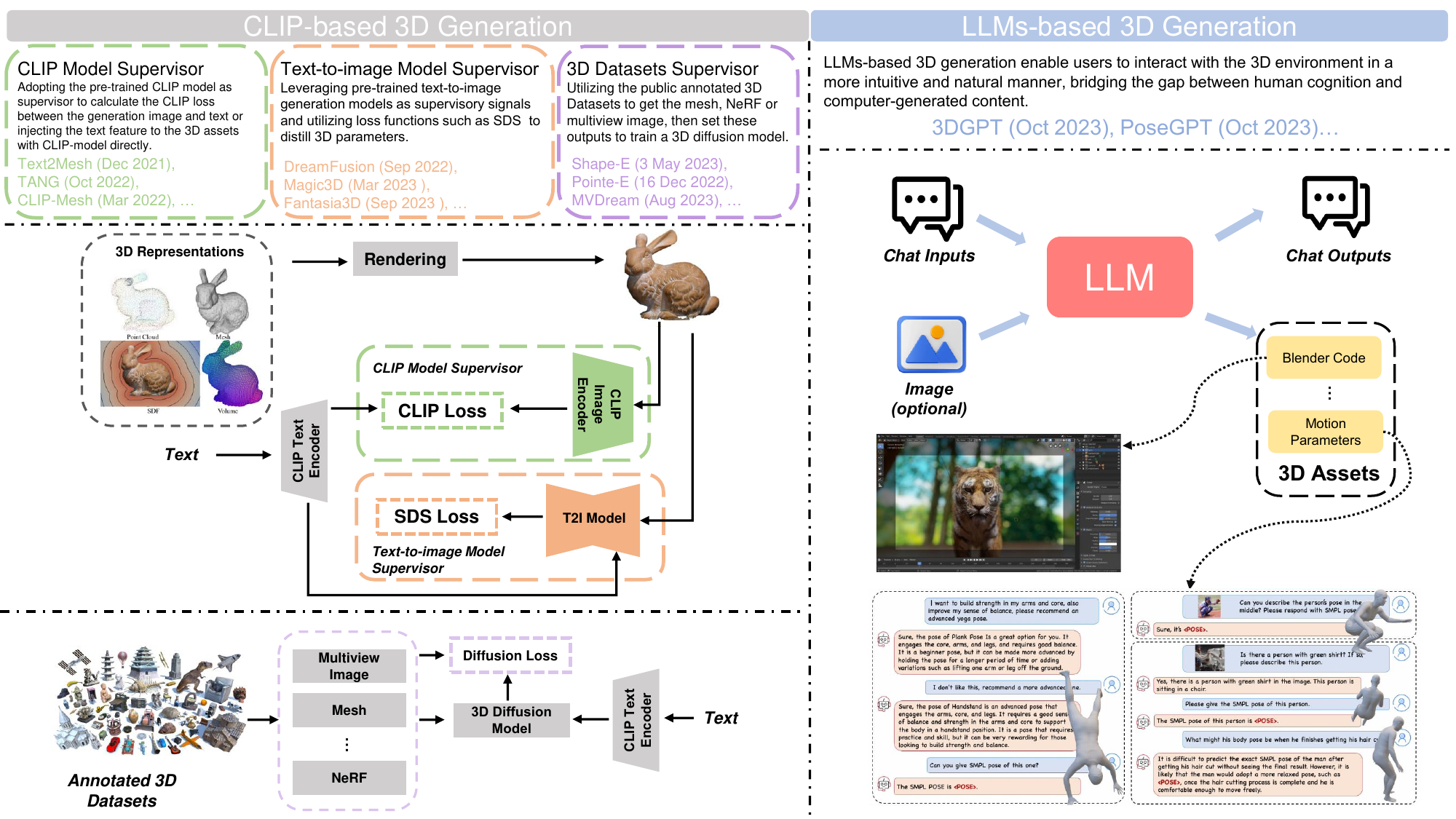}
    \caption{A generic pipeline of 3D generation with CLIP and LLMs. The CLIP-based models optimize the 3D representation by minimizing the distance between the rendering image and the text prompt. In order to better improve interaction efficiency, LLMs-based methods try to transfer the language outputs of LLMs to blender code or 3D representation (i.e. human motion) directly. Some images are borrowed from~\cite{li2024advances, feng2023posegpt, deitke2023objaverse}
    }
    \label{fig:3d_pipeline}
\end{figure*}
Recent studies have focused on establishing a connection between 3D assets and texts. There are two types of methods that can bring the text information for 3D assets, including LLMs and CLIP/T5 models. LLMs can make the output of 3D generation and understanding iterative updates according to user requirements, thereby facilitating highly effective human-computer interaction (i.e., guiding the generation of human motions based on language). Different from LLMs, which directly influence the 3D assets in an interactive iterative manner, the CLIP/T5 model integrates the features of both rendered images and text, enabling the injection of textual information into the 3D assets. In this section, we will explore various methods that leverage the CLIP model or LLMs to guide the processes of 3D generation, editing, and understanding. The generic pipeline is shown in Fig.~\ref{fig:3d_pipeline}. The corresponding overview of the 3D generation and editing methods is shown in Table~\ref{tab:3d}.

\subsection{3D Generation}
\subsubsection{3D Generation with CLIP/T5}
By leveraging the multi-modal representation capabilities of CLIP, researchers have been able to guide the generation and editing of 3D assets using textual descriptions or queries, thereby enabling more precise control and customization. Specifically, CLIP~\cite{radford2021learning} extracts the feature of image and text with two encoders and aligns them in a contrastive learning manner. This alignment builds the connection between image and text effectively which brings significant improvements to text-guided 2D~\cite{ramesh2021zero, ldm, dalle2hierar, nichol2021glide, saharia2022photorealistic, sanghi2022clip}, text-guided 3D~\cite{sanghi2022clip, wang2022clip, poole2022dreamfusion} generations. There are three typical methods to utilize the CLIP model to provide text information during 3D generation: 1. Adopting the pre-trained CLIP model as supervisor to calculate the CLIP loss between the generation image and text or injecting the text feature to the 3D assets with CLIP model directly. 2. Leveraging pre-trained text-to-image generation models as supervisory signals and utilizing distillation loss functions such as SDS~\cite{poole2022dreamfusion,wang2023score}  to distill 3D assets~\cite{mildenhall2021nerf, wang2021neus,shen2021deep}. 3. Utilizing the public annotated 3D Datasets~\cite{deitke2023objaverse, deitke2024objaverse, reizenstein2021common,nichol2022point} to get the mesh, NeRF or multiview image, then set these outputs to train a 3D diffusion model. As mentioned above, we will elaborate on the current methods in two categories in the following content.

{\flushleft \bf CLIP/T5 Model Supervisor.}
Text2Mesh~\cite{michel2022text2mesh} focuses on stylizing 3D meshes by predicting color and local geometric details that align with target text prompts. The entire process is guided by a CLIP loss, which helps ensure the generated meshes conform to the desired textual specifications. This method offers enhanced control over the visual appearance and geometric attributes of 3D meshes, enabling the creation of visually appealing and semantically meaningful shapes.

TANGO~\cite{lei2022tango} proposes a pipeline for generating texture on a given mesh. By leveraging the CLIP model, TANGO aligns the texture generation process with textual descriptions, allowing for the synthesis of textured meshes that match specific visual or semantic criteria. This approach facilitates the creation of realistic and visually coherent 3D models with detailed surface textures. CLIP-Mesh~\cite{mohammad2022clip} addresses unsupervised text-guided 3D generation by optimizing the texture, normal, and vertical position of a 3D object using the CLIP loss. This approach enables the generation of 3D objects that align with textual prompts, providing a powerful tool for text-driven content creation and design. X-mesh~\cite{ma2023x} further improves the performance of CLIP-Mesh by adopting an attention-based network, enhancing the fidelity and accuracy of the generated 3D meshes. CLIP-forge~\cite{sanghi2022clip} introduces a zero-shot text-to-shape method for predicting volumetric occupancy by leveraging the text feature of the CLIP model in conjunction with a conditional normalizing flow network~\cite{dinh2016density}. This approach enables the generation of 3D shapes based on textual prompts without the need for explicit supervision or labeled training data.

Some methods employ CLIP guidance to output NeRF representations~\cite{mildenhall2021nerf}, which are used to model complex scenes and capture high-frequency spatial information. DreamFields~\cite{jain2022zero} introduces general-purpose priors to assist in aligning the optimized NeRF with the given text prompt, improving the quality and fidelity of the generated scenes. CLIP-NeRF~\cite{wang2022clip} takes a two-step approach, first training a disentangled conditional NeRF and then utilizing the text feature to adjust the parameters of the learned NeRF, enabling more fine-grained control over the generated scenes.

ShapeGPT~\cite{yin2023shapegpt} utilizes a "word-sentence-paragraph" pipeline to convert shapes into words. These words are then assembled to form shape sentences, which can be integrated with instructional text to create multi-modal paragraphs describing 3D shapes. These multi-modal paragraphs help the ShapeGPT do several applications, including text-to-shape generation, image-to-shape generation, and multimodal-to-shape completion and editing.  

Moving beyond object generation, MotionCLIP~\cite{tevet2022motionclip} proposes a 3D human motion auto-encoder that predicts pose sequences. By leveraging CLIP, this method enables the generation of realistic and contextually coherent human motion based on textual prompts, providing a means for text-driven animation and virtual character control. MotionGPT~\cite{jiang2024motiongpt} considers human motion as a distinct language and trains a motion-language model with the T5 model. The approach incorporates discrete vector quantization to represent human motion and transforms 3D motion into motion tokens. By establishing a comprehensive "motion vocabulary," the model conducts language modeling on both motion and text in a cohesive manner.

While the aforementioned methods have achieved notable success in text-guided generation, they still face challenges related to visual artifacts. This can be attributed to the semantic-level nature of the CLIP loss, which tends to reduce the high-frequency spatial information in the generated images. Addressing this limitation remains an active area of research, with the aim of further improving the visual quality and fidelity of text-guided 3D generation approaches.

{\flushleft \bf Text-to-image Model Supervisor.}
In contrast to the aforementioned methods that directly utilize the CLIP model or CLIP loss for 3D generation, recent approaches~\cite{seo2023let,wang2023score,lorraine2023_att3d,zhu2023_hifa,zhou2023sparsefusion,xu2023_dream3d,hoellein2023text2room,li2023_sweetdreamer,tsalicoglou2023textmesh,wan2023cad,yang2024dreamspace} have focused on distilling 3D assets from pre-trained text-to-image generation models (i.e., Stable Diffusion~\cite{ldm}). These models employ score distillation sampling (SDS) loss within the framework of DreamFusion~\cite{poole2022dreamfusion}. Specifically, text-to-2D image generation models utilize the text feature from CLIP to train a generative model (i.e., diffusion model) within a text-image pair dataset, SDS~\cite{poole2022dreamfusion} simulates the training process of diffusion model and calculates the spatial gradient of the rendered image to optimize the parameters of NeRF.  Many subsequent methods have utilized the SDS loss function as a supervisory signal and introduced various techniques to improve the performance of 3D generation.

For instance, Magic3D~\cite{lin2023magic3d} combines DMTed~\cite{shen2021deep} to transfer the NeRF model to a mesh representation, enabling high-resolution rendering results. The experimental results presented in the paper demonstrate the effectiveness of Magic3D in enhancing the generation quality of both geometry and texture. Similar to the Magic3D, TextMesh~\cite{tsalicoglou2023textmesh} replaces the NeRF model with a textured mesh representation for 3D asset generation.   
LatentNeRF\cite{metzer2023latent} introduces a latent space optimization strategy. It takes advantage of the latent space structure learned by the text-to-image model, which has already been trained on a large-scale text-image dataset. By aligning the NeRF optimization process with the latent space structure, LatentNeRF improves the convergence and stability of the training process.

Fantasia3D~\cite{Chen_2023_ICCV} presents a novel approach to generating 3D assets by separating the geometry and texture components. It first refines the normal map to generate the geometry and then fixes the geometry to predict the color field. Moreover, Fantasia3D set the color field as a Physically Based Rendering (PBR) material model to enhance the generation fidelity. Moreover, artists and designers can manipulate and modify the geometry and texture components independently, allowing for a wider range of artistic expression and customization possibilities.

Despite the significant improvements brought about by SDS, there remain some limitations in the results generated. These include over-saturation, over-smoothing, multi-face Janus artifacts, and time-consuming computations. To address these issues and further enhance the generation performance, some methods~\cite{katzir2023noisefree,yu2023_csd,liang2023_luciddreamer} have been proposed. 

For example, Perp-Neg~\cite{armandpour2023re} debiases the score-distillation framework for view-consistent text-to-3D generation, aiming to mitigate the multi-face Janus problem.  Prolificdreamer~\cite{wang2023prolificdreamer} introduces variational score distillation (VSD) to avoid over-saturation and over-smoothing. It minimizes the KL divergence between the optimized 3D assets and the target distribution, which makes the generation process utilize the normal CFG weight.
 DreamPropeller~\cite{zhou2023dreampropeller} presents a drop-in acceleration algorithm to expedite the training process. It extends the concept of Picard iterations, a well-established algorithm for parallel sampling of an ODE path, to encompass various scenarios beyond ODEs. This includes accommodating momentum-based gradient updates and handling dimensional changes during the optimization process, which are often encountered in the context of 3D generation. Additionally, some methods~\cite{tang2023dreamgaussian, yi2023gaussiandreamer} have replaced the NeRF model with Gaussian splatting~\cite{kerbl3Dgaussians} techniques to improve training efficiency and generate high-quality geometry.

Apart from the methods mentioned, which focus on general object generation, some methods~\cite{yu2023towards,zhang2023styleavatar3d,wang2023rodin,aneja2023clipface,wu2023high,liao2023tada,huang2023avatarfusion,han2023headsculpt,cao2023dreamavatar,zhang2023avatarverse,zhang2023dreamface,hong2022avatarclip,kolotouros2023dreamhuman,huang2023humannorm,zeng2023avatarbooth,wang2024headevolver,liu2023headartist} try to explore 3D generation in 3D avatar generation.

{\flushleft \bf 3D Datasets Supervisor.}
Some methods~\cite{shi2023MVDream,kant2024spad,liu2023unidream,qiu2023richdreamer} try to train the multi-view diffusion model with diffusion loss or reconstruction model~\cite{li2023instant3d,tang2024lgm,xu2024grm} for text-to-3D generation, and these methods always adopt the text-to-3D dataset~\cite{deitke2023objaverse, deitke2024objaverse, reizenstein2021common} as a training label. 

Point-E~\cite{nichol2022point} and Shape-E~\cite{jun2023shap} utilize the Blender to collect a point cloud 3D dataset first, then train a diffusion model to generate the point cloud and mesh with text condition, respectively.

MVDream~\cite{shi2023MVDream}, RichDreamer~\cite{qiu2023richdreamer}, SPAD~\cite{kant2024spad} and UniDreamer~\cite{liu2023unidream} get the multi-view images from the Objaverse dataset. With these multi-view images, they Then, fine-tune the 2D text-to-image diffusion to a 3D diffusion model to predict the multi-view images under the guidance of text and camera poses.

While CLIP-based 3D generation methods have demonstrated impressive progress, the CLIP model itself cannot maintain flexibility in human-computer interaction during the 3D generation process. More recently, Large Language Models (LLMs) have been introduced to enhance human-computer interaction in 3D generation, which will be discussed in the following section.

\subsubsection{3D Generation with LLMs}
The integration of LLMs with 3D assets has emerged as a promising research direction recently. By harnessing the powerful language understanding capabilities of LLMs, researchers aim to directly enhance the performance of generation, manipulation,  or understanding of 3D assets through textual instructions. These approaches enable users to interact with the 3D environment more intuitively and naturally, bridging the gap between human cognition and computer-generated content.  

3D-GPT~\cite{sun20233d} proposes a training-free framework that utilizes LLMs. This framework consists of three agents: the task dispatch agent, the conceptualization agent, and the modeling agent. By employing these agents, 3D-GPT can produce the code for blender corresponding to the language, and the efficiency of end-users engaged in procedural 3D modeling is improved. Similar to the 3D-GPT, SceneCraft~\cite{hu2024scenecraft} introduces an LLM agent to transfer input text query into a 3D scene by generating a Blender script. Specifically, sceneCraft has a dual-loop self-improving pipeline: in the inner-loop, per each scene, an LLM autonomously writes a script to interact with Blender, receives rendered images, and keeps improving the script until getting good scenes; in the outer-loop, SceneCraft summarizes common functions over a batch of written scripts to maintain a reusable design skill library.

LL3DA~\cite{chen2023ll3da} proposed a Large Language 3D Assistant that utilizes a transformer network to predict querying tokens. These tokens are projected to the prefix of textual instructions as the input of a frozen LLM. Finally, the LLM will produce the answer to the textual instructions.
 
PointLLM~\cite{xu2023pointllm} processes colored point clouds using human instructions and predicts responses to user questions with the help of LLMs. This enables users to analyze and interpret point clouds more effectively. 

3D-LLM~\cite{hong20233d} takes 3D points with features and language prompts as input and performs various 3D-related tasks, leveraging the capabilities of LLMs.  The pipeline involved in this study involves the collection of a comprehensive dataset comprising over 300,000 instances of 3D-language data. This dataset encompasses a wide range of diverse 3D-related tasks, which include, but are not limited to, 3D captioning, dense captioning, 3D question answering, task decomposition, 3D grounding, 3D-assisted dialog, navigation, and various other tasks.

In summary, the integration of LLMs and 3D assets has opened up new possibilities for generating, manipulating, and understanding 3D content through natural language instructions. These methods have demonstrated significant advancements in enhancing human-computer interaction in the 3D domain.

\subsection{3D Editing}
Similar to the generation, we separate the text-to-3D editing into two aspects. And we will first discuss the CLIP/T5-based editing methods.  

\subsubsection{3D Editing with CLIP/T5}
{\flushleft \bf CLIP/T5 model Supervisor.}
Blended-NeRF~\cite{gordon2023blended} introduced a framework for modifying a specific region of interest within an existing NeRF scene using CLIP loss. This approach allows for targeted editing of NeRF scenes by leveraging the power of contrastive learning with CLIP. NeRF-Art~\cite{wang2023nerf} proposed a global-local contrastive learning strategy to stylize pre-trained NeRF models. By employing contrastive learning, NeRF-Art enables the creation of artistic and stylized renderings through the manipulation of pre-existing NeRF scenes. TextDeformer~\cite{gao2023textdeformer} employed text-to-geometry manipulation by introducing a mesh deformation technique based on Jacobians. This approach utilizes text descriptions to deform the geometry of objects, offering a novel way to manipulate and edit 3D models. Sine~\cite{bao2023sine} presented a prior-guided editing field that encodes fine-grained geometric and texture modifications. By leveraging this approach, Sine enables precise and detailed editing of both geometry and texture in 3D scenes, providing a powerful tool for creative expression. 

{\flushleft \bf Text-to-image Model Supervisor.}
SKED~\cite{Mikaeili_2023_ICCV} utilizes sketches as a guiding input for text-to-image generation models in SDS. By incorporating sketch information, SKED enhances the generation process, resulting in more accurate and contextually aligned text-to-image translations. DreamEditor~\cite{zhuang2023dreameditor} transfers NeRF representations to meshes and employs a personalized text-to-image generation model called DreamBooth for mesh editing. This approach allows for interactive and personalized editing of meshes using text-based instructions, enabling users to modify and shape 3D scenes according to their preferences. Instruct-NeRF2NeRF~\cite{haque2023instruct} utilizes the InstructPix2Pix model to incorporate SDS loss for editing NeRF scenes with text-based instructions. By combining the power of text instructions with NeRF editing, this approach enables users to modify scenes in a controlled and precise manner, enhancing the interactive editing capabilities of NeRF. 3D Paintbrush~\cite{decatur20233d} proposes a cascaded score distillation method for editing the texture of local semantic regions in meshes. By distilling scores, this approach allows for targeted and localized texture editing, providing users with a powerful tool for enhancing the visual appearance of specific regions in 3D models.

\subsubsection{3D Editing with LLMs}
Different from CLIP-based methods,  there is no specific method to do the 3D editing with LLMs. Editing is more like a sub-task of LLMs-based 3D generation, so some generation methods (i.e., 3D-GPT~\cite{sun20233d}, SceneCraft~\cite{hu2024scenecraft} ) can edit the 3D assets directly.  We will follow the latest developments in LLMs-based 3D editing and discuss them in the future.

\subsection{Summary}
The utilization of CLIP or LLMs in the context of 3D generation and editing offers several advantages. Firstly, it enables users to express their creative intentions or desired modifications in natural language, simplifying the interaction process and reducing the need for specialized software or technical expertise. Additionally, the incorporation of textual information into the 3D generation pipeline enhances the interpretability and explainability of the generated outputs, allowing users to better understand and fine-tune the results according to their requirements.

In conclusion, the integration of CLIP or LLMs with 3D assets has opened up new avenues for highly effective human-computer interaction. By aligning textual information with the visual features of 3D assets, researchers have been able to facilitate more intuitive and precise control over the generation and editing of 3D content. These advancements hold great potential for applications in fields such as computer graphics, virtual reality, and augmented reality, offering enhanced user experiences and empowering users to unleash their creativity more seamlessly and efficiently.

\begin{table*}[]
\centering
\caption{Summary for 3D general object generation. 
The optimization target means the essential constraint during the learning process. The representation means the type of the 3D output. The method without an optimization target means this method is not guided by CLIP loss or SDS-based loss.
}
\begin{tabular}{ccccc}
\hline
Method                                     & Venue                  & Optimization Target  & Representation        & Guided Model         \\ \hline
\multicolumn{1}{l}{\textbf{\textit{CLIP/T5 for 3D generation}}} & \multicolumn{1}{l}{}   & \multicolumn{1}{l}{} & \multicolumn{1}{l}{} & \multicolumn{1}{l}{} \\
MotionCLIP~\cite{tevet2022motionclip}      & ECCV 2022              & CLIP Loss            & Motion Sequences     & CLIP                    \\
MotionGPT~\cite{jiang2024motiongpt}        & NeurIPS 2023           & -                 & Motion Sequences           & T5                    \\
MDM~\cite{tevet2023human}                  & ICLR 2023           & -                 & Motion Sequences                &  CLIP                    \\
CLIP-Mesh~\cite{mohammad2022clip}          & SIGGRAPH Asia 2022     & CLIP Loss            & Mesh                 & CLIP                    \\
TANGO~\cite{lei2022tango}                  & NeurIPS 2022 Spotlight & CLIP Loss            & Mesh                 & CLIP                    \\
DreamFields~\cite{jain2022zero}            & CVPR 2022              & CLIP Loss            & NeRF                 & CLIP                    \\
Clip-forge~\cite{sanghi2022clip}           & CVPR 2022              & CLIP Loss            & Voxel                & CLIP                    \\
Text2Mesh~\cite{michel2022text2mesh}       & CVPR 2022              & CLIP Loss            & Mesh                 & CLIP                    \\
TextMesh~\cite{tsalicoglou2023textmesh}    & 3DV 2023               & CLIP Loss            & Mesh                 & CLIP                    \\
X-Mesh~\cite{ma2023x}                      & ICCV 2023              & CLIP Loss            & Mesh                 & CLIP                    \\
ShapeGPT~\cite{yin2023shapegpt}            & arXiv 2023                  & -                 & SDF                  & T5                    \\
Shape-E~\cite{jun2023shap}                 & arXiv 2023             & Diffusion Loss        & Mesh/NeRF          & CLIP               \\
Point-E~\cite{nichol2022point}             & arXiv 2023             & Diffusion Loss        & PointCloud          & CLIP               \\
DreamFusion~\cite{poole2022dreamfusion}    & ICLR 2023 Oral         & Score Distillation   & NeRF                 & Imagen              \\
SJC~\cite{wang2023score}                   & CVPR 2023              & Score Distillation   & NeRF                 & SD                    \\
Magic3D~\cite{lin2023magic3d}              & CVPR 2023 Highlight    & Score Distillation   & NeRF                 & SD                    \\
Perp-Neg~\cite{armandpour2023re}           & arXiv 2023                  & Score Distillation   & NeRF                 & SD                    \\
Latent-NeRF~\cite{metzer2023latent}        & CVPR 2023              & Score Distillation   & NeRF                 & SD                  \\
Fantasia3D~\cite{chen2023fantasia3d}       & ICCV 2023              & Score Distillation   & NeRF                 & SD                   \\
ATT3D~\cite{lorraine2023_att3d}            & ICCV 2023              & Score Distillation   & NeRF                 & SD                   \\
ProlificDreamer~\cite{wang2023prolificdreamer} & NeurIPS 2023 Spotlight & Score Distillation   & NeRF                 & SD                   \\
Text2Room~\cite{hoellein2023text2room}     & ICCV 2023              & Score Distillation   & Mesh                 & SD                   \\
3DFuse~\cite{seo2023let}                   & ICLR 2024              & Score Distillation   & NeRF                 & SD                   \\
GaussianDreamer~\cite{yi2023gaussiandreamer} & CVPR 2024            & Score Distillation   & Gaussian Splatting   & SD                   \\
DreamGaussian~\cite{tang2023dreamgaussian}  & ICLR 2024             & Score Distillation   & Gaussian Splatting   & SD                   \\
NFSD~\cite{katzir2023noisefree}            & ICLR 2024              & Score Distillation   & NeRF                 & SD                   \\
MVDream~\cite{shi2023MVDream}              & ICLR 2024              & Diffusion Loss       & Multi-view images    & SD                   \\
RichDreamer~\cite{qiu2023richdreamer}     & CVPR 2024              & Diffusion Loss       & Multi-view images    & SD                   \\
SPAD~\cite{kant2024spad}                   & CVPR 2024              & Diffusion Loss       & Multi-view images    & SD                   \\
UniDreamer~\cite{liu2023unidream}          & CVPR 2024              & Diffusion Loss       & Multi-view images    & SD                   \\
Enhancing3D~\cite{pan2024enhancing}         & ICLR 2024              & Score Distillation   & NeRF                 & SD                   \\
LucidDreamer~\cite{liang2023_luciddreamer} & CVPR 2024              & Score Distillation   & Gaussian Splatting   & SD                   \\
CSD~\cite{yu2023_csd}                      & ICLR 2024              & Score Distillation   & NeRF                 & SD                   \\
SweetDreamer~\cite{li2023_sweetdreamer}    & ICLR 2024              & Score Distillation     & NeRF                 & SD                   \\
HiFA~\cite{zhu2023_hifa}                   & ICLR 2024              & Score Distillation     & NeRF                 & SD                   \\
AToM~\cite{qian2024atom}                   & arXiv 2023                   & Score Distillation     & Mesh                 & SD                     \\
Consistent3D~\cite{wu2024consistent3d}     & arXiv 2023                   & Score Distillation     & Mesh/NeRF           & SD                      \\
DreamControl~\cite{huang2023dreamcontrol}  & CVPR 2024              & Score Distillation     & NeRF                  & SD                \\
IT3D~\cite{chen2023it3d}                   & AAAI 2024               & Score Distillation     & NeRF                  & SD                \\
Efficientdreamer~\cite{zhao2023efficientdreamer} & CVPR 2024            & Score Distillation     & NeRF                  & SD               \\
GSGEN~\cite{chen2023text}                  & CVPR 2024              & Score Distillation     & Gaussian Splatting    & SD                \\
X-Dreamer~\cite{ma2023xdreamer}             & arXiv 2023              & Score Distillation     & Gaussian Splatting    & SD                \\
HD-Fusion~\cite{wu2024hd}                  & WACV 2024              & Score Distillation     & Gaussian Splatting    & SD                \\
LODS~\cite{yang2023learn}                   & arXiv 2023                    & Score Distillation     & Gaussian Splatting    & SD                \\
Sherpa3d~\cite{liu2023sherpa3d}             & CVPR 2024               & Score Distillation     & NeRF                 & SD                \\
DreamPropeller~\cite{zhou2023dreampropeller} & CVPR 2024                & Score Distillation     & NeRF                 & SD                   \\
DreamPolisher~\cite{lin2024dreampolisher}   & arXiv 2024                & Score Distillation     & Gaussian Splatting                 & SD    \\
\hline
\textbf{\textit{LLM for 3D generation}}    & \multicolumn{1}{l}{}   & \multicolumn{1}{l}{} & \multicolumn{1}{l}{} & \multicolumn{1}{l}{} \\
3D-GPT~\cite{sun20233d}                    & arXiv 2023                   & -                     & Blender Code         & GPT-3.5              \\
PoseGPT~\cite{feng2023posegpt}             & CVPR 2024                  & -                    & Motion Sequences   & LLaVA                \\
HOLODECK~\cite{yang2023holodeck}           & CVPR 2024                 & -                     & Scene               & GPT-4                   \\
LL3DA~\cite{chen2023ll3da}                 & arXiv 2023                   & -                     & PointCloud           & GPTV                    \\  
SceneCraft~\cite{hu2024scenecraft}         & arXiv 2023                   & -                     & Blender Code         & GPT-3.5                    \\ \hline
\textbf{\textit{CLIP for 3D editing}}     & \multicolumn{1}{l}{}   & \multicolumn{1}{l}{} & \multicolumn{1}{l}{} & \multicolumn{1}{l}{} \\
CLIP-NeRF~\cite{wang2022clip}              & CVPR 2022              & CLIP Loss            & NeRF                 & CLIP                    \\
Blended-NeRF~\cite{gordon2023blended}      & ICCVW 2023          & CLIP Loss            & NeRF                 & CLIP                    \\
SKED~\cite{Mikaeili_2023_ICCV}              & ICCV 2023              & Score Distillation   & NeRF                 & SD                   \\
DreamEditor~\cite{zhuang2023dreameditor}    & SIGGRAPH Asia 2023     & Score Distillation   & NeRF                 & SD                    \\
Instruct-NeRF2NeRF~\cite{haque2023instruct}  & SIGGRAPH Asia 2023     & Score Distillation   & NeRF                 & SD                    \\
TextDeformer~\cite{gao2023textdeformer}     & TVCG 2022              & Score Distillation   & Mesh                 & SD                    \\
SINE~\cite{bao2023sine}                     & CVPR 2023              & Score Distillation   & NeRF                 & SD                   \\
Blending-NeRF~\cite{song2023blending}        & ICCV2023               & CLIP Loss           & NeRF                 & CLIP                     \\
CustomNeRF~\cite{he2023customize}            & CVPR 2024              & Score Distillation   & NeRF                 & SD               \\
Paint3D~\cite{zeng2023paint3d}               & arXiv 2023              & -                     & Mesh                 & SD               \\ 
3D Paintbrush~\cite{decatur20233d}          & arXiv 2023                & Score Distillation   & NeRF                 & SD                  \\
\bottomrule
\end{tabular}
\label{tab:3d}
\end{table*}
\section{Audio Generation, Understanding and Editing}
\label{sec: audio}
Recently, a surge of innovative works, such as ~\cite{yuan2024chatmusician, ghosal2023text, huang2023audiogpt, huang2023make2, zhang2023speechgpt, vyas2023audiobox}, has demonstrated the utilization of LLMs in various audio-related tasks. These tasks span across domains including the creation of audio effects, speech processing, and music composition, showcasing the versatility of LLMs.The roles of LLMs in these areas are varied, acting as backbones for complex systems~\cite{zhang2023speechgpt, gong2023listen,deshmukh2023pengi, rubenstein2023audiopalm, liu2024music, chen2023lauragpt, gardner2023llark, tang2023salmonn, chu2023qwen-audio, hussain2023m,shu2023llasm,yuan2024chatmusician,ding2024songcomposer}, conditioners for specific tasks~\cite{wu2023music, ghosal2023text, wu2024improving}, labellers for audio content~\cite{huang2023make2, wang2023assessing, vyas2023audiobox}, agents in interactive environments~\cite{shen2023hugginggpt, huang2023audiogpt, liu2023wavjourney, yu2023musicagent, zhang2023loop, zhuo2023lyricwhiz}, and as inspiration for some approaches~\cite{wang2023valle, agostinelli2023musiclm, dhariwal2020jukebox, copet2024musicgen, borsos2023audiolm, yang2023uniaudio}. This surge in the application of LLMs in the audio field is not only reshaping how we interact with sound and music but also opening new frontiers in the intersection of AGI and audio technologies.

\begin{table*}[!ht]
    \centering
    \caption{Audio datasets that can be adopted for language-based audio research. For each dataset, we list the following information in each column: dataset name (Dataset), paper conference venue (Venue), average duration per clip (Dur./Clip), total number of clips ($\#$Clips), total number of hours ($\#$Hours), and dataset domain (Domain).}
    \begin{tabular}{llllll}
    \toprule
        \textbf{Dataset} & \textbf{Venue} & \textbf{Dur./Clip} & \textbf{\#Clips} & \textbf{\#Hours} & \textbf{Domain} \\ \midrule

        MagnaTagATune~\cite{law2009evaluation} & ISMIR 2009 & 29s & 25,863 & 208h & Music \\
        Librispeech~\cite{panayotov2015librispeech} & ICASSP 2015 & - & - & 1,000h & Speech \\
        Audioset~\cite{gemmeke2017audio} & ICASSP 2017 & 10s & 2M & -  & Audio \\
        MAESTRO~\cite{hawthorne2018enabling} & ICLR 2019 & - & - & 200h & Music \\
        Libri-TTS~\cite{zen2019libritts} & INTERSPEECH 2019 & - & - & 585h & Speech \\
        MTG-Jamendo~\cite{bogdanov2019mtg} &	ICMLw 2019 & - & 55,000 & - & Music \\
        Librilight~\cite{kahn2020libri} & ICASSP 2020 & - & - & 60,000h &Speech \\
        Vggsound~\cite{chen2020vggsound} &	ICASSP 2020 & 10s & 210,000 & 550h & Audio \\
        WenetSpeech~\cite{zhang2022wenetspeech} & ICASSP 2022 & - & - & 22,400h & Speech \\ 
        Libri-heavy~\cite{kang2024libriheavy} & ICASSP 2024 & - & - & 50,000h & Speech \\ 
    \bottomrule
    \end{tabular}
        \label{tab:audio_dataset}
\end{table*}

\begin{table*}[]
\centering
\caption{Summary of approaches for LLMs-related audio tasks: Generation (G), Understanding (U), and Editing (E). 
We categorize the methods into five types according to the role of LLMs: LLMs as backbone, LLMs inspired backbone, LLMs as conditioner, LLMs as agent, and LLMs as labeller.
}
\begin{tabular}{ccccc}
\hline
    \textbf{Task}  & \textbf{Method} & \textbf{Venue}  & \textbf{LLM Model} & \textbf{Domain}    \\ \hline
    \multicolumn{1}{c}{\textbf{\textit{LLMs as backbone}}} & \multicolumn{1}{l}{}   & \multicolumn{1}{l}{} & \multicolumn{1}{l}{} & \multicolumn{1}{l}{} \\
      G, U   & SongComposer~\cite{ding2024songcomposer} & arXiv 2024     &  SongComposer & Audio music, Speech         \\
      G, U   & ChatMusician~\cite{yuan2024chatmusician} & arXiv 2024     & Llama 2  & Symbolic music             \\
      G, U   & AnyGPT~\cite{zhan2024anygpt} & arXiv 2024     &      Llama 2         & Audio, Audio music         \\
      G   & Boosting Large~\cite{hao2023boosting} & arXiv 2023     & LLaMA         & Speech    \\
      G, U   & Unified-IO 2~\cite{lu2023unified}         & arXiv 2023     & Unified-IO 2  & Speech, Audio, Audio music  \\
      G, U   & M$^2$UGen~\cite{hussain2023m} & arXiv 2023     & Llama 2        & Audio music \\
      G, U   & LauraGPT~\cite{chen2023lauragpt}             & arXiv 2023 & -   & Speech    \\
      U   & LLaSM~\cite{shu2023llasm}  & arXiv 2023     &  Llama 2    & Speech    \\
      G, U   & AudioPaLM~\cite{rubenstein2023audiopalm} & arXiv 2023     & PaLM          & Speech    \\
      U   & Pengi~\cite{deshmukh2023pengi}  & NeurIPS 2023 &       -  &  Speech, Audio, Audio music  \\
      G, U   & Speechgpt~\cite{zhang2023speechgpt} & EMNLP 2023      & LLaMA         & Speech    \\
      G, U   & Sparks~\cite{bubeck2023sparks} & arXiv 2023 & GPT-4         & Symbolic music             \\
      U   & Qwen-Audio~\cite{chu2023qwen-audio}   & arXiv 2023     & Qwen-LM       & Audio, Speech, Audio music  \\
      U   & SALMONN~\cite{tang2023salmonn} & arXiv 2023     & Vicuna        & Audio, Speech, Audio music  \\
      U   & Llark~\cite{gardner2023llark}  & arXiv 2023     &     Llama 2         & Audio music \\
      U   & MU-LLaMA~\cite{liu2024music} & arXiv 2023 & LLaMA & Audio music \\
      U   & Speech-LLaMA~\cite{wu2023decoder}  &  ASRU 2023      & LLaMA         & Speech    \\
      U   & LTU~\cite{gong2023listen} & ICLR 2024      & LLaMA         & Audio     \\
      U   & Yu et al.~\cite{yu2023connecting}     & ICASSP 2024     & Vicuna      & Speech    \\
      
     \hline
    \multicolumn{1}{c}{\textbf{\textit{LLMs inspired backbone}}} & \multicolumn{1}{l}{}   & \multicolumn{1}{l}{} & \multicolumn{1}{l}{} & \multicolumn{1}{l}{} \\ 
      G, E  & UniAudio~\cite{yang2023uniaudio} & arXiv 2023 & -  & Audio, Speech, Audio music  \\
      G  & AudioLM~\cite{borsos2023audiolm} & IEEE/ACM TASLP & - & Audio     \\
      G  & MusicGen~\cite{copet2024musicgen} & NeurIPS 2023 & - & Audio music \\
      G  & Jukebox~\cite{dhariwal2020jukebox} & arXiv 2020 & - & Audio music \\
      G  & MusicLM~\cite{agostinelli2023musiclm} & arXiv 2023 & - & Audio music \\
      G  & VALL-E~\cite{wang2023valle} & arXiv 2023 & - & Speech    \\
      U  & SICL~\cite{wang2023can-Whisper} & arXiv 2023 & - & Speech    \\
      
     \hline
    \multicolumn{1}{c}{\textbf{\textit{LLMs as conditioner}}} & \multicolumn{1}{l}{}   & \multicolumn{1}{l}{} & \multicolumn{1}{l}{} & \multicolumn{1}{l}{} \\      
      G  & TANGO~\cite{ghosal2023text}  & arXiv 2023     & FLAN-T5       & Audio     \\     
      G  & Music ControlNet~\cite{liu2024music} & ICASSP 2024     & ChatGPT       & Audio music \\
      U  & Wu et al.~\cite{wu2024improving}    & ICASSP 2024     & - & Audio     \\
     \hline
     \multicolumn{1}{c}{\textbf{\textit{LLMs as agent}}} & \multicolumn{1}{l}{}   & \multicolumn{1}{l}{} & \multicolumn{1}{l}{} & \multicolumn{1}{l}{} \\  
      G, E     & Loop Copilot~\cite{zhang2023loop} & arXiv 2023     & GPT-4         & Audio music \\
      G, U     & MusicAgent~\cite{yu2023musicagent} & EMNLP (Demos) 2023      &     ChatGPT          & Audio music, Symbolic music \\
      G, U     & Audiogpt~\cite{huang2023audiogpt} & AAAI 2024 & GPT-3.5 & Audio, Speech, Audio music  \\
      G, U     & Hugginggpt~\cite{shen2023hugginggpt}   & NeurIPS 2023      &      ChatGPT         & Audio, Speech, Audio music  \\
      G     & Wavjourney~\cite{liu2023wavjourney}           & arXiv 2023     & ChatGPT       & Audio, Audio music         \\
      G     & ComposerX~\cite{deng2024composerx}           & arXiv 2024     & GPT-4 & Symbolic music         \\

     \hline
     \multicolumn{1}{c}{\textbf{\textit{LLMs as labeller}}} & \multicolumn{1}{l}{}   & \multicolumn{1}{l}{} & \multicolumn{1}{l}{} & \multicolumn{1}{l}{} \\  
      G  & Audiobox~\cite{vyas2023audiobox} & arXiv 2023     & LLAMA2 7B     & Audio, Speech, Audio music  \\      
      G  & Make-An-Audio 2~\cite{huang2023make2} & arXiv 2023 & GPT-3.5 & Audio \\

\bottomrule
    \label{tab:audio_method}
\end{tabular}
\end{table*}

\begin{figure*}[h]
    \centering
    \includegraphics[width=\linewidth]{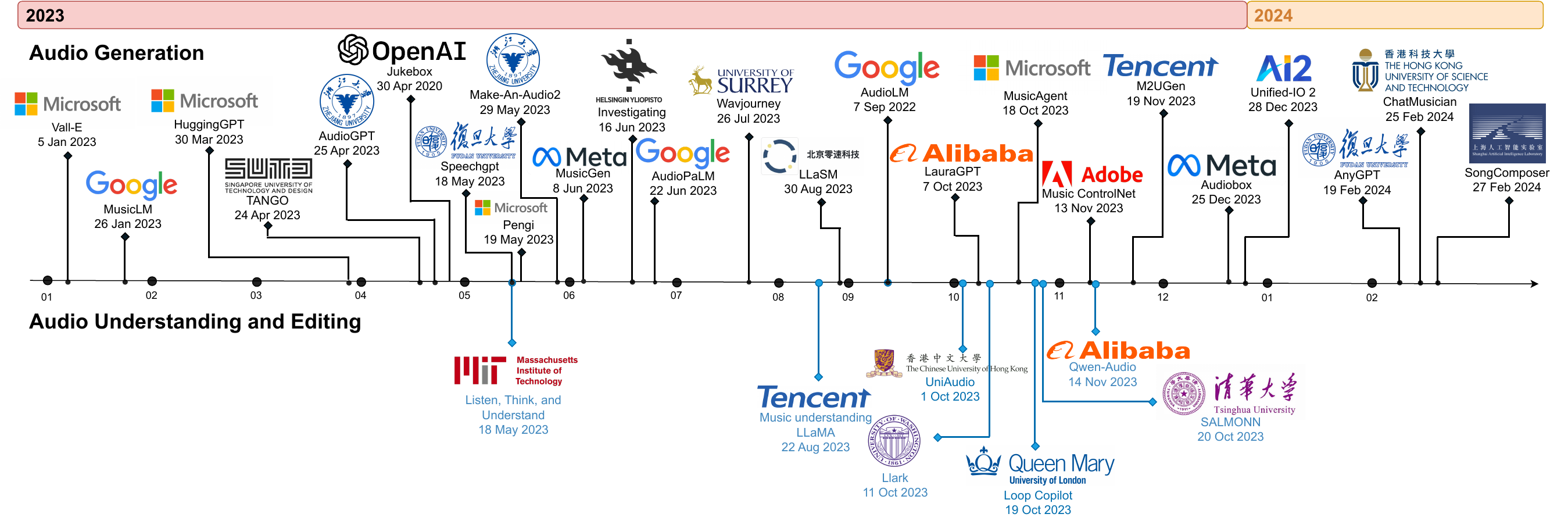}
    \caption{Milestone works of LLMs-based audio research, including audio generation, understanding, and editing.
    }
    \label{fig:Audio_timeline}
\end{figure*}

\begin{figure*}[h]
    \centering
    \includegraphics[width=0.9\linewidth]{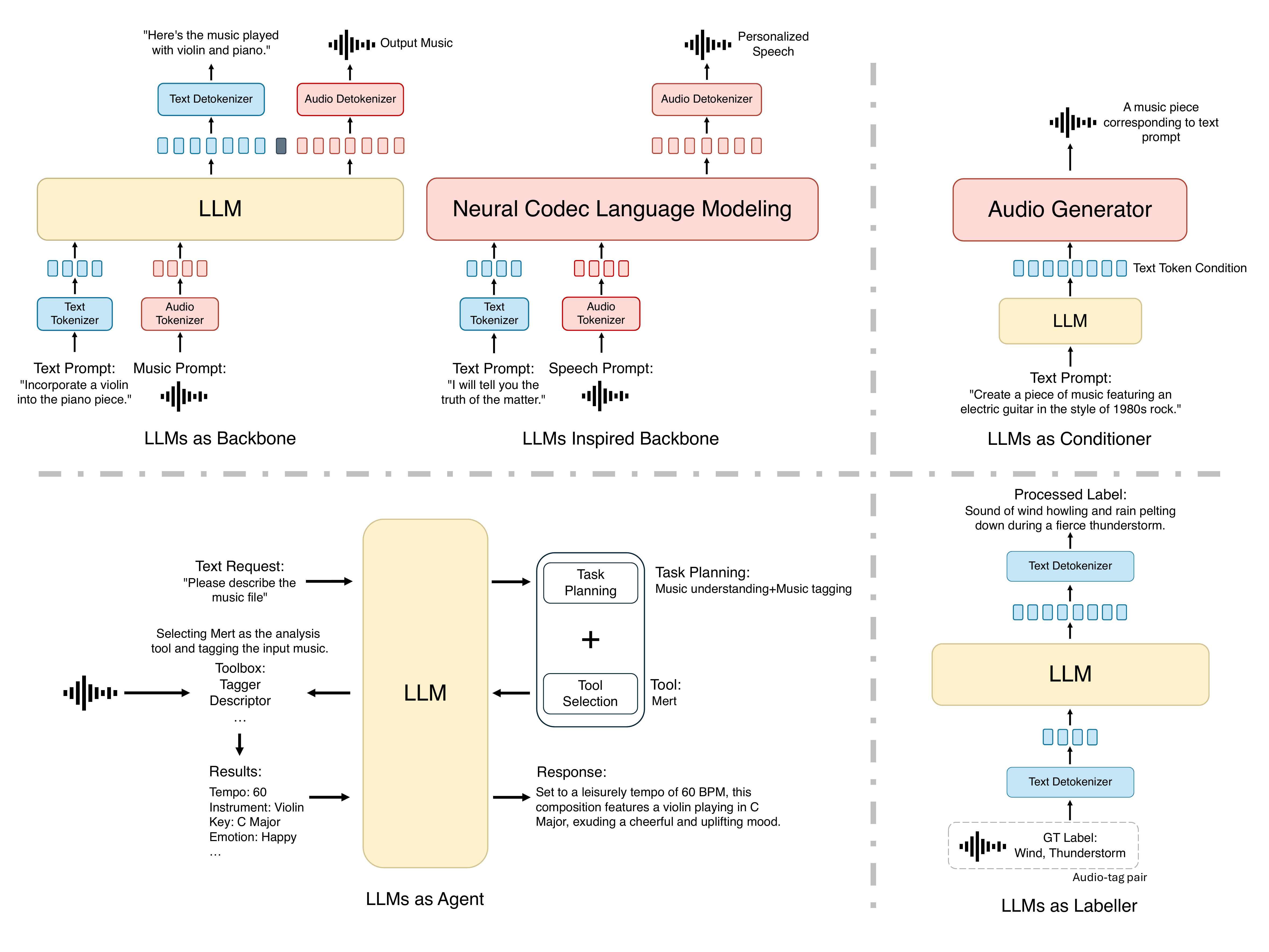}
    \caption{
    Method summary of LLMs-related audio research according to different roles of LLMs.
    LLMs as backbone: Language pre-trained LLMs checkpoint serves as the central unit for processing text and audio tokens, either continuous or discretized. 
    LLMs as inspiration: Different from LLMs as backbone, this method trains on randomly initialized LLMs architecture over discretized audio tokens. 
    LLMs as Conditioner: LLMs encode text prompts into embeddings which serve as the condition of the audio generator. LLMs as agent: LLMs solve user requests by leveraging external tools. LLMs as labeller: LLMs convert class labels into audio captions.
    }
    \label{fig:Audio_methods}
\end{figure*}
\subsection{Domains}

The works that integrate LLMs in the field of audio can be divided into three categories: general audio sounds, music, and speech. Each category poses the challenges and prospects from three pivotal perspectives—generation, understanding, and editing. The summary of the key technical components of LLMs-based approaches for LLMs-related audio tasks is shown in Table~\ref{tab:audio_method}, and the related audio-language dataset in Table~\ref{tab:audio_dataset}. The timeline of milestone works is shown in Fig.~\ref{fig:Audio_timeline}.

\subsubsection{General Audio Sounds}

General Audio Sounds refers to any sound that can be heard. It includes various auditory experiences, including natural sounds (like birds chirping and wind rustling through trees), human activities (such as traffic noise and machinery), and other environmental noises. 
In the past few months, the field of general audio has experienced significant advancements through the application of LLMs~\cite{ghosal2023text, huang2023audiogpt, gong2023listen, huang2023make2, shen2023hugginggpt, borsos2023audiolm, wu2024improving, yang2023uniaudio, tang2023salmonn, chu2023qwen-audio, vyas2023audiobox, lu2023unified}. In the following sections, we will explore the specific areas of audio generation, audio understanding, and audio editing to analyze how these LLMs-driven developments are reshaping the audio domain.

\noindent\textbf{Audio Understanding.}
General audio understanding involves the analysis and interpretation of a wide array of sounds from our environment, beyond just speech and music. This task includes identifying and classifying sounds (such as distinguishing a car horn from a dog bark), recognizing patterns in environmental sounds (like detecting the sound of rainfall or an approaching vehicle), and even understanding the context or source of sounds. 


A suite of groundbreaking models, LTU (Listen, Think, and Understand)~\cite{gong2023listen}, SALMONN (Speech Audio Language Music Open Neural Network)~\cite{tang2023salmonn}, Qwen-Audio~\cite{chu2023qwen-audio}, and UNIFIED-IO 2~\cite{lu2023unified}, have utilized LLMs as their backbone for audio understanding. Unlike LTU~\cite{gong2023listen}, which is the first multimodal LLM to focus on general audio understanding beyond just speech, SALMONN~\cite{tang2023salmonn} is the first multimodal LLM capable of perceiving and understanding general audio inputs with speech, audio events, and music.
By integrating audio with other data modalities, UNIFIED-IO 2~\cite{lu2023unified} leverages LLMs to enhance the understanding of complex interactions between various types of input. 
Qwen-Audio~\cite{chu2023qwen-audio} improves interaction capabilities of pre-trained audio models by covering over 30 different tasks and various audio types, including human speech, natural sounds, music, and songs, thereby promoting comprehensive audio understanding capabilities. 
To improve user interaction, models like AudioGPT~\cite{huang2023audiogpt} and HuggingGPT~\cite{shen2023hugginggpt} also leverage LLMs to serve as intelligent interfaces.
The works~\cite{wu2024improving, gong2023listen, chu2023qwen-audio, lu2023unified, tang2023salmonn} demonstrate how LLMs can be used to enhance automated audio captioning.

LTU~\cite{gong2023listen} combines the audio perception model AST~\cite{gong2021ast} with LLaMA~\cite{touvron2023llama} to improve audio understanding through a perception-to-understanding curriculum. For this purpose, this work also constructs the OpenAQA-5M dataset, which includes 1.9 million closed-ended and 3.7 million open-ended tuples. This dataset facilitates LTU's training within an autoregressive framework.

SALMONN~\cite{tang2023salmonn} processes general audio inputs, including speech, events, and music, by integrating a text-based LLM with speech and audio encoders together. This fusion improves SALMONN's understanding ability across various audio phenomena.

UNIFIED-IO 2~\cite{lu2023unified} is the first autoregressive multimodal model that integrates text, images, audio, and actions into a unified framework. Using a single encoder-decoder transformer model, it tokenizes inputs from different modalities into a shared semantic space for processing.

Qwen-Audio~\cite{chu2023qwen-audio} scales up audio-language pre-training to include over 30 tasks. To address the interference problems that arise from training all tasks and datasets together, a multi-task training framework was designed. This framework uses a sequence of hierarchical tags for the decoder, which helps share knowledge and prevent interference by using both shared and specific tags. Further developed upon Qwen-Audio, Qwen-Audio-Chat can take input from different audio and text sources, allowing for multi-turn conversations and supporting various audio-focused scenarios.

AudioGPT~\cite{huang2023audiogpt} and HuggingGPT~\cite{shen2023hugginggpt} showcase the use of LLMs for audio understanding by coordinating tools through LLMs-driven interfaces. AudioGPT utilizes ChatGPT as a central node for audio and speech applications, depending on external audio systems for functionality. HuggingGPT functions as an agent that combines ChatGPT's language capabilities with a diverse set of AI models from the Hugging Face community, improving its ability to understand audio content.

Wu et al.~\cite{wu2024improving} focuses on advancing automated audio captioning (AAC), a field dedicated to generating descriptive text for sounds from nature and human activities. This work pushes this development further by extensively integrating pretrained models and LLMs. Wu et al. employ BEATS for extracting detailed audio features and use the INSTRUCTOR LLM for obtaining text embeddings of captions. Additionally, Wu et al. introduce a data augmentation technique using ChatGPT to create caption mix-ups and enrich the training data in terms of quantity, complexity, and diversity.

\noindent\textbf{Audio Generation.}
Audio generation is an emerging field that focuses on modeling the creation of diverse audio content. The application of LLMs has significantly advanced the generation of audio. Significant contributions in text-to-audio generation include TANGO~\cite{ghosal2023text}, Make-an-Audio 2~\cite{huang2023make2}, WavJourney~\cite{liu2023wavjourney}, AudioLM~\cite{borsos2023audiolm}, and Audiobox~\cite{vyas2023audiobox}. From employing text embedders and diffusion models in TANGO~\cite{ghosal2023text} and Make-an-Audio 2~\cite{huang2023make2} to integrating multi-modal approaches in WavJourney~\cite{liu2023wavjourney} and advanced tokenization in AudioLM~\cite{borsos2023audiolm} and Audiobox~\cite{vyas2023audiobox}, these initiatives highlight the versatility and impact of LLMs in driving forward the capabilities of audio generation technologies.

TANGO employs FLAN-T5~\cite{chung2022scaling} as a text embedder, while Make-an-Audio 2 uses pre-trained LLMs for text parsing into structured pairs, both utilizing latent diffusion-based models for audio synthesis. 
WavJourney~\cite{liu2023wavjourney} leverages the LLM agent to integrate various audio models for producing cohesive audio content, including speech, music, and sound effects, based on textual descriptions.
AudioLM~\cite{borsos2023audiolm} generates high-quality audio with an emphasis on long-term consistency. The method converts input audio into discrete tokens, effectively making audio generation a task akin to language modeling within a discrete space.
Audiobox~\cite{vyas2023audiobox} uses LLMs for data construction, including tagging audio with high-quality, detailed captions and using LLMs to evaluate the quality of these annotations automatically. Then, Audiobox uses flow-matching techniques to produce diverse audio types with precise attribute control, from speech to music and sound effects.
UniAudio~\cite{yang2023uniaudio} introduces a versatile system utilizing LLMs to generate a diverse range of audio types, including speech, sounds, music, and singing, under various input conditions. Unlike task-specific models, UniAudio tokenizes different audio types and their respective conditions into a unified sequence, enabling next-token prediction with LLMs.

\subsubsection{Music}
Music is an art form characterized by the arrangement of sounds in time, typically including elements such as melody, harmony, rhythm, and timbre. It is produced using musical instruments and/or the human voice and is often organized according to pitch (which influences melody and harmony), rhythm (including tempo, meter, and articulation), dynamics (variations in loudness), and the sonic properties of timbre and texture. Music serves multiple functions, including aesthetic enjoyment, ceremonial purposes, and the expression of cultural identity. 
In the realm of music research, audio music refers to the actual recorded sound waves of music, while symbolic music pertains to the notational representation of music, such as MIDI~\cite{rothstein1995midi} files. Each of these forms requires distinct approaches for analysis and manipulation.
We explore the interconnected fields of music understanding, generation, and editing, each of which utilizes different techniques and technologies to analyze, create, and refine music, further enriching its cultural and artistic impact.

\noindent\textbf{Music Understanding.}
Music understanding involves the analysis and interpretation of musical elements such as melody, harmony, rhythm, and timbre, to recognize patterns, genres, emotions, and contextual meanings within music. It includes the analysis of simple motifs to complex structures.

The field of music understanding has made significant progress with the development of models like the Music Understanding LLaMA (MU-LLaMA)~\cite{liu2024music}, LLARK~\cite{gardner2023llark} MusicAgent~\cite{yu2023musicagent}, LyricWhiz~\cite{zhuo2023lyricwhiz}, and ChatMusician~\cite{yuan2024chatmusician}. These models showcase a range of methods, from analyzing detailed music features to improving lyric transcription. They highlight how LLMs can assist our understanding and interaction with music in different ways and applications.

MU-LLaMA~\cite{liu2024music} uses a pretrained MERT encoder for initial music representation, which is then integrated into the LLaMA model with an adapter. This process leverages the capabilities of LLMs to understand music by analyzing its comprehensive features.
LLARK~\cite{gardner2023llark} improves music understanding by using a multimodal model that is fine-tuned with instructions from refined annotations in music datasets. It combines a generative music model with a language model to analyze music in a unified way.
MusicAgent~\cite{yu2023musicagent} assists music understanding and generation through LLMs, automating tasks to meet user needs and using tools for task execution. This simplifies processes and encourages exploration in music processing.
In summary, MU-LLaMA focuses on the analysis of musical features, LLARK leverages refined labels for a more general understanding, while MusicAgent emphasizes user-friendly interactions.

LyricWhiz~\cite{zhuo2023lyricwhiz} proposes a multilingual, zero-shot method for automatic lyrics transcription, performing well across various genres. It uses ``Whisper'' for speech recognition and ``GPT-4'' for context-aware annotations, acting as the transcription's ``ear'' and ``brain''. This combination greatly reduces the Word Error Rate in English and provides effective transcription in multiple languages.

\noindent\textbf{Music Generation.}
AI music generation, especially with the use of LLMs, is changing the industry by creating various and complex musical pieces. Notable examples include MusicLM~\cite{agostinelli2023musiclm}, Jukebox~\cite{dhariwal2020jukebox}, MusicGen~\cite{copet2024musicgen}, Music ControlNet~\cite{wu2023music}, M$^2$UGen~\cite{hussain2023m}, ChatMusician~\cite{yuan2024chatmusician}, and SongComposer~\cite{ding2024songcomposer}. Specifically, these models use different techniques, from using LLMs as text embedders to employing diffusion processes and autoregressive Transformers. 

MusicLM~\cite{agostinelli2023musiclm}, Jukebox~\cite{dhariwal2020jukebox}, and MusicGen~\cite{copet2024musicgen} represent significant strides in text-to-music generation, each drawing inspiration from the capabilities of LLMs and employing Transformer architectures to handle complex audio tasks. MusicLM~\cite{agostinelli2023musiclm} treats conditional music generation as a hierarchical sequence-to-sequence task, utilizing decoder-only Transformers to create music in both the semantic and acoustic stages.
Jukebox~\cite{dhariwal2020jukebox} addresses the challenges of long audio contexts by compressing raw audio into discrete codes using a multiscale VQ-VAE and then modeling these codes with autoregressive Transformers.

On the other hand, MusicGen~\cite{copet2024musicgen} directly incorporates an LLM as a text embedder. It combines text tokens and melody conditions into a Transformer decoder, which then processes the inputs in an autoregressive way. The final step involves a codec model that converts the processed tokens back into music.

Music ControlNet~\cite{wu2023music} uses an LLM as a text embedder and introduces a diffusion-based model for music generation that offers precise control over dynamic and temporal aspects of audio. Inspired by image-domain ControlNet's pixel-wise control, it applies similar precision to audio through controls for melody, dynamics, and rhythm derived from training audio.

M$^2$UGen~\cite{hussain2023m} introduces a multi-modal music understanding and generation framework that leverages LLMs along with pretrained models like MERT, ViT, and ViViT to analyze and create music from diverse inputs such as music, images, and videos. The decoder part utilizes AudioLDM 2~\cite{liu2023audioldm2} and MusicGen~\cite{copet2024musicgen} for generating music. 

ChatMusician~\cite{yuan2024chatmusician} and SongComposer~\cite{ding2024songcomposer} both focus on generating symbolic music but use different methods and representations. ChatMusician is an open-source LLM with intrinsic musical abilities, using continual pre-training and fine-tuning of Llama2 with ABC notation, a text-compatible music representation. It treats music as a second language, allowing it to understand and generate music using only a text tokenizer, without the need for external multi-modal neural structures.
In contrast, SongComposer uses MIDI for its symbolic music representation and introduces a unique tuple design to format lyrics alongside three note attributes: pitch, duration, and rest duration. This design ensures the correct interpretation of musical symbols and precise alignment between lyrics and melodies, differentiating it from ChatMusician's approach.

Music editing involves refining and altering musical elements to improve the sound quality and artistic expression. 
Loop Copilot~\cite{zhang2023loop} combines LLMs with specialized AI music models to create a conversational interface for collaborative human-AI music loop creation. It uses a large language model to interpret user intentions and directs specialized AI models to generate and refine music through interactive dialogue. Key musical attributes are centralized and managed to ensure consistent quality throughout the creative process.

\subsubsection{Speech}
Speech specifically refers to the sounds that humans produce when they speak. It is the verbal manifestation of language, including a variety of linguistic elements such as words, sentences, tone, intonation, and rhythm. Speech is a fundamental mode of human communication and can vary greatly depending on factors like language, dialect, emotional state, and context. In the realm of artificial intelligence, LLMs have been advancing both the understanding and generation of speech, facilitating machines to interpret and replicate human-like spoken communication with increasing accuracy and naturalness.

\noindent\textbf{Speech Understanding.}
Speech understanding empowers machines to interpret spoken language. This aspect of AI captures not just the words but also the speaker's intent and nuances, with progress driven by LLMs. Key contributions in this field include SpeechGPT~\cite{zhang2023speechgpt}, AudioPaLM~\cite{rubenstein2023audiopalm}, and other studies~\cite{wu2023decoder, wang2023can-Whisper, yu2023connecting}, showcasing the improved capabilities of LLMs in recognizing and processing speech across diverse contexts.

SpeechGPT~\cite{zhang2023speechgpt}, AudioPaLM~\cite{rubenstein2023audiopalm}, and Speech-LLaMA~\cite{wu2023decoder} represent pivotal developments in speech understanding, all utilizing LLMs as the structural backbone of their frameworks.
SpeechGPT~\cite{zhang2023speechgpt} and Speech-LLaMA~\cite{wu2023decoder} specifically use LLaMA as their foundation. SpeechGPT not only facilitates understanding and generating multi-modal content but also supports inter-modal knowledge transfer. It introduces SpeechInstruct, a large-scale cross-modal speech instruction dataset built upon discrete speech representations, highlighting its multi-modal capabilities. Meanwhile, Speech-LLaMA integrates speech signals with LLMs, emphasizing a mixture of auditory and linguistic data processing.
Similarly, AudioPaLM~\cite{rubenstein2023audiopalm} combines the strengths of PaLM-2~\cite{anil2023palm} and AudioLM~\cite{borsos2023audiolm} into a unified multimodal framework that performs well in both speech understanding and generation. It retains paralinguistic features such as speaker identity and intonation from AudioLM and blends them with the textual linguistic capabilities of PaLM-2, demonstrating an approach to multimodal speech processing.

Recent research in Automatic Speech Recognition (ASR) has focused on improving model accuracy with LLMs.
Wang et al.~\cite{wang2023can-Whisper} investigate the in-context learning abilities of the Whisper~\cite{radford2023robust}, which is an ASR model released by OpenAI. SICL~\cite{wang2023can-Whisper} is introduced to reduce the Word Error Rates (WERs) with only a small number of labeled speech samples without gradient descent.
Yu et al.~\cite{yu2023connecting} presents a study of structures that include fully connected layers, multi-head cross-attention, and Q-Former as connectors for integrating ASR models with LLMs.

\noindent\textbf{Speech Generation.}
Speech generation, the process of converting text or other inputs like speech prompts into spoken language, has significantly evolved with the integration of LLMs. These models facilitate the naturalness and contextual relevance of generated speech, making it increasingly realistic and similar to human speech.

Inspired by the capabilities of LLMs, Wang et al. introduced VALL-E~\cite{wang2023valle}, a transformative approach in speech generation. VALL-E utilizes a neural codec language model that employs discrete codes from an existing neural audio codec, reframing text-to-speech (TTS) synthesis as a conditional language modeling task rather than traditional continuous signal regression. Building on this VALL-E, Hao et al. further advanced the field with their study~\cite{hao2023boosting}. They conducted an investigation to improve LLMs' speech generation capabilities by integrating the pre-trained LLM frameworks, LLaMA/OPT, with the TTS model VALL-E. This research showcases a combination of language modeling and speech synthesis technology, aiming to produce more natural and effective speech outputs.

Different from previous works, using an LLM as the backbone, LauraGPT~\cite{chen2023lauragpt}, developed by Wang et al., is a unified GPT model capable of handling both audio and text for recognition, understanding, and generation tasks. It performs well in a variety of functions including speech recognition, translation, text-to-speech synthesis, and more.

Another work proposed by Kakouros et al.~\cite{kakouros2023investigating} examines the potential of word surprisal, a measure of word predictability in context, to improve speech synthesis prosody.

\subsection{Roles of LLMs}

Language provides a great abstraction of our world.
With the flexibility and rich descriptive power of language, researchers have unified language understanding and language generation, into a paradigm so-called generative understanding. 
Audio research benefits a lot from language in the LLM era, with LLMs serving as a bridge to collect and process information, the field is now able to reach a similar generative understanding stage. 
Broadly, we categorize the roles of LLMs into the following: LLMs as Backbone, LLMs as Conditioner, LLMs as Labeller, LLMs as Agent, and LLMs Inspired Backbone. The method summary is shown in Fig.~\ref{fig:Audio_methods}.

\subsubsection{LLMs as Backbone}
Using an LLM as a backbone entails leveraging a pre-trained LLM, like LLaMA, as the central architecture of a system. These backbones, essential for the system’s learning and processing abilities, are integrated with various network components and undergo fine-tuning. In the realm of multimodal LLM applications within the audio domain, the LLM backbone plays a pivotal role. It is either coupled with structures specialized for modality-specific understanding or generation, or it employs a tokenizer to convert audio into discrete tokens.

Much of the current research in LLMs typically adopts a cascade approach, which often involves the use of modality-specific encoders and/or decoders. In LTU~\cite{gong2023listen}, Yuan et al. suggest utilizing an Audio Spectrogram Transformer (AST)~\cite{gong2021ast}, a transformer encoder pre-trained with CAV-MAE~\cite{gong2022contrastive}. This encoder's representations are aggregated and input into a LLaMA-7B backbone. These audio pre-trained representations are then paired with corresponding texts. To fine-tune the LLM backbone, LoRA adapters~\cite{hu2022lora} are employed, tasked with predicting text pairs based on the audio representations.
Pengi~\cite{deshmukh2023pengi} follows a similar paradigm. Soham et al. call it a `audio-text-to-text` format. Except for an CLAP~\cite{elizalde2023clap} audio encoder, it also uses a text encoder to encode the task instructions. The audio representation, as well as the text instruction representation, are together fed into the LLM as a prefix. The LLM is then trained to predict the paired text output, e.g. a sound description. Both LTU and Pengi show improvement in close-ended audio understanding tasks and a certain level of open-ended audio understanding tasks.
Similar approaches can also be found in LLaSM~\cite{shu2023llasm}, Mu-LLaMA~\cite{liu2024music}, MusicLingo~\cite{deng2023musilingo}, Llark~\cite{gardner2023llark}, Qwen-Audio~\cite{chu2023qwen-audio}. Popular audio encoder may include CLAP~\cite{elizalde2023clap}, MERT~\cite{li2023mert}, Whisper~\cite{wang2023can-Whisper}, AST~\cite{gong2021ast}.

Besides the model focused solely on understanding, there has also been researches extending into generation. Part of these researches adopt the design philosophy of the cascade approach, incorporating not just an audio encoder but also introducing an audio decoder. For instance, in M$^2$UGen~\cite{hussain2023m}, Atin et al. adapt both a MERT encoder and an Audioldm2~\cite{liu2023audioldm2} / MusicGen~\cite{copet2024musicgen} decoder. The output projection layer then maps the LLaMA2 model's output embeddings to the music decoder. A similar approach is observed in NExT-GPT~\cite{wu2023next}, a recently proposed any-to-any multimodal language model. 

However, the cascade approach requires training heterogeneous neural structures. In scenarios with abundant data and computational resources, these heterogeneous neural structures could lead to decreased training efficiency and lower system scalability. Recently, a unified approach has garnered the attention of researchers. This method typically necessitates the use of audio codecs~\cite{defossez2022high, kumar2024high}, to tokenize raw audio into discrete tokens, which are then flattened into a one-dimensional sequence for input into an LLM. This requires the LLM's vocabulary to include audio tokens, thereby necessitating an expansion of the LLM vocabulary, akin to integrating audio as a new language into the LLM. This method employs a uniform LLM structure, facilitating scalability. AudioPaLM~\cite{rubenstein2023audiopalm}, LauraGPT~\cite{chen2023lauragpt}, SpeechGPT~\cite{zhang2023speechgpt} follows this paradigm.

\subsubsection{LLMs as Conditioner}
In this setting, LLMs usually act as text embedders, encoding input text to condition the system’s response or output, thus enabling a more nuanced and context-aware processing of audio data. 

Tango~\cite{ghosal2023text} follows this paradigm. It comprises three main components: a text encoder, a Latent Diffusion Model (LDM), and a Mel-Spectrogram/Audio Variational Autoencoder (VAE). The text encoder is a Flan-T5, which translates the audio's input text prompt into a textual representation. This representation is then utilized to construct a latent audio representation or audio prior from standard Gaussian noise through reverse diffusion. Following this, the Mel-Spectrogram VAE's decoder generates a Mel-Spectrogram from the latent audio representation. Finally, this Mel-Spectrogram is input into a vocoder to produce the final audio output. MusicGen~\cite{copet2024musicgen} follows a similar paradigm. The study tested both T5 and Flan-T5 models along with CLAP, and found that the T5 encoder, serving as a text conditioner, achieved the highest relevance subjective test score in relation to the text input.

\subsubsection{LLMs as Labeller}
Currently, the majority of large-scale audio datasets, such as AudioSet~\cite{gemmeke2017audio} and VGGSound~\cite{chen2020vggsound}, are annotated solely with class labels, akin to ImageNet~\cite{deng2009imagenet}. Researchers aiming to undertake text-to-audio tasks are compelled to transform these class labels into full-sentence audio descriptions, also known as audio captioning. A prevalent method involves utilizing LLMs to achieve this transformation.

A common pipeline for text description augmentation involves initially crafting description templates manually for a labeled audio dataset, thereby parsing audio classes into more uniformly formatted descriptions. Subsequent steps leverage self-instruction methods, employing LLMs such as ChatGPT~\cite{chatgpt} that are capable of following instructions to paraphrase these descriptions, often making use of self-instruct~\cite{wang2022self} techniques to enrich the dataset further.

\subsubsection{LLMs as Agent}
In `LLMs as Agent', LLMs are employed to interface with various tools, orchestrating multiple functionalities to accomplish diverse tasks. This role highlights the versatility of LLMs in managing and executing complex, multi-dimensional operations.

Communicating with LLMs can be approached in various ways. A notably straightforward yet effective method is through a text interface. During the nascent phase of multi-modal LLM audio research, Huang et al. introduced AudioGPT. This system, capitalizing on advanced audio foundation models, tackles tasks such as sound detection, audio-to-text conversion, speech recognition, and speech translation. The data gleaned from these audio processes is then transformed into text, seamlessly integrating with LLM interactions. Within this framework, task analysis, model assignment, and response generation all function through textual operations. 
AudioGPT drew inspiration from its contemporary work, HuggingGPT, which employs a similar approach. HuggingGPT uses LLMs to invoke various models on Hugging Face, a platform hosting a diverse array of machine learning models. 
Similarly, MusicAgent is proposed to streamline AI-powered music processing by integrating a variety of tools and an autonomous workflow, primarily facilitated by LLMs like ChatGPT. It features a diverse toolset sourced from platforms like Hugging Face, GitHub, and various web APIs. Tasks like music classification, music separation, lyric recognition are supported.

\subsubsection{LLMs Inspired Backbone}

With the success of the next token prediction paradigm in language modeling, the audio domain has also sought to apply this approach by discretizing audio into tokens for modeling. Researchers aim to achieve emergent capabilities on audio tokens similar to those observed in LLMs, such as in-context learning and the chain of thoughts ability. Currently, modest in-context learning abilities have been confirmed to be attainable through the language modeling pretraining of audio tokens. 

In VALL-E\cite{wang2023valle}, researchers combined autoregressive and non-autoregressive language models to model encoded tokens. Thanks to the residual vector quantized (RVQ) modeling\cite{defossez2022high} of acoustic information, VALL-E can continue speech with only short audio and text prompts, preserving the speaker’s timbre, prosody, and acoustic environment while following text constraints. In AudioLM\cite{borsos2023audiolm}, researchers discovered that unconditional training on RVQ-based acoustic tokens did not yield semantic-level consistency. Consequently, they proposed introducing semantic tokens based on Self-Supervised Learning (SSL) representations. The representations from an SSL-pretrained teacher contain rich semantic information, and performing k-means clustering on these representations yields a k-means quantizer, allowing for the extraction of semantic tokens for the training set. Language modeling on these semantic tokens achieves better semantic consistency, enabling unconditional speech continuation to maintain semantic coherence.

\section{Tool-augmented Multimodal Agents}
\label{sec: agent}
\begin{figure*}[h]
    \centering
    \includegraphics[width=\linewidth]{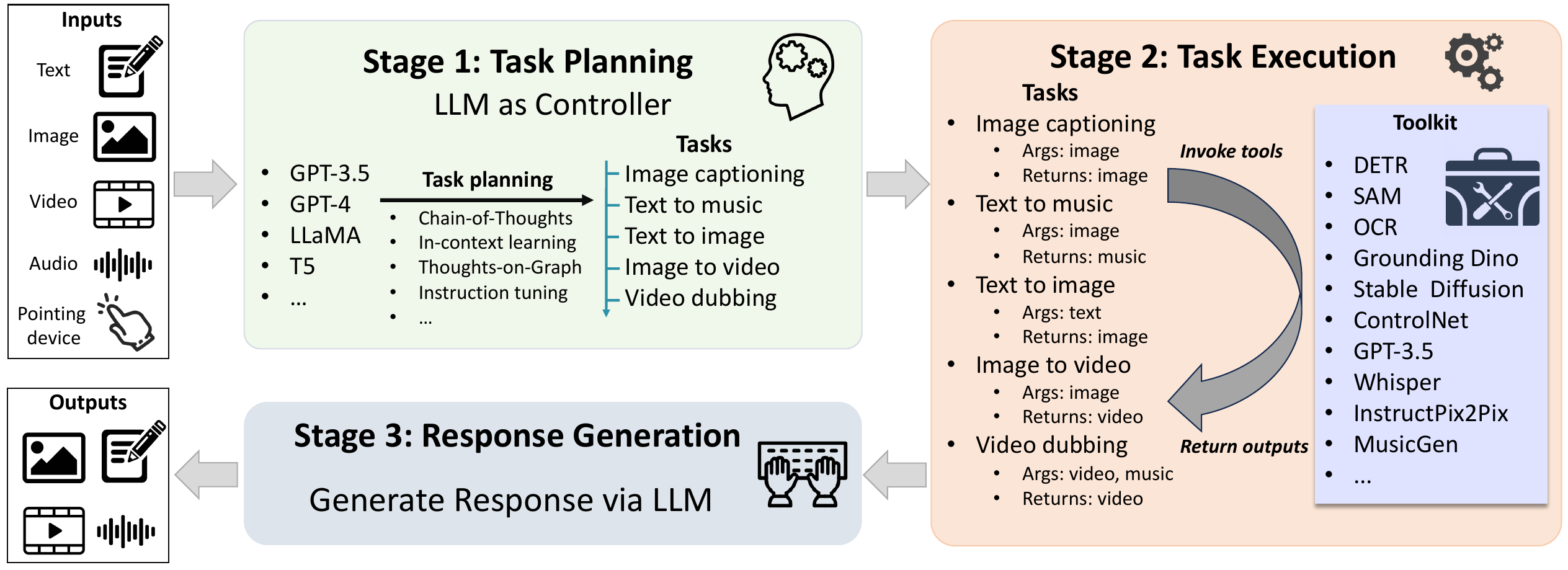}
    \caption{\centering The pipeline of tool-augmented multimodal agents.
    }
    \label{fig:pipeline_of_mm_agents}
\end{figure*}

\begin{figure*}[t]
    \centering
    \includegraphics[width=0.9\linewidth]{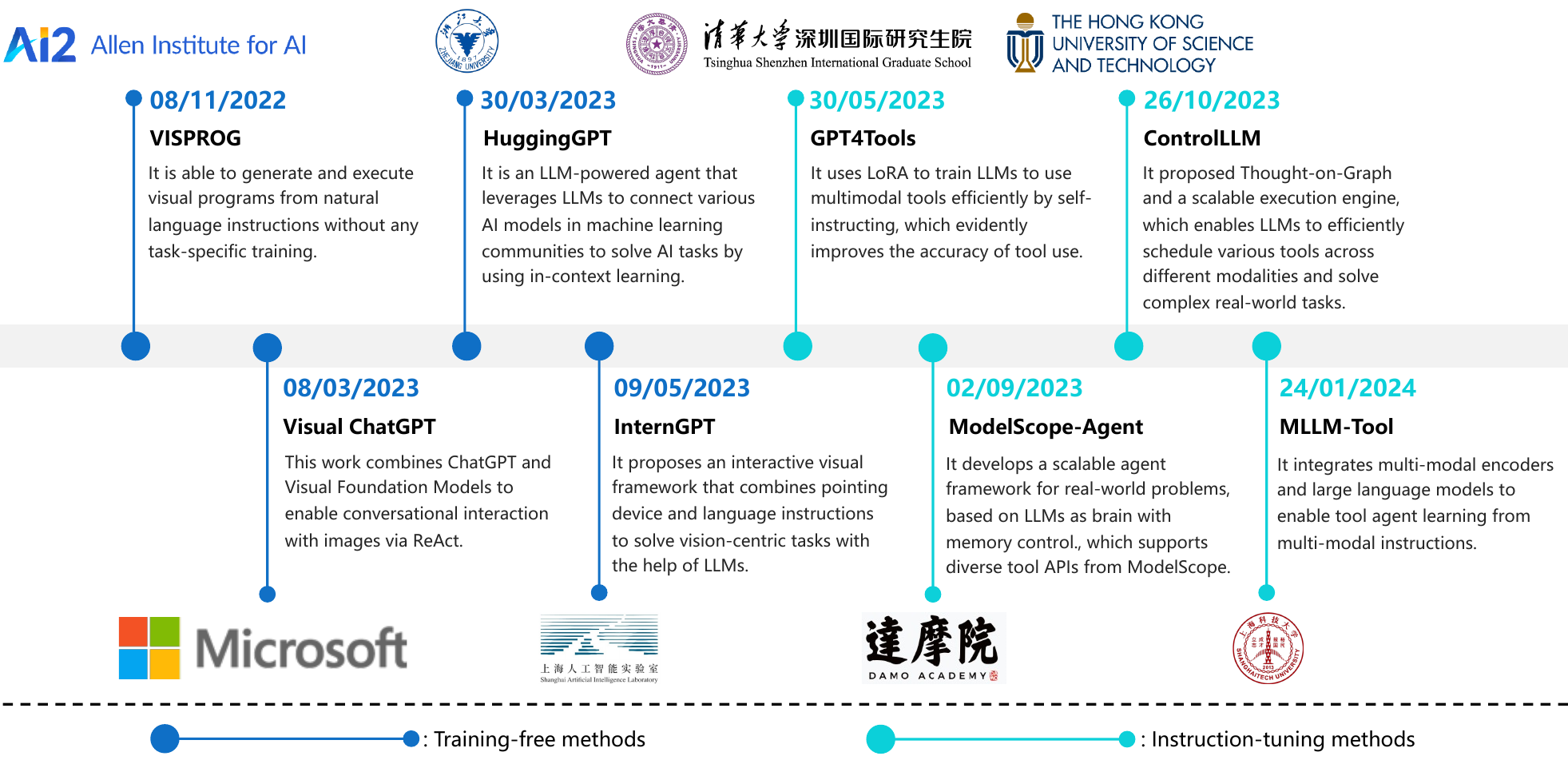}
    \caption{\centering The milestone of multimodal agents that focus on multimodal generation and editing. 
    }
    \label{fig:milestone_of_mm_agents}
\end{figure*}

Over the past few months, lots of works~\cite{qin2023toolllm,tang2023toolalpaca,schick2023toolformer,shen2023hugginggpt,liu2023internchat,yang2023gpt4tools,farn2023tooltalk,hao2023toolkengpt,hsieh2023tool,ruan2023tptu,liu2023controlllm,parisi2022talm,zhang2023graph,zhuang2023toolchain,gou2023critic,jin2023genegpt,paranjape2023art,gou2023tora,song2023restgpt,qiao2023making,zhang2023syntax,wu2023visual,shen2024small,wang2024mobile} known as tool-augmented LLMs have been emerged as a promising direction in human-computer interaction. They empower LLMs to use external tools to enhance the models' capabilities. Among them, several works~\cite{shen2023hugginggpt,wu2023visual,liu2023internchat,yang2023gpt4tools,li2023modelscope,liu2023controlllm} which extend LLMs to other modalities beyond the natural language stand out.
In contrast to those expert models which only focus on optimizing one specific task, \eg., image generation or video generation, these pioneering works built upon LLMs can interactively generate or edit images, videos, and audio by invoking corresponding tools. In this section, we mainly focus on reviewing the recent works that aim to extend LLMs to multimodal generation by augmenting them with external tools.

\subsection{Motivation}
It is known that LLMs have limitations in accessing and processing information that is not available in their training data, such as ephemeral, changing, or private data. For example, LLMs may not be able to answer questions that involve factual knowledge updated frequently, such as the current weather or stock prices.

To overcome these limitations, several works have proposed to augment LLMs with external tools or APIs, such as retrieval-augmented generation (RAG), calculator or visual foundation models, that can provide additional information or functionality for LLMs. These tools can be invoked by LLMs through natural language instructions, and the results can be integrated into the LLMs’ outputs. For example, an LLM can use a weather API to obtain the current temperature and humidity of a given location, and use them to generate a natural language response. The tool-augmented paradigm have been validated the efficacy by many works~\cite{qin2023toolllm,tang2023toolalpaca}. In practice, Microsoft Copilot\footnote{\url{https://www.microsoft.com/en/microsoft-copilot}} augmented by various tools has been integrated into their applications including Bing, Edge and Windows operating system, which dramatically facilitates user experience. OpenAI also releases a Function Calling\footnote{\url{https://platform.openai.com/docs/assistants/tools}} service that can give assistants access to OpenAI-hosted tools like Code Interpreter and Knowledge Retrieval or build your own tools.  

It is well known that LLMs can not generate or edit content in other modalities, such as images, videos, or audio, which is helpful for creative purposes. Motivated by tool-augmented LLMs, some pioneers develop multimodal agents that can control diverse tools across different modalities like image, video and audio. By augmenting LLMs with external tools, it can enable more natural and versatile human-computer interaction, as well as more powerful and creative applications.

\subsection{Methods}
As depicted in Fig.~\ref{fig:pipeline_of_mm_agents}, the general framework of tool-augmented LLMs for multimodal interaction consists of three main stages: 1) \textbf{Task Planning} which tasks LLM as a controller to interpret the natural language instructions into a tool invocation scheme. Specifically, the core aim of this stage is to decide which tools will be used and prepare the arguments for tools. In this stage, the selected tools are organized into the tool invocation scheme that specifies the sequence of tool invocation and the inputs of tools. 2) \textbf{Task Execution} which hosts lots of external multimodal tools, \eg, image generation, video editing, or audio synthesis. The tools are invoked based on the invocation scheme obtained in task planning. It is noteworthy that most of the tools are based on deep learning models including stable diffusion~\cite{ldm}, ControlNet~\cite{zhang2023controllable}, Blip~\cite{li2023blip}, LLaVA~\cite{chen2023llava}, \etc. 3) \textbf{Response Generation} which can make a user-friendly response by prompting the LLM with execution outputs from task execution. The overall system connects the LLM and external multimodal tools, which not only enhances the capabilities of LLM but also evidently improves the user experience.

The main difference among the existing works lies in how they perform task planning. To this end, the tool-augmented LLMs for multimodal interaction can be roughly divided into two categories: (1) training-free methods~\cite{wu2023visual,shen2023hugginggpt,liu2023internchat,gupta2023visual}, and (2) instruction-tuning methods~\cite{yang2023gpt4tools,liu2023controlllm,li2023modelscope,wang2024mllmtool}. The evolution path is shown in Fig.~\ref{fig:milestone_of_mm_agents}. In addition, Table~\ref{tab:feature_comparison} wraps up works related to multimodal agents. Next, we will elaborate on the details of two types of methods.

\subsubsection{Training-free Methods}
Training-free methods~\cite{gupta2023visual,surismenon2023vipergpt,shen2023hugginggpt,wu2023visual,liu2023internchat,gao2023clova} primarily reply on prompt engineering, \eg, ReAct~\cite{yao2023react}, in-context learning~\cite{wei2022chain}, to boost the reasoning capabilities of LLMs. These methods exhibit their effectiveness on many tasks.

Gupta \etal~\cite{gupta2023visual} and Sur\'is \etal~\cite{surismenon2023vipergpt} propose a code generation-based method for solving complex and compositional visual tasks given natural language instructions. These two methods mainly focus on image perception and visual question answering. Taking~\cite{gupta2023visual} as an example, they develop the framework of VISPROG, which consists of two main components: a program generator and a program executor. The program generator uses a large language model (GPT-3) to generate python-like modular programs from natural language instructions. The authors prompt GPT-3 with pairs of instructions and the desired high-level programs, along with a new instruction. GPT-3 then generates a program that can be executed on the input image(s) to perform the described task. Each line of the program invokes one of the 20 modules supported by VISPROG, such as object detection, segmentation, image editing, knowledge retrieval, etc. Then, the program executor executes the generated program on the input image(s) and produces the output and a visual rationale. The executor steps through the program line-by-line and invokes the correct module with the specified inputs. The modules are implemented as python classes that use off-the-shelf computer vision models, image-processing routines, or Python functions. The executor also updates the program state with the output variable name and value after each step. It is noticed that VISPROG does not have stage 3 explicitly, and directly returns the output from tools to users. This work opens up a new direction for multimodal human-computer interaction and inspires the following studies.

In contrast to VISPROG~\cite{gupta2023visual,surismenon2023vipergpt}, Visual ChatGPT~\cite{wu2023visual} does not generate code, directly. It combines a large language model (ChatGPT) with various visual foundation models (VFMs) to enable conversational interaction with both text and images. To achieve this, Wu \etal~\cite{wu2023visual} design a Prompt Manager that bridges the gap between ChatGPT and VFMs. The Prompt Manager converts all non-language signals into language prompts that ChatGPT can understand and process. Essentially, Visual ChatGPT performs task planning via ReAct~\cite{yao2023react}, which is the simplest and most straightforward way to augment LLMs with tools. Instead of generating a full solution containing all related tools once, ReAct, which extends Chain-of-Thought to tool use, immediately executes the tool after each thought. It means that the system carries out stage 1 and stage 2 alternately. In addition, to accurately call the tools, each tool is equipped with a crafted natural language prompt that instructs the LLMs how to use it. In practice, one can use a prompt like ``Please use the tool named ImageBind to generate an image of a cat wearing a hat'' to invoke an image generation tool, and expect the system to return an image path as the response. 
This approach does not require any modification or retraining of the LLMs, and can leverage the existing pre-trained LLMs, such as GPT-3 or ChatGPT. InternGPT~\cite{liu2023internchat} shares a similar pipeline to Visual ChatGPT, but it supports pointing devices. Thus, InternGPT offers more interesting and diverse modes of interaction, including clicking and drawing. For example, After the clicking operation is triggered, InternGPT utilizes SAM~\cite{kirillov2023segment} to select the chosen semantic region, which can be used to remove or replace the object. Furthermore, InternGPT supports user directly in drawing the sketch based on which it can generate a new image.

\renewcommand{\arraystretch}{1.2}
\begin{table*}[!tbh]
    \centering
    \caption{\textbf{The instruction samples of multimodal agents, which are used to train or assess models.}}
    \begin{tabular}{|m{17cm}|}
       \Xhline{1.5pt}
        \multicolumn{1}{|C{17cm}|}{\textbf{Image}}\\
        \hline
        1. Can you remove the dog in the image\_1.png? \\
        2. Can you create a new image that depicts a family having a picnic in the foreground? \\
        3. Generate a new image based on a segmentation map named seg.png. The new image should show a beautiful landscape. \\
        4. Using the pose map pose.png, generate a new image that features a cozy cabin in the woods with a campfire, trees, and a couple roasting marshmallows. \\
        5. Using the segmentation from the given image, generate a new image that showcases a magical forest with glowing mushrooms and fairies flying around. \\
        \hline
        \multicolumn{1}{|C{17cm}|}{\textbf{Video}}\\
        \hline
        1. Can you remove the dog in the video\_1.mp4? \\
        2. Can you provide a video that is related to the given image? \\
        3. Please give me a new video derived from this file image\_2.png? \\
        4. Make a video that illustrates a calm landscape of a lake surrounded by vegetation. \\
        5. I would appreciate it if  you can dub the video\_1.mp4 with given audio file named aud\_1.wav. \\
        \hline
        \multicolumn{1}{|C{17cm}|}{\textbf{Audio}}\\
        \hline
        1. Can you create a song that visually represents the image? \\
        2. You need to generate a piece of background music for video\_3.mp4. \\
        3. Please generate a piece of music by given prompt: An 80s driving pop song with heavy drums in the background.\\
        4. I want to create a piece of background music for video\_4.mp4 and dub the video with this piece of music. \\
        5. translate the text into speech: ``Hope is the thing with feathers That perches in the soul, And sings the tune without the words, And never stops at all''. \\
        \Xhline{1.5pt}
    \end{tabular}
    \label{tab:instructions}
\end{table*}

\begin{figure*}[h]
    \centering
    \includegraphics[width=0.9\linewidth]{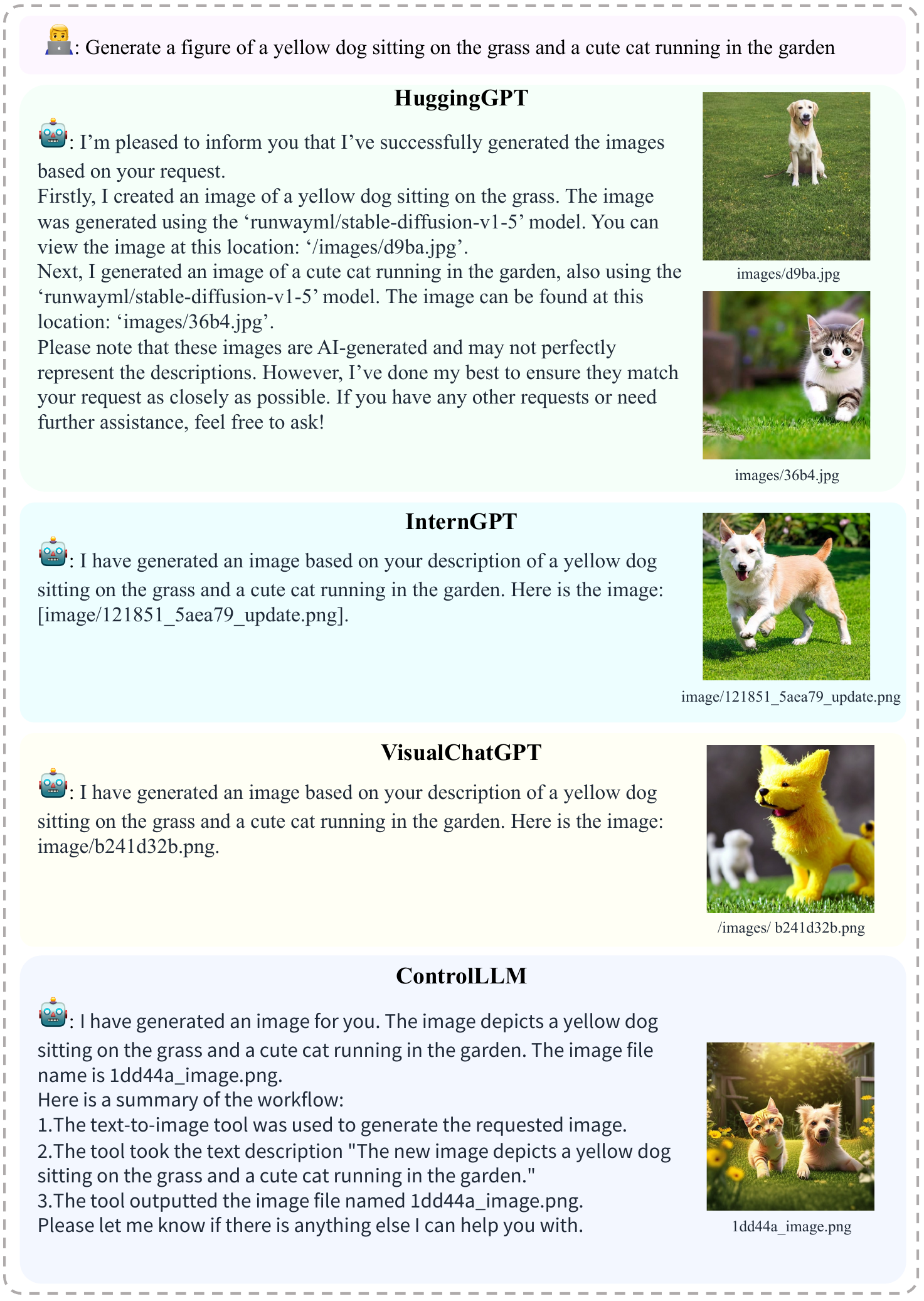}
    \caption{\centering We show some case studies on image generation for different methods.
    }
    \label{fig:agent_vis_1}
\end{figure*}

\begin{figure*}[h]
    \centering
    \includegraphics[width=0.9\linewidth]{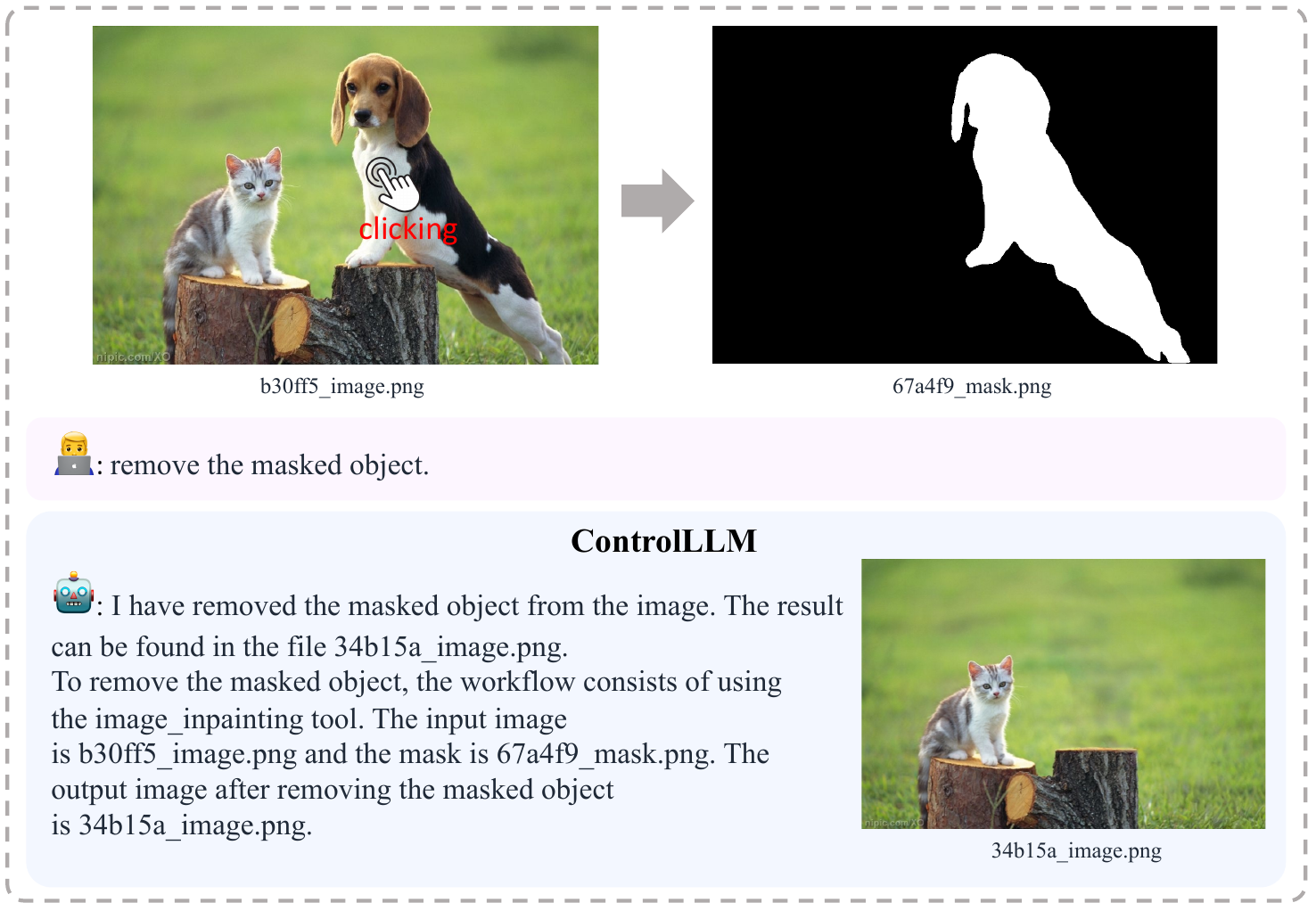}
    \caption{Interactive image editing by clicking.
    }
    \label{fig:agent_vis_2}
\end{figure*}

\begin{figure*}[h]
    \centering
    \includegraphics[width=0.9\linewidth]{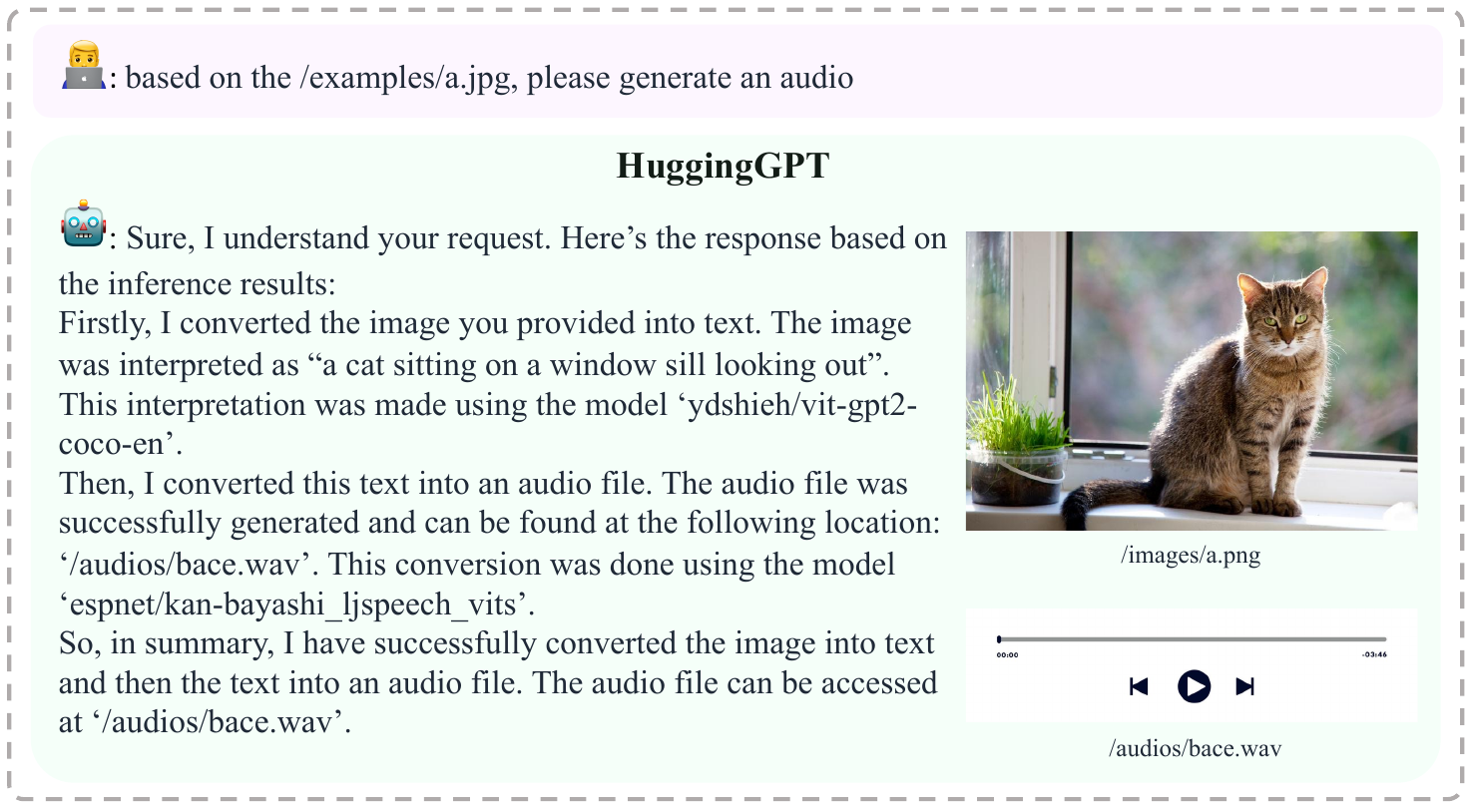}
    \caption{An example of generating audio from the given image.
    }
    \label{fig:agent_vis_3}
\end{figure*}

\begin{figure*}[h]
    \centering
    \includegraphics[width=0.9\linewidth]{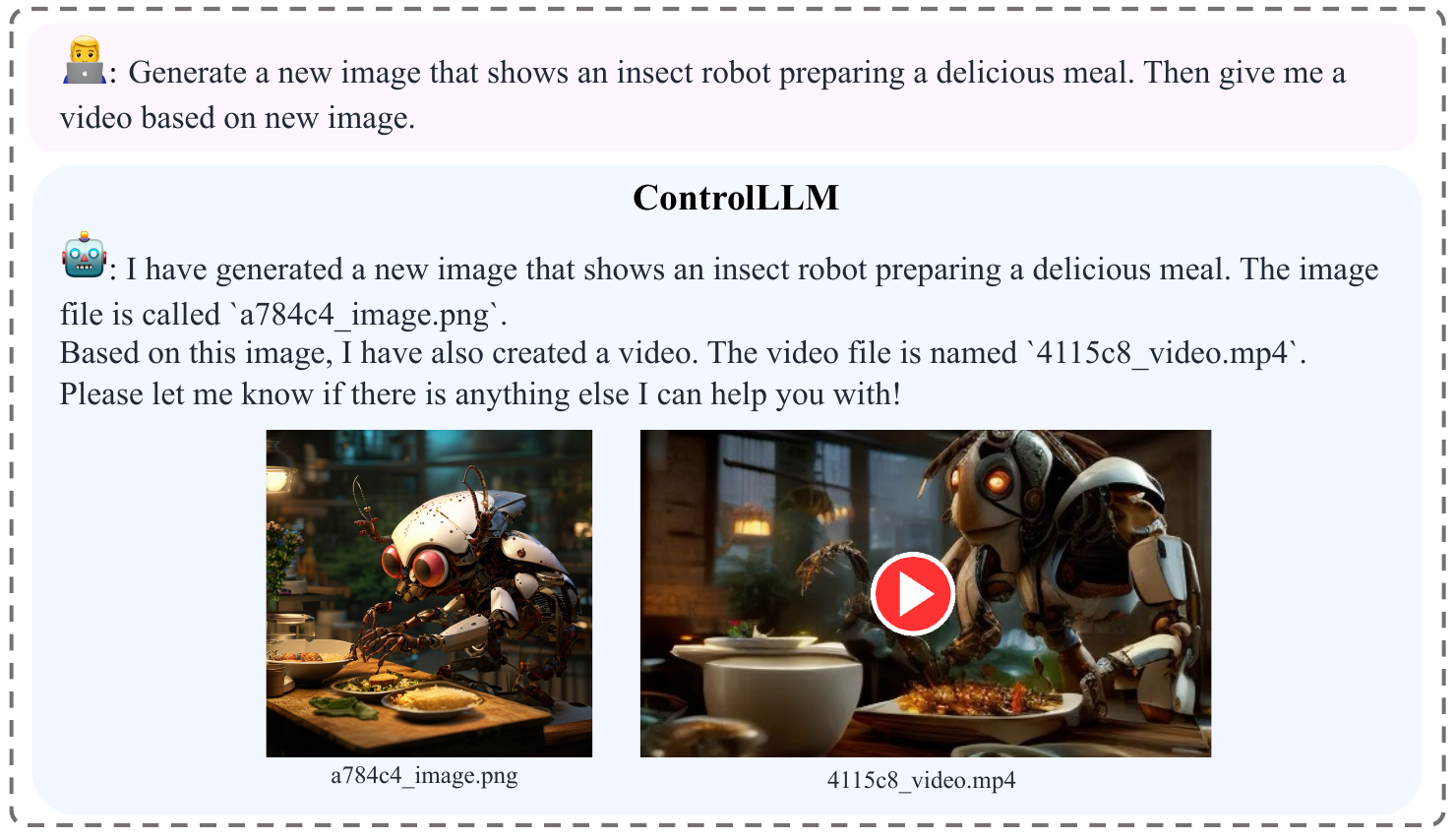}
    \caption{This example shows the multimodal agent~\cite{liu2023controlllm} generate multiple outputs from the instruction. It first generates an image conditioned on the text prompt and then creates a video from the generated image.
    }
    \label{fig:agent_vis_4}
\end{figure*}

\begin{figure*}[h]
    \centering
    \includegraphics[width=0.9\linewidth]{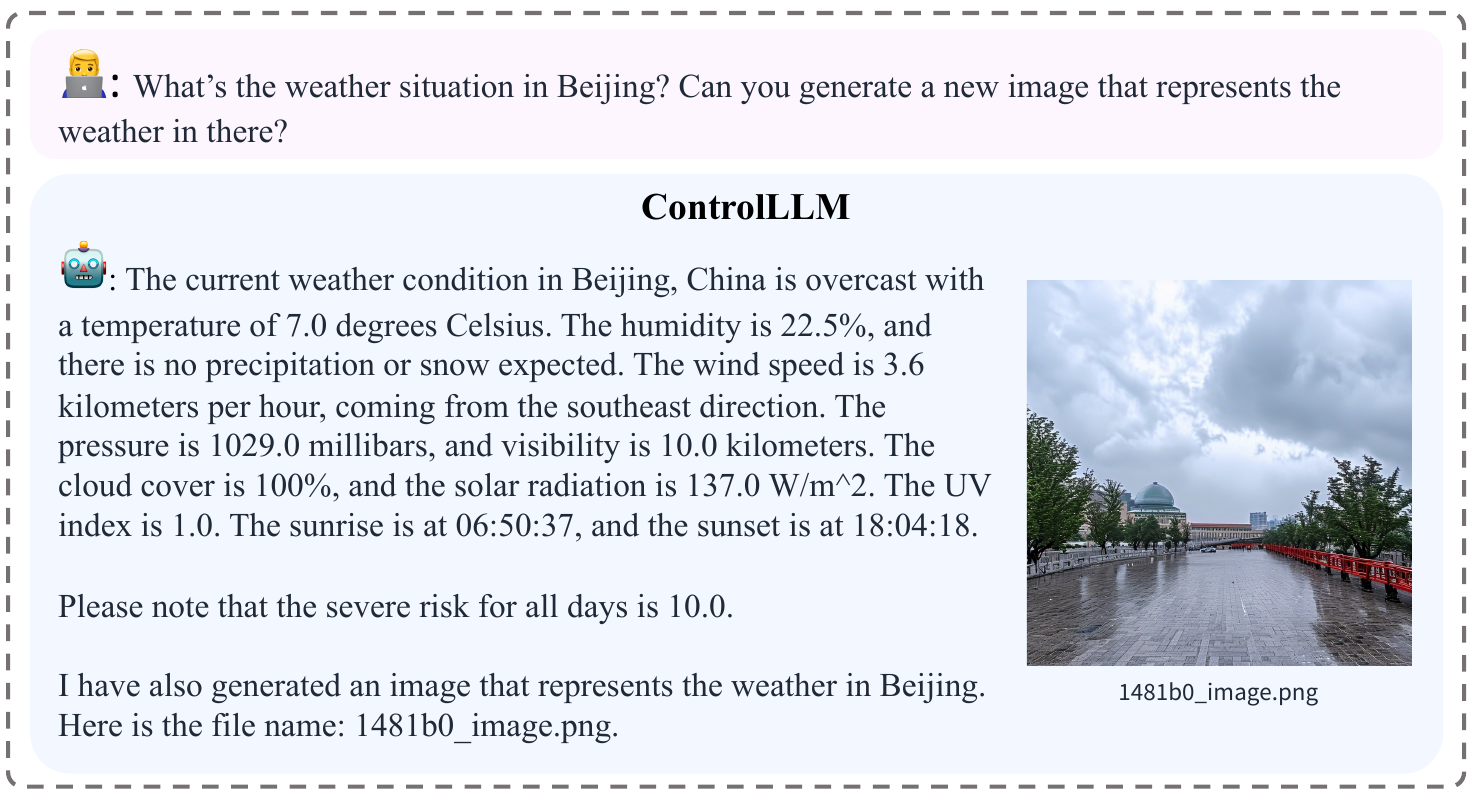}
    \caption{\centering An example of multimodal generation that visualizes the weather condition.
    }
    \label{fig:agent_vis_5}
\end{figure*}

In addition, HuggingGPT~\cite{shen2023hugginggpt} builds upon a large language model used as the core controller to manage and organize the cooperation of expert models from machine learning communities such as Hugging Face. HuggingGPT consists of four stages: task planning, model selection, task execution, and response generation. HuggingGPT separates model selection from task planning. In the task planning stage, Shen \etal~\cite{shen2023hugginggpt} use ChatGPT to parse the user request into a list of tasks, and determine the execution order and resource dependencies among them. In the model selection stage, they use ChatGPT to assign appropriate models to each task, based on the model descriptions available on Hugging Face. In the task execution stage, the system invokes and executes the selected models. Last, the response generation stage uses ChatGPT to integrate the predictions from all models and generate a response for the user. It is noticeable that HuggingGPT uses in-context learning in task planning and model selection. Therefore, it can perform well on some easy cases, but almost invariably fails to address the hard problems. Different from Fig.~\ref{fig:pipeline_of_mm_agents}, HuggingGPT actually decomposes the task planning into two steps, namely, one step to parse the task and the other step to identify tools for each task. HuggingGPT also introduces some techniques to handle resource dependencies, hybrid endpoints, and prompt design in HuggingGPT. 

\textbf{Limitations.} Training-free methods have some disadvantages. First, these methods rely on the availability and accessibility of the pre-trained LLMs, which are often proprietary and expensive to use.
Second, these methods require manual design and tuning of the prompts, which can be time-consuming and error-prone.
Last, these methods assume that the LLMs have sufficient knowledge and capability to use the tools. However, it usually fails to address complex problems. Furthermore, we found directly using off-the-shelf LLMs leads to a performance drop when extending the tool set to a large scale. This is due to the fact that LLMs normally are not trained for this purpose.

\subsubsection{Instruction-tuning Methods}
Instruction-tuning methods\cite{tang2023toolalpaca,schick2023toolformer,qin2023toolllm} involve training a language model to follow human instructions more accurately, which can vastly improve the capabilities of tool use for LLMs. As such, several multimodal agents~\cite{yang2023gpt4tools,li2023modelscope,liu2023controlllm} finetune an LLM in the first stage of task planning to use the tools across different modalities. In this type of method, the key is how to generate the instruction corpus to train the LLMs.

Taking GPT4Tools~\cite{yang2023gpt4tools} as an example, it aims to efficiently enable LLMs to use multimodal tools, such as visual models, by self-instructing from advanced LLMs, such as LLaMA~\cite{touvron2023llama} and OPT~\cite{zhang2022opt}. The first challenge is how to construct the training corpus. Yang \etal~\cite{yang2023gpt4tools} utilize a simple yet effective method to generate an instruction-following dataset by prompting an advanced LLM such as ChatGPT with various multimodal contexts and tool descriptions. Next, they filter out similar or invalid instructions from the raw data, resulting in 41K items. GPT4Tools augments the data by introducing negative samples (instructions that do not require tool usage) and context samples (instructions that involve multiple turns or actions). Finally, This work builds a dataset of 71.4K instruction-response pairs, which covers 31 tools for various visual tasks. We here show some examples of instructions in Table~\ref{tab:instructions}. After dataset construction, GPT4Tools incorporates Low-Rank Adaptation (LoRA) to finetune the open-source LLMs on the generated dataset, thus adapting them to use tools for various visual tasks, such as visual comprehension and image generation. Since the LLMs are tuned on the instruction corpus, the capabilities of tool use are dramatically improved. 

Li \etal~\cite{li2023modelscope} proposes ModelScope-Agent, utilizing the models in ModelScope to augment the open-sourced LLMs. In this work, the authors also provide a tool dataset named MSAgent-Bench. Distinct from the above methods, Li \etal ~\cite{li2023modelscope} design a module of tool retrieval with memory control to identify tools instead of directly prompting LLMs. Such a design makes the overall system more flexible and scalable.

Instead of directly training an LLM as a controller to generate the solution, Liu \etal~\cite{liu2023controlllm} train a language model to perform tool-agnostic task decomposition and propose Thoughts-on-Graph (ToG) to generate solutions for each sub-task, which enables LLMs to use various tools across different modalities, such as text, image, audio, and video, to solve complex real-world tasks. This paper argues there are three main challenges identified in tool-augmented LLMs: a) ambiguous user prompts, b) inaccurate tool selection and parameterization, and c) inefficient tool scheduling. To this end, Liu \etal~\cite{liu2023controlllm} proposes ControlLLM, a powerful framework that consists of three components:
Task decomposition, which is able to parse the user input into several subtasks with specific inputs and outputs;
Thoughts-on-Graph (ToG) paradigm, which finds the optimal solution path on a pre-built tool graph by depth-first search (DFS) algorithm;
The execution engine is equipped with a powerful toolbox, which interprets the solution path and schedules the tools efficiently on different computational devices. ControlLLM supports lots of multimodal tools and provides a user-friendly 
demo interface. 

Followed by ControLLM, Wang~\etal~\cite{wang2024mllmtool} develop a multimodal tool agent system called MLLM-Tool. In contrast to prior works~\cite{yang2023gpt4tools,li2023modelscope,liu2023controlllm}, this is the first work to train a large multimodal model for tool learning. In task planning, they utilize a multimodal encoder based on ImageBind~\cite{girdhar2023imagebind} as well as a projection layer to extract a unified embedding space for six modalities: text, image, video, audio, speech, and music. Then, the user instruction combining the multimodal embeddings is fed into a language model to predict the corresponding API names.

\textbf{Limitations.} Currently, the instruction-training methods still have some weaknesses. On the one hand, training an LLM is forbiddingly expensive for most researchers. Despite that lots of works~\cite{hu2022lora,dettmers2023qlora,peft}, such as LoRA and its variants, have been proposed to train LLMs efficiently, it still needs to make a trade-off between performance and training costs. On the other hand, it needs to generate to diverse instruction corpus for training purposes. Self-instruct~\cite{wang2022self} is an effective method that can prompt LLMs to generate more instruction automatically from seed instruction. However, it is hard to control the quality of the generated corpus, which unavoidably poses a negative impact on training LLMs. In addition, there still is an open problem to be addressed. That is, how to make an LLM that learns from a closed corpus generalize to unseen instructions. This problem is about whether the language models are able to generate novel solutions that are unseen in the training corpus, to solve more complex problems.

\subsection{Demonstrations}
Some works~\cite{wu2023visual,shen2023hugginggpt,liu2023internchat,liu2023controlllm} not only have open-sourced code, but also release online demos. As a result, in this section, we simply showcase the functionalities of some works by using their available online demos. 

Currently, several multimodal agents~\cite{wu2023visual,shen2023hugginggpt,liu2023internchat,liu2023controlllm} can interact with images, either by generating, editing, or understanding them. For example, they augment LLMs with image generation or editing models, such as Stable Diffusion~\cite{ldm}, ControlNet~\cite{zhang2023controllable} and InstructPix2Pix~\cite{brooks2022instructpix2pix}, that can create or modify images based on text prompts. As shown in Fig.~\ref{fig:agent_vis_1}, we make comparisons between Visual ChatGPT, HuggingGPT, InternGPT, and ControlLLM by an example of image generation, which is beneficial to study their capabilities directly. Interestingly, HuggingGPT decomposes the instruction into two tasks and returns two generated images. The generated image from ControlLLM is more aligned to the provided instruction. We can find a slight performance gap between different methods due to the underlying models used in their systems. In contrast to those text2image models like Stable Diffusion~\cite{ldm}, multimodal agents are able to generate vivid images in an interactive manner instead of stiffly returning images to the user. In addition, InternGPT and ControlLLM support pointing devices as input to enhance interactivity. As depicted in Fig.~\ref{fig:agent_vis_2}, taking ControlLLM as an example, the user can click the region of interest from the image, and the object lying in this region is then segmented into a mask by using SAM~\cite{kirillov2023segment}. Next, the user can send an instruction to edit the image such as removing the masked object in the image. Such a manner not only can edit the image more precisely and efficiently, but also improves the rate of success for tool use.

In addition, several works~\cite{shen2023hugginggpt,liu2023controlllm} are able to generate the video as shown in Fig.~\ref{fig:agent_vis_3} and also can dub the video with the audio track. However, there is no multimodal agent available that supports direct editing on the video frame. This is partly due to the fact that video editing is extremely challenging and still needs to be further studied. 
Some agents also support to generate audio like speech~\cite{shen2023hugginggpt,liu2023controlllm} and music~\cite{liu2023controlllm}. It is really interesting to combine the tools (\eg, image\_captioning and text\_to\_music) to generate a piece of music for an image as depicted in Fig.~\ref{fig:agent_vis_4}. Furthermore, multimodal agents are not limited to these audio-visual tasks. For example, ControlLLM~\cite{liu2023controlllm}  supports querying weather and can even visualize the weather condition by an image as demonstrated in Fig.~\ref{fig:agent_vis_5}.

\begin{table*}[!t]
    \centering
    \renewcommand{\arraystretch}{1.3}
    \caption{
    \textbf{Multimodal agents.} We only showcase the methods that build upon LMMs to solve the user's question by invoking expert models.}
    \setlength{\tabcolsep}{1.5mm}{
    \begin{tabular}{p{3.2cm}|cccccccccc}
        \Xhline{1.5pt}
        \textbf{Methods}& Venue & \makecell{Image\\Editing}& \makecell{Image\\Generation} & 
        \makecell{Video\\Editing} & \makecell{Video\\Generation} & \makecell{Audio\\Editing} & \makecell{Audio\\Generation} & 
        \makecell{3D\\Generation} & \makecell{Pointing\\Device} \\
         \Xhline{1pt}
         \textbf{Idea-2-3D}~\cite{chen2024idea} & arXiv 2024 & \xmark & \cmark & \xmark  & \xmark & \xmark & \xmark & \cmark & \xmark\\ 
         \hline
         \textbf{MLLM-Tool}~\cite{wang2024mllmtool}  & arXiv 2024 & \cmark & \cmark & \cmark  & \cmark & \xmark & \cmark & \xmark & \xmark\\ 
         \hline
          \textbf{ControlLLM}~\cite{liu2023controlllm}  & arXiv 2023 & \cmark & \cmark & \cmark  & \cmark & \xmark & \cmark & \xmark & \cmark\\ 
          \hline
          \textbf{ModelScope-Agent}~\cite{li2023modelscope}  & EMNLP 2023 & \xmark & \cmark & \xmark  & \cmark & \xmark & \cmark & \xmark & \xmark\\ 
          \hline
          \textbf{GPT4Tools}~\cite{yang2023gpt4tools} & NeurIPS 2023 & \cmark & \cmark & \xmark  & \xmark & \xmark & \xmark & \xmark & \xmark\\
          \hline
          \textbf{InternGPT}~\cite{liu2023internchat} & arXiv 2023& \cmark & \cmark & \cmark  & \cmark & \xmark & \xmark & \xmark & \cmark\\
          \hline
          \textbf{HuggingGPT}~\cite{shen2023hugginggpt} & NeurIPS 2023& \cmark & \cmark & \xmark  & \cmark & \xmark & \xmark & \xmark & \xmark \\
          \hline
          \textbf{Visual ChatGPT}~\cite{wu2023visual} & arXiv 2023& \cmark & \cmark & \xmark  & \xmark & \xmark & \xmark & \xmark & \xmark\\
          \hline
          \textbf{VISPROG}~\cite{gupta2023visual} & CVPR 2022& \cmark & \xmark & \xmark  & \xmark & \xmark & \xmark & \xmark & \xmark\\
         \Xhline{1.5pt}
    \end{tabular} }
    \label{tab:feature_comparison}
\end{table*}

\subsection{Summary}
This section investigates LLMs with external tools to enhance their capabilities, particularly in multimodal interactions. The motivation behind this integration is to address LLMs' limitations in processing information not present in their training data, such as real-time or private data. This is achieved by augmenting LLMs with tools that can provide additional information or functionalities, which are invoked through natural language instructions.

The methods fall into two categories: training-free methods, which rely on prompt engineering and in-context learning, and instruction-tuning methods, which involve training LLMs to follow instructions more accurately. These methods generally involve a three-stage framework: task planning, task execution, and response generation. Task planning interprets instructions into tool invocation schemes, Task execution involves the use of multimodal tools for tasks like image generation or audio synthesis, and Response Generation creates user-friendly responses from the execution outputs.

In summary, tool-augmented multimodal agents represent a significant advancement in human-computer interaction, enabling more natural and versatile interactions and fostering creative applications across various modalities. However, they also present challenges that need to be addressed to realize their full potential.

\section{Generative AI Safety}
\label{sec: safety}
The security concerns of multimodal-generated content are drawing increasing attention. 
The research mainly focuses on mitigating biased and toxic content generation, safeguarding copyright, and alleviating the impact caused by fabricated content originating from generative models. 

The vulnerability of generative models to attacks or malicious usage presents unique challenges and has attracted significant research attention. 
Recent research includes optimization-based attacks, prompt-level manipulations, and data poisoning methods: 
(i) Optimization-based attacks demonstrate the effectiveness of adversarial techniques to degrade model performance~\cite{wallace2019universal} or induce biases and harmful outputs~\cite{fu2023misusing,bailey2023image,zou2023universal,jones2023automatically}. 
Adversarial attack and detection-based defense research are also conducted in the field of audio and video ~\cite{zelasko2021adversarial,zelasko2021adversarial,chen2021appending,liu2023coherent,lo2021defending,lee2023defending}. 
(ii) Prompt-level attacks~\cite{wu2023jailbreaking,xie2023defending,liu2023prompt,perez2022ignore} reveal the risks at the inference level, where human-crafted inputs can bypass safeguards and elicit unsafe outputs, bringing the security challenges. 
(iii) Data poisoning methods~\cite{carlini2023poisoning,jia2017adversarial} expose that models can be manipulated by injecting malicious data inputs when the integrity of training data is broken. 
These research works underscore the need for comprehensive approaches to enhance model robustness, secure data integrity, and recognize user's unsafe interactions, addressing the generative AI vulnerabilities. 

Following the discussion on techniques to attack large generative models, there are generally two main approaches to defending against undesirable generation content. 
The first approach involves not modifying the existing parameters of the model but employing detection mechanisms or manipulating the input prompt context. 
~\cite{van2023detecting} utilizes VLMs to detect and correct hate speech in multimodal memes. ~\cite{wei2023jailbreak,robey2023smoothllm} manipulate LLMs through in-context learning and defending against jailbreaking attacks. For text-to-image generation, an effective and efficient framework named Latent Guard~\cite{liu2024latent} is proposed to detect the unsafe input prompt. Compared to traditional blacklist-based approaches, it is much more robust because the input prompt is checked in a latent space. And the speed is much faster than the LLM-based unsafe detection method. 
The second strategy enhances safety by aligning the model with human preferences or values using alignment algorithms. 
The Proximal Policy Optimization (PPO) algorithm, introduced in ~\cite{schulman2017proximal}, has been widely used for aligning LLMs. Recently, Direct Preference Optimization(DPO) and related methods ~\cite{rafailov2024direct,pi2024strengthening} introduce an improvement over ~\cite{schulman2017proximal} by presenting a more efficient alignment algorithm capable of learning alignment directly from preference data. For image generative models some preference datasets ~\cite{wu2023better} and alignment methods ~\cite{dong2023raft} have also been proposed. 

Some studies focus on preventing multi-modal generative models from fabricating facts. 
Powerful generative models that are capable of producing highly realistic videos have attracted significant attention for their potential misuse like Deepfakes ~\cite{korshunov1812deepfakes,mirsky2021creation,masood2023deepfakes}. 
Deepfake is a technique that can generate realistic content of a certain identity in the form of an image or video. 
Methods ~\cite{verdoliva2020media,wodajo2023deepfake,wodajo2021deepfake} focus on distinguishing deepfake videos by detecting visual artifacts. 
While these detection methods are still limited~\cite{hussain2021adversarial} that they do not insist on adversarial attacks so they are hard to work well on new AIGC models that rely on the pattern seen in training. 

Generated content can also bring up copyright issues. 
To solve these problems, we may need to use data attribution and embed watermarks in the generated outputs. 
Data origin attribution, which involves tracing the model prediction to its original training data, can help find the data causing copyright issues~\cite{shi2023detecting,park2023trak,wang2024did}. 
Watermark technique ~\cite{kirchenbauer2023watermark,cui2023diffusionshield,fernandez2023stable} can make generated content distinguishable from real content by integrating the ownership information in the generated content. 

In addition, a variety of datasets also have been proposed to evaluate different aspects of generative AI safety. 
The SafetyBench~\cite{zhang2023safetybench} dataset is a multiple-choice question dataset for assessing unsafe contents which contains 11,435 items spanning 7 safety categories. 
This effort is complemented by the GOAT-Bench dataset~\cite{lin2024goat} that evaluates unsafe memes with over 6K diverse topics, including implicit hate speech, gender discrimination, and cyberbullying.
Furthermore, datasets like ToViLaG and others ~\cite{wang2023tovilag,gong2023figstep,liu2023query} have been developed specifically for visual LLMs, shedding light on the challenges of addressing the generation of toxic content, such as offensive texts and inappropriate images. 
These datasets offer a comprehensive evaluation for further enhancing the safety of generative models, spanning across text and image contents. 

In summary, safety techniques for generative AI models can mitigate ethical risks and protect copyrights. Detection and data algorithm technologies are being utilized by commercial models. And some open-source projects also offer safety checks for users by default. Watermarking and data tracing technologies have made significant progress in alleviating copyright protection concerns. Adopting safety technologies for impactful public projects can enhance the security and trustworthiness of multimodal generation applications. Table~\ref{tab:safety} summarizes selected research related to generative AI safety issues.

\begin{table}[htbp]
\centering
\caption{Overview of generative AI safety across various modalities and methods. The term "Adv." denotes "adversarial attack". }
\label{tab:safety}
\begin{tabular}{@{}lllll@{}}
\toprule
\textbf{Name}            & \textbf{Media}       & \textbf{Type}    & \textbf{Method}    & \textbf{Venue}                 \\ \midrule
Wallace et al.\cite{wallace2019universal}  & T           & Attack  & Adv.      & EMNLP 2019            \\
Fu et al.\cite{fu2023misusing}       & T           & Attack  & Adv.      & arXiv 2023            \\
Image hijacks\cite{bailey2023image}   & I           & Attack  & Adv.      & arXiv 2023            \\
Jones et al.\cite{jones2023automatically}    & T           & Attack  & Adv.      & ICML 2023             \\
Wu et al.\cite{wu2023jailbreaking}       & T + I       & Attack  & Prompt    & arXiv 2024            \\
Xie et al.\cite{xie2023defending}      & T           & Attack  & Prompt    & NMI 2023 \\
Liu et al.\cite{liu2023prompt}      & T           & Attack  & Prompt    & arXiv 2023            \\
Carlini et al.\cite{carlini2023poisoning}  & T           & Attack  & Data      & arXiv 2023            \\
Jia et al.\cite{jia2017adversarial}      & T           & Attack  & Data      & EMNLP 2017            \\
Latent Guard\cite{liu2024latent} & T+I       & Defense & Detection    & arXiv 2023            \\
Van et al.\cite{van2023detecting}      & T+I       & Defense & Detection    & arXiv 2023            \\
Wei et al.\cite{wei2023jailbreak}      & T           & Defense & Prompt    & arXiv 2023            \\
Smoothllm\cite{robey2023smoothllm}       & T           & Defense & Prompt    & arXiv 2023            \\
Rafailov et al.\cite{rafailov2024direct} & T           & Defense & Alignment & arXiv 2023            \\
Raft\cite{dong2023raft}            & T+I         & Defense & Alignment & TMLR 2023             \\
Wodajo et al.\cite{wodajo2023deepfake}   & V       & Defense & Detection    & arXiv 2023            \\
Safetybench\cite{zhang2023safetybench}     & T           & Dataset & -         & arXiv 2023            \\
GOAT-Bench\cite{lin2024goat}      & T           & Dataset & -         & arXiv 2024            \\
ToViLaG\cite{wang2023tovilag}         & T+I         & Dataset & -         & EMNLP 2023            \\
Figstep\cite{gong2023figstep}         & T+I         & Dataset & -         & arXiv 2023            \\
Liu et al.\cite{liu2023query}      & T+I         & Dataset & -         & arXiv 2023            \\ \bottomrule
\end{tabular}
\end{table}
\section{Applications}
\label{sec: applications}
The rapid advancements in LLMs from companies like OpenAI, Google, Meta, Baidu, and Microsoft have led to the development of a wide range of impressive AI-powered applications. These models, such as GPT-4, Gemini, and Claude, have demonstrated remarkable capabilities in multimodal tasks, particularly in multimodal understanding.

The ability of these models to comprehend, interpret, and generate multimodal content is a significant milestone in artificial intelligence. This multimodal capability holds immense potential for various industries and showcases the effectiveness of LLMs in multimodal generation. 

In this section, we will review some remarkable applications that have already been published. Beginning with image generation and progressing to video, audio, and 3D generation, these showcases demonstrate the remarkable impact of LLMs in generating content across multiple modalities.

\subsection{Image}
The rapid advancements in diffusion models have witnessed a remarkable increase in the quality and realism of synthesized images. This has sparked the emergence of numerous companies developing high-quality text-to-image generation tools and multimodal conditional image editing or generation solutions.

Midjourney\cite{midjourney} is making significant strides in the industry. It enables content creation and design by offering users the ability to generate high-quality, realistic images from text prompts. Its user-friendly interfaces and robust performance make it a top choice for professionals and enthusiasts in image generation.

Moreover, Stability AI\cite{StabilityAI} has provided a powerful open-source generation model. The user community has provided various usage methods, indeed handing over creativity and tools to the users. Opening up fine-tuning has created a sizeable open-source image usage community. Even artists who are not computer scientists can easily make their small models based on their basic models. Users integrate various modal tools for deployment, enabling their image generation model to play a better role.

DALLE3\cite{dalle3} stands out as a remarkable example of seamlessly integrating image generation capabilities into the powerful ChatGPT4 chatbot\cite{gpt4page}. With DALLE3, users can generate and modify images through text-based prompts. The success of DALL-E\cite{dalle} and DALL-E 2\cite{dalle2} from OpenAI\cite{openai} has paved the way for highly sophisticated image generation capabilities within LLMs. These models can create detailed, photorealistic images from textual descriptions, allowing for rapid prototyping and content creation across numerous domains.

In addition to the industry-leading solutions mentioned above, lots of text-to-image generation tools have emerged that leverage LLMs to enhance the robustness and overall quality of the user experience. By leveraging LLMs to expand and refine captions, these tools can improve the quality of the generated images and the platforms' overall reliability and user-friendliness.

\subsection{Video}
With the advent of large-scale video generation models, individuals can now obtain a high-quality video clip by simply inputting a textual description. 
Users do not need specialized skills in traditional video production, such as CG modeling, 3D modeling, or other professional knowledge. 
Users can generate desired video clips prompted by a textual description and then assemble them to create a captivating short film or animated video.
Existing prominent tools in this domain include commercial tools like Pika~\cite{pikalabs} and Runway's Gen2~\cite{gen2}, as well as open-source video generation models such as AnimateDiff~\cite{guo2023animatediff}, VideoCrafter~\cite{chen2023videocrafter1} and SVD~\cite{chai2023stablevideo}. 
Regarding human video generation, Heygen~\cite{heygen} is a popular tool widely applied in various domains, including e-commerce, social media, and advertising videos.

After some demo videos generated by Sora\cite{sora} were released, significant advancements have been made in terms of realism and prompt-following capabilities, instilling greater confidence in the application of large-scale text-to-video models. Many efforts have been made to reduce the video production cost of the film and television industry.

\subsection{Audio}
The application of multimodal AI in audio has been explored for a long time. The use cases are more well-defined, and the demand for customized and diversified sounds is more established. Technologies such as text-to-speech generation, sound transfer, music generation, and other audio generation have demonstrated promising prospects in education, video dubbing, intelligent terminals, voice assistance, and the medical field.

Microsoft's Azure platform\cite{azure} is taking a leading position in speech generation and is driving the integration of AI-generated sounds across short-video platforms. Descript\cite{descript}, an AI-based audio and video editor, can transcribe speech in audio and video into text, enabling users to modify the audio and video akin to editing a Word document. Moreover, numerous video platforms, video editing software, and audio platforms have devoted significant attention to applying multimodal models and audio generation.

Besides speech and audio generation, music generation is also a hot spot in industries. The passion for music has driven countless AI researchers and scientists to dedicate immense effort to advancing this field. Suno AI\cite{Suno} has ushered in the "Sora era" of music generation, where users can now create vivid, high-quality songs simply by providing a text prompt describing the desired lyrical style. Additionally, companies like Stability Audio\cite{StabilityAudio}, Google's MusicFX\cite{musicfx}, Tuneflow\cite{tuneflow}, and Deepmusic\cite{deepmusic} have also provided their music generation products, further expanding the capabilities in this domain.

\subsection{3D}
The generation of 3D models is crucial across diverse domains, including film, gaming, industrial design, architecture, interior design, product design, and virtual reality. It provides realistic visual experiences and immersive interactions, facilitating the creation of characters, scenes, products, and virtual environments to enhance creativity and engagement.
Meta\cite{meta} has been heavily investing in 3D modeling and virtual reality technologies. Epic Games' MetaHuman Creator\cite{MetaHuman}, a cloud-streamed app designed to elevate real-time digital human creation, is another noteworthy development that can be used in conjunction with Unreal Engine, a state-of-the-art real-time engine and editor.

As for 3D reconstruction and generation, Luma AI\cite{lumalabs} is making significant advancements, with their technology capable of generating 3D models from 2D images, simplifying the process of creating 3D content. Other industry players, such as Adobe\cite{adobe} and Kaedim3D\cite{kaedim3d}, are also making substantial strides in this field. Adobe's\cite {adobe} 3D and AR tools enable the creation of immersive content, while Kaedim3D's\cite {kaedim3d} AI technology can convert 2D images into 3D models.

Wonder Studio\cite{wonderdynamics} is a powerful AI tool for character replacement in videos, where it can replace the original characters in a video with user-created 3D models, unlocking exciting possibilities for personalized content creation.

Recent advancements in Language-to-Language Models (LLMs) have revealed the significant potential of text interaction and generation, opening new possibilities for creating and manipulating 3D models using natural language commands, making the process more intuitive and accessible. For instance, SceneScript~\cite{avetisyan2024scenescript} from Meta is able to reconstruct environments and represent the layout of physical spaces based on their powerful language-based model, Llama\cite{touvron2023llama}. However, compared to the image-to-3D, the text-to-3D is still the research topic of companies like Meta~\cite{meta}, Google~\cite{google}, Tencent~\cite{tencent}, \etc.

Integrating LLMs in the 3D world is transforming how we create and interact with digital content. As these technologies continue to evolve, we anticipate even more interesting and practical applications.

\subsubsection{Others}
An AI-driven software normally needs to handle various modalities of input data. This growing demand for multimodal solutions highlights the importance of advanced AI models that can seamlessly integrate and process various data types.
For instance, AI-generated movies incorporate 3D technology for video, music, and speech generation, collaborating with human artists to produce high-quality cinematic experiences. Digital humans have also emerged as prominent figures across various industries, from live streaming and gaming to memorial services and large-scale interactive displays. Furthermore, LLM $+$ multimodal generative tools have found diverse applications in mathematics, law, education, and robotics fields. In summary, we are currently witnessing the dawn of multimodal generative models with LLMs, which will undoubtedly change our lives.

\section{Future prospects}
\label{sec: future_work}
LLMs augmented multimodal generation stands out as a promising research topic, which harnesses the linguistic knowledge of LLMs to enhance the generation across various modalities, such as image, video, 3D, and audio. 
This series of approaches not only can improve the quality, diversity, and controllability of the generated content but also can facilitate interactivity during multimodal generation. In line with this direction, we intend to present prospects for future works.

\subsection{Technical Prospects}
In this section, we focus on the technical prospects for multimodal generation, which are expected to provide more insights and facilitate future work.

\subsubsection{High-resolution Generation}
High-resolution multimodal generation is pivotal as it directly impacts the quality and usability of generated content across various domains such as image~\cite{scalecrafter, make-a-cheap-scaling}, video~\cite{videogigagan, upscale-a-video}, audio and 3D generation~\cite{lin2023magic3d}. Accordingly, high fidelity also needs to be taken into account in audio generation\cite{kumar2024high,agostinelli2023musiclm,roman2023discrete,roman2023discrete,yao2023jen}. 
The ability to produce High-resolution multimodal generation is crucial for applications requiring detailed and realistic representations, ranging from virtual reality to film production. Because it enhances the perceptual experience, provides more information for analysis, and improves the performance of subsequent tasks like object recognition and scene understanding.

LLMs hold the potential to address the challenges in a high-resolution multimodal generation. They can provide a more seamless integration of visual and textual modalities, offering a dialogue-based interface and instruction-following capabilities~\cite{betker2023improving}. It could enhance the generation process by improving the understanding of complex instructions and generating more accurate and diverse outputs. Recent advancements across different modalities, including image~\cite{ldm,ding2022cogview2,betker2023improving,chen2023pixart}, 
video~\cite{chen2023videocrafter1,blattmann2023align,zhang2024videoelevator,henschel2024streamingt2v},
3D~\cite{pan2024enhancing,or2022stylesdf,huang2023refsr} and audio~\cite{liu2023audioldm, kreuk2022audiogen, huang2023make2}, have led to significant improvements in the quality of generated content. We are extremely eager to witness the increase in future works that integrate LLMs, thereby offering enhanced support for high-resolution generation. Moreover, high-resolution content generation typically entails substantial hardware expenses and time costs. Consequently, the efficient generation of high-resolution content is also a topic worthy of research.

\subsubsection{Long-term Sequence Generation}
Long-term sequence generation is crucial for creating immersive experiences in the video~\cite{henschel2024streamingt2v,wang2023gen} and the audio~\cite{yoo2023towards,borsos2023audiolm,liu2023audioldm,lin2024arrange}. In video, it allows for the portrayal of evolving scenes and narratives, while in audio, it supports the development of music and dialogue that can adapt and flow over time.
The ability to generate long sequences over time is not just a technical challenge but also a creative one, where the model must understand and predict complex patterns and progressions.
It should maintain continuity, prevent repetition, and introduce novel elements that align with the overarching theme and input conditions. Only when we are able to generate long sequences for video and audio, it can potentially lead to practical significance.

Recent advancements in LLMs, such as OpenAI's GPT series and Meta's LLaMA~\cite{touvron2023llama}, addressing the challenges of long-term sequence generation. LLMs build upon pre-trained language representations and fine-tuning techniques to capture intricate patterns and dependencies in text data, enabling them to generate coherent and contextually relevant sequences over extended lengths. 
By harnessing the contextual understanding and generative capabilities of LLMs, researchers can explore long-term sequence generation. For example, fine-tuning pre-trained LLMs on multimodal datasets could enable them to generate coherent and diverse sequences across different modalities, including video and audio. Additionally, techniques such as prompt engineering and conditioning can guide the generation process toward desired outcomes, allowing for the creation of long sequences with specific themes or narratives. We argue that LLMs can enhance coherence and consistency for generated long sequence generation.

Generally, long-term sequence generation represents a complex yet compelling area of research for various domains. By leveraging the capabilities of LLMs and addressing the associated challenges, researchers can unlock new opportunities for creating immersive and engaging sequences that captivate audiences and push the boundaries of content creation and storytelling. 

\subsubsection{More Accurate and Fine-grained Generation Control}
Accurate and fine-grained generation control is a significant topic in AIGC for several reasons. First, it allows for the creation of more realistic and high-quality multimodal content. This is particularly important in fields such as entertainment, advertising, and education, where high-quality content can significantly enhance user experience.
Second, fine-grained control can facilitate more effective communication between humans and AI. For instance, an AI model with fine-grained control can generate a specific image or sound based on a user's detailed description, thereby improving the interaction between the user and the AI.
Last, fine-grained control can also contribute to the advancement of other AI fields. For example, in reinforcement learning, an AI agent can learn more effectively if it can generate detailed and accurate simulations of its environment.

Lots of methods~\cite{betker2023improving,zhang2023adding,zhao2023controlvideo,chen2023videocrafter1} have been proposed to address accurate and fine-grained generation control. 
However, these methods still have some limitations. For instance, they still struggle with generating fine details, such as fingers or body parts, which can lead to unrealistic outputs. Moreover, they may also fail to accurately capture the nuances in the control signals, resulting in a mismatch between the generated content and the control signal.

Large language models have shown remarkable capabilities in understanding and generating text. By leveraging these capabilities, we can potentially improve the accuracy and granularity of generation control.
One of prominent examples is text rendering~\cite{liu2022character,chen2024textdiffuser,chen2023textdiffuser2,ma2023glyphdraw,yang2024glyphcontrol} on images or videos. It has been observed that by using powerful language models, such as T5-XXL, as the encoder, the image generation models will exhibit better spelling ability. In this context,  the integration of more potent LLMs into generative models is worthy of further exploration. Generally, a large language model can be trained to better understand the nuances in the control signals, thereby improving the alignment between the control signal and the generated content. 

\subsubsection{Multi-view Consistency}
Multi-view consistency (MVC) is a fundamental aspect of visual generation, particularly in 3D generation, ensuring the coherence and continuity of an object's appearance from different viewpoints. This consistency is crucial for applications in augmented reality (AR), virtual reality (VR), and computer graphics, where users interact with 3D objects in a seemingly real-world context. Inconsistent appearances can break immersion and lead to a less realistic experience. 
The significance of multi-view consistency lies in its ability to provide a seamless and integrated perception of 3D objects, enhancing the user's experience and interaction with digital content. 

MVC is particularly challenging due to the complex nature of translating 2D images into consistent 3D models, where issues such as occlusions, lighting variations, and geometric distortions can arise.
Recent advancements pay lots of attention to multi-view consistency. In 3D generation, Sculpt3D~\cite{chen2024sculpt3d} introduces a sparse 3D prior to improve consistency without retraining the 2D diffusion model. HarmonyView~\cite{woo2023harmonyview} addresses the balance between consistency and diversity by employing a diffusion sampling technique. Additionally, MVDream~\cite{shi2023MVDream} lacks comprehensive multi-view knowledge or 3D-awareness during score distillation, leading to unstable generation and artifacts. In image and video generation, the works~\cite{ye2023consistent,zuo2024videomv} have contributed to the field by focusing on novel view synthesis and multi-view image generation based on large video datasets, respectively. 

Despite these advancements, there are still several challenges to be further studied: 
1) Limited Generalization: Many methods struggle to generalize well across diverse datasets and object categories.
2) Struggling in Complex Geometries: Accurately rendering objects with complex geometries or textureless surfaces.
Since linguistic prompts can provide more prior knowledge for generation, we believe it can enhance the multi-view consistency as well as generation quality by incorporating LLMs in the pipeline. 

\subsubsection{Unified Training for Multimodal Generation} 
Multimodal generation is defined as the ability to simultaneously create content across different modalities, including images, videos, 3D objects, and audio. 
Currently, most methods~\cite{betker2023improving,ldm,chai2023stablevideo,kondratyuk2023videopoet,hu2024scenecraft,feng2023posegpt,liu2023audioldm,borsos2023audiolm} only focus on one aspect such as text-to-image or text-to-video synthesis. This inevitably prompts consideration: Can a single model possess the capability to generate multiple modalities?

Several recent works~\cite{girdhar2023imagebind,wang2023one,zhang2023video,boletsis2023invizar,lu2023unified,chen2024vast} have made notable strides in feature alignment for text, image, video, audio and other modalities.
Some multimodal agents~\cite{shen2023hugginggpt,wu2023visual,yang2023gpt4tools,liu2023internchat,liu2023controlllm} offer awesome generation capabilities for various modalities, but atom tools they used are not jointly trained. Furthermore, the pioneer works~\cite{girdhar2023imagebind,hussain2023m,wu2023next} have made their preliminary efforts to explore how to generate multimodal content in one model.

However, despite these advancements, challenges persist in achieving effective unified training for multimodal generation. One prominent obstacle lies in feature alignment across different modalities, as each modality possesses distinct statistical properties and underlying structures, necessitating robust alignment mechanisms to ensure consistency and coherence in generated outputs. Moreover, the mutual interference during training poses a significant hurdle, as optimizing for multiple modalities concurrently may lead to conflicts or competition among modal-specific objectives, hindering the overall training stability and convergence. In addition, the inherent complexity of multimodal data imposes computational overhead, necessitating efficient algorithms and scalable architectures to handle the diverse modalities efficiently.

The pursuit of unified training for multimodal generation represents a crucial advancement in AI research, offering immense potential for advancing the capabilities of generative models across diverse domains. In the future, we even look forward to models capable of generating different modalities in an interleaved manner.

\subsubsection{Efficient Training and Deployment Strategies}
Efficient training and deployment strategies also still remain to be studied in multimodal generation. As datasets and models continue to scale exponentially, the challenge of achieving efficient training and deployment becomes increasingly significant, in line with the scaling law, which posits that the computational resources required for training and deploying models grow rapidly with model size and dataset size~\cite{kaplan2020scaling}. Efficient strategies are essential not only for reducing computational costs but also for enabling real-time or resource-constrained applications of multimodal generation technologies. By minimizing computational overhead and resource utilization, efficient training and deployment strategies not only reduce time and energy costs but also enhance scalability and accessibility, democratizing access to advanced generative capabilities across diverse domains. 

Several approaches have been proposed to address the challenge of efficient training in multimodal generation. Several works study low-rank approximation techniques, such as LoRA~\cite{hu2022lora} and Q-LoRA~\cite{dettmers2023qlora}, which aim to reduce the computational complexity of model training by approximating weight matrices with low-rank structures. Additionally, mixed precision training~\cite{micikevicius2017mixed}, which involves using reduced precision (e.g., 16-bit floating-point) arithmetic for certain computations, has emerged as a powerful tool for accelerating training without sacrificing model accuracy. 
Despite their effectiveness, these efficient training techniques still have limitations. Low-rank approximation methods may introduce approximation errors that degrade the quality of generated outputs, particularly in scenarios where high-fidelity synthesis is crucial. Similarly, mixed precision training may encounter numerical instability issues, especially when dealing with extremely large models or datasets, leading to suboptimal convergence or even training failures.

Efficient deployment strategies, such as quantization~\cite{jacob2018quantization,dettmers2022gpt3,choukroun2019low,wu2023understanding} to int8 or even int4 precision, offer another avenue for reducing the computational and memory requirements of multimodal generation models during inference. By quantizing model weights and activations to lower precision formats, significant savings in memory bandwidth and computational resources can be achieved, enabling faster inference and deployment on resource-constrained devices.
However, quantization also presents issues, particularly in preserving model accuracy and generative quality. Lowering the precision of model parameters and activations can lead to information loss and degradation in output fidelity, especially in complex multimodal synthesis tasks where fine-grained details are crucial. 

In conclusion, efficient training and deployment strategies are indispensable for realizing the full potential of multimodal generation technologies across diverse applications. By overcoming the challenges associated with scalability and resource constraints, researchers can accelerate the adoption of multimodal generation systems in real-world scenarios, unlocking new possibilities for content creation, human-computer interaction, and beyond.

\subsubsection{Ethically Safe Content Generation}
While there have been many works exploring how to strengthen the security of generative models of text and images\cite{van2023detecting}\cite{wei2023jailbreak}\cite{robey2023smoothllm}, the increasing capabilities of video generation models should raise security concerns. Due to the emergence of security issues like Deepfakes\cite{masood2023deepfakes} even with the use of previously less powerful video models, the growing strength of video models magnifies the societal impact of potential risks. 

Adversarial attacks have demonstrated effective transferability from open-source models to commercial closed-source models\cite{zou2023universal}. Future commercial closed-source models should consider guarding against attacks from open-source models, such as by implementing corresponding adversarial token detection mechanisms. Simultaneously, efforts can also be considered to mitigate the impact of transferable attacks like minimizing the commercial models similarities with open-source models, such as in network architecture, data usage, and weights. 

Currently, most research articles focus on ensuring security from individual perspectives such as detection\cite{van2023detecting}, alignment\cite{schulman2017proximal}, post-hoc checking\cite{qu2023unsafe}, etc. Each of these methods generally has their own advantages and disadvantages. For instance, detection techniques offer fast checking but may overlook certain vulnerabilities. Alignment methods also cannot guarantee that the data used for training alignment covers all security cases. Additionally, post-hoc checking can be computationally expensive, especially for the generation of images and videos. There has not been much work integrating these techniques into a holistic system to ensure the security of large generative models. For example, the system can first detect user inputs, then simultaneously apply securely aligned models, and finally conduct security checks on the output to determine whether to proceed. Integrating these techniques can lead to higher efficiency and security.

\subsection{Application Prospects}
In this section, we make an effort to build blueprints for the application of multimodal generative models.

\subsubsection{Semantic Audio Synthesis}
Semantic audio synthesis involves generating of audio signals based on semantic descriptions or contextual cues, enabling the creation of immersive auditory experiences with specific characteristics or attributes. Multimodal generative models offer a promising approach to semantic audio synthesis by leveraging contextual information from other modalities, such as text or images. For instance, text-based descriptions of soundscapes or music compositions can be translated into audio waveforms using generative models trained on multimodal data. Similarly, images or videos depicting scenes or environments can inform the generation of corresponding audio accompaniments, enhancing the realism and richness of multimedia content. By integrating semantic information across modalities, multimodal generative models enable the creation of highly personalized and contextually relevant audio experiences, spanning applications in entertainment, virtual reality, and assistive technologies.

\subsubsection{Multi-modal Storytelling.}
Multimodal storytelling involves fusing different modalities to craft compelling narratives that engage multiple senses simultaneously. This approach not only enriches storytelling experiences but also opens up new avenues for creative expression and audience engagement. In multimodal storytelling, the synthesis of content can occur in several directions.

From text prompts to image sequences, multimodal storytelling can begin with a topic, a script, or even a story outline, which serves as the basis for generating complementary modalities such as text and image sequences. For instance, given a prompt about a fantastical adventure, a multimodal generative model could generate vivid imagery depicting characters and scenes, produce an animated video sequence illustrating key events, or compose a thematic musical score to accompany the narrative.

From text prompts or images to videos and audio, in this scenario, an image serves as the starting point for generating accompanying textual descriptions, video sequences, or audio narratives. For example, given an image depicting a scenic landscape, a multimodal generative model can generate descriptive text detailing the setting, produce a video animation depicting the scene in motion, or create an immersive audio experience capturing ambient sounds and atmosphere.

Multimodal storytelling holds immense potential for enhancing traditional narrative formats and creating immersive, multi-sensory experiences that resonate with audiences across various mediums and platforms. By harnessing the capabilities of multimodal generative models, storytellers, content creators, and media producers can unlock new dimensions of creativity and engagement in the digital age.

\subsubsection{Interactive Content Design}
Interactive content design aims to create and manipulate media elements in real time, empowering users to actively participate in the creative process. Traditionally, content creation processes involve iterative steps of ideation, design, and refinement, often requiring extensive time and resources. However, with the interactive capabilities afforded by foundation generative models, creators can swiftly explore a multitude of design possibilities, rapidly iterate on concepts, and refine compositions in real time, thereby streamlining the overall creative workflow.

By enabling real-time interaction and manipulation of media elements, it can improve the efficiency of multimodal generative models. Creators can efficiently experiment with different visual and auditory elements, explore diverse artistic styles, and generate high-quality content without the need for extensive manual labor or specialized expertise. Consequently, this not only accelerates the production process but also minimizes the expenses incurred in hiring additional resources or outsourcing tasks. 
Moreover, the integration of multimodal generative models in Interactive Content Design contributes to the democratization of creativity by lowering the barriers to entry for aspiring artists and designers. Unlike traditional design tools that often require proficiency in complex software interfaces or artistic skills, these models offer intuitive and accessible interfaces that empower individuals from diverse backgrounds to engage in creative expression. By democratizing access to advanced content creation capabilities, these tools foster inclusivity and diversity within the creative community, enabling a broader range of voices to be heard.

Looking ahead, the evolution of multimodal generative models holds exciting prospects for the future of Interactive Content Design. As advancements continue to expand the scope and fidelity of generated content across different modalities, we can anticipate even greater opportunities for innovation in areas such as virtual reality, augmented reality, and immersive storytelling. Additionally, ongoing research efforts aimed at enhancing the interpretability, controllability, and scalability of these models will further fuel their adoption in diverse creative domains, paving the way for transformative changes in how we conceive, design, and interact with digital content.

\subsubsection{3D Scene Generation}
3D scene generation refers to the creation of immersive and realistic environments in virtual worlds, games, simulations, and architectural visualization. This application domain leverages multimodal generative models to synthesize complex 3D scenes comprising objects, textures, lighting, and spatial arrangements. The ability to generate 3D scenes has profound implications for various industries, including entertainment, education, design, and virtual reality.

In the context of games and virtual environments, multimodal generative models can automate the process of scene creation, reducing the reliance on manual modeling and asset creation. By inputting textual descriptions or conceptual sketches, developers can generate entire 3D environments populated with interactive objects, characters, and atmospheric effects. This not only accelerates the game development pipeline but also enables the creation of dynamic and immersive gameplay experiences.
Moreover, in architectural visualization and design, multimodal generative models can assist architects, urban planners, and designers in visualizing and exploring different design options. By inputting architectural blueprints or design parameters, designers can generate realistic 3D renderings of buildings, landscapes, and interior spaces, allowing for rapid iteration and exploration of design concepts. This facilitates collaboration, decision-making, and communication among stakeholders involved in the design process.

By harnessing the capabilities of multimodal generative models, 3D scene generation might revolutionize how virtual environments are created, experienced, and interacted with. Whether in games, simulations, or architectural visualization, the ability to generate immersive and realistic 3D scenes programmatically opens up new possibilities for creativity, exploration, and storytelling in virtual worlds.

\subsubsection{Customizable Avatars}
Customizable avatars represent digital representations of users that can be personalized and adapted to reflect individual preferences, identities, and characteristics. 

Multimodal generative models offer a compelling approach to customizable avatar creation by synthesizing diverse media types such as images, text, and audio to create lifelike and expressive avatars. For example, generative models trained on multimodal data can generate photorealistic images of avatars based on textual descriptions or user preferences, incorporating details such as facial features, clothing styles, and expressions. Similarly, audio-based avatars can be generated using voice synthesis techniques, enabling avatars to communicate with users using natural-sounding voices that reflect their personalities or preferences. By enabling the creation of customizable avatars across multiple modalities, multimodal generative models empower users to express themselves in virtual environments, fostering deeper engagement and personalization in social interactions, gaming, and virtual communication platforms.

Currently, there are several aspects that could be further studied: 1) Personalization and Customization: Multimodal generative models can generate avatars that closely resemble users based on input parameters such as facial features, body type, and clothing preferences. Users can interactively customize their avatars using intuitive interfaces, adjusting attributes such as hairstyle, facial expression, and accessories in real time.
2) Emotional Expression and Limb Movements: Avatars generated by multimodal models can exhibit a wide range of emotional expressions, gestures and physical movements, enhancing their ability to convey nonverbal communication cues in virtual environments. Users can dynamically control their avatar's behaviors, allowing for more immersive social interactions and collaborative experiences in virtual worlds.
3) Integration with Virtual Environments: Customizable avatars can be seamlessly integrated into various virtual environments, including social platforms, online games, and virtual reality applications. Users can navigate these environments using their avatars, interacting with other users and objects in real time, fostering a sense of presence and belonging in digital spaces.

\subsection{Towards the World Model}
World models~\cite{ha2018world,liu2024world,lecun2022path,min2023uniworld,hafner2023mastering} have recently emerged as a hotspot topic. Many renowned researchers have expressed that world models will come true in the foreseeable future, and researchers around the world hold great expectations for this development. We found that all the topics mentioned in the survey correspond precisely to the principal components of the world modeling, encompassing perceptual modalities such as vision, audition, and speech, as well as spatial understanding and generation. Once world models advance to a usable stage, they will endow numerous industries with new possibilities. We here highlight several core applications for reference.

\textbf{Multimodal education and communication.} World models hold immense promise for revolutionizing education and communication by facilitating multimodal learning experiences and immersive interactions. By integrating diverse sensory modalities such as text, images, audio, and video, these models enable the creation of rich educational content that caters to different learning styles and preferences. Furthermore, they empower learners to engage with complex concepts and environments more intuitively and interactively, thereby enhancing comprehension and retention. Additionally, world models facilitate seamless communication by enabling the synthesis of natural and expressive multimodal dialogue, fostering more engaging and personalized interactions in virtual learning environments and online collaboration platforms.

\textbf{Movie Generation.} The application of world models in movie generation represents a paradigm shift in filmmaking, offering filmmakers unprecedented creative freedom and flexibility. By leveraging multimodal generative techniques, filmmakers can seamlessly integrate various elements such as dialogue, visuals, sound effects, and music to craft immersive cinematic experiences that resonate with audiences on a deeper level. Moreover, world models enable the generation of dynamic and personalized narratives tailored to individual viewer preferences, thereby enhancing viewer engagement and immersion. Furthermore, these models facilitate the exploration of alternative storytelling formats and experimental filmmaking techniques, pushing the boundaries of cinematic creativity and expression.

\textbf{Metaverse.} The emergence of the metaverse presents exciting opportunities for leveraging world models to create immersive and interactive virtual worlds. By synthesizing multimodal sensory experiences, including visual, auditory, and haptic feedback, these models enable the creation of highly realistic and immersive virtual environments that blur the boundaries between physical and digital reality. Moreover, world models facilitate the development of intelligent virtual agents and NPCs that exhibit lifelike behaviors and interactions, enhancing the sense of presence and social immersion within the metaverse. Additionally, these models empower users to customize and personalize their virtual experiences, fostering creativity and exploration within digital worlds.

\section{Conclusion}
In this survey, we systematically review multimodal editing and generation works augmented by LLMs, delving into the advancements across various modalities including image, video, 3D and audio. Following that, multimodal agents that integrate a plethora of state-of-the-art generative models are thoroughly discussed with diverse case studies. 
We also investigate the concern for the safety of multimodal generative models, which play an indispensable role in practical applications. Our comprehensive review highlights the significant contributions of LLMs in enhancing the quality and capabilities of generative systems. 
Looking forward, we anticipate further innovations at the intersection of AI and generative content, driving progress toward a more unified and capable multimodal generation framework. 
In conclusion, we ardently expect that our investigation will provide insights and inspiration for the development of the multimodal generation, particularly the world models, which have garnered the attention and anticipation of the majority of researchers.

\clearpage
\bibliographystyle{IEEEtran}
\bibliography{references}

\end{document}